\documentclass[11pt]{article}
\usepackage[preprint]{acl}

\usepackage{amsmath}
\usepackage{amssymb}     
\usepackage{amsthm}
\newtheorem{definition}{Definition}[section]

\usepackage{times}
\usepackage{latexsym}

\usepackage{graphicx}
\usepackage{subcaption}
\usepackage{svg}
\usepackage{tikz}
\usetikzlibrary{positioning}

\usepackage{booktabs}
\usepackage{tabularx}
\usepackage{array}
\newcolumntype{Y}{>{\raggedright\arraybackslash}X}
\newcolumntype{M}[1]{>{\centering\arraybackslash}m{#1}}

\usepackage{makecell}
\usepackage{bookmark}
\usepackage{etoolbox}
\usepackage{placeins}
\usepackage[hang,flushmargin]{footmisc}
\usepackage{needspace}
\usepackage{xcolor}

\usepackage{hyperref}
\usepackage{url}

\usepackage[ruled,vlined]{algorithm2e}
\SetKwInput{KwIn}{Input}
\SetKwInput{KwOut}{Output}
\SetKw{Return}{return}
\SetKw{True}{true}
\SetKw{False}{false}

\usepackage{fontawesome5}

\usepackage{listings}
\usepackage[most]{tcolorbox}
\tcbuselibrary{listings,skins}

\usepackage{textcomp} 

\IfFileExists{latexml.sty}{\usepackage{latexml}}{}
\makeatletter
\@ifundefined{iflatexml}{%
  \newif\iflatexml
  \latexmlfalse
}{}
\makeatother

\tcbset{colback=gray!4,colframe=gray!40,boxrule=0.5pt,arc=2mm}

\iflatexml
  \usepackage{fancyvrb}

  \newenvironment{promptbox}[2][]{%
    \par\smallskip\noindent\textbf{#2}\par
    \VerbatimEnvironment
    \begin{Verbatim}[fontsize=\small]
  }{%
    \end{Verbatim}\par\smallskip
  }

  \newenvironment{tightpromptbox}[2][]{%
    \par\smallskip\noindent\textbf{#2}\par
    \VerbatimEnvironment
    \begin{Verbatim}[fontsize=\small]
  }{%
    \end{Verbatim}\par\smallskip
  }

\else
  \newtcblisting{promptboxwide}[2][]{%
    enhanced,
    colback=gray!4, colframe=gray!40, coltitle=black, boxrule=0.5pt, arc=2mm,
    top=1mm,bottom=1mm,left=1.2mm,right=1.2mm,
    width=\textwidth,
    title={\strut #2}, fonttitle=\bfseries\footnotesize,
    attach boxed title to top left={xshift=1mm,yshift=-1mm},
    boxed title style={colback=white, colframe=gray!60, boxrule=0.5pt},
    listing only, listing engine=listings,
    listing options={
      basicstyle=\ttfamily\footnotesize,
      breaklines=true, breakatwhitespace=false,
      postbreak=\mbox{\textcolor{gray}{$\hookrightarrow$}\space},
      columns=fullflexible, keepspaces=true, showstringspaces=false, upquote=true, tabsize=2,
      literate={•}{{\textbullet}}1 {–}{{-}}1 {—}{{---}}3
               {“}{{``}}1 {”}{{''}}1 {‘}{{`}}1 {’}{{'}}1
               {…}{{\ldots}}1 {×}{{$\times$}}1 {≤}{{$\le$}}1 {≥}{{$\ge$}}1
               {π}{{$\pi$}}1 {✓}{{\checkmark}}1 {✗}{{\texttimes}}1 {→}{{$\rightarrow$}}1
               {<}{{\textless}}1 {>}{{\textgreater}}1
    },
    float*=t, floatplacement=tp,
    before skip=0pt, after skip=0.5\baselineskip,
    #1
  }

  \newtcolorbox{tightpromptbox}[2][]{%
    enhanced,
    breakable,
    title={#2},
    colback=white,
    colframe=black!35,
    boxrule=0.6pt,
    arc=1.5pt,
    left=4pt,right=4pt,top=3pt,bottom=3pt,
    boxsep=2pt,
    before skip=6pt,
    after skip=10pt,
    fontupper=\small\ttfamily,
    fonttitle=\bfseries,
    before=\Needspace{12\baselineskip},
    #1
  }

  \newtcblisting{promptbox}[2][]{%
    enhanced, unbreakable,
    colback=gray!4, colframe=gray!40, coltitle=black, boxrule=0.5pt, arc=2mm,
    top=1mm,bottom=1mm,left=1.2mm,right=1.2mm,
    width=\columnwidth,
    title={\strut #2}, fonttitle=\bfseries\footnotesize,
    attach boxed title to top left={xshift=1mm,yshift=-1mm},
    boxed title style={colback=white, colframe=gray!60, boxrule=0.5pt},
    listing only, listing engine=listings,
    listing options={
      basicstyle=\ttfamily\footnotesize,
      breaklines=true, breakatwhitespace=false,
      postbreak=\mbox{\textcolor{gray}{$\hookrightarrow$}\space},
      columns=fullflexible, keepspaces=true, showstringspaces=false, upquote=true, tabsize=2,
      literate={•}{{\textbullet}}1 {–}{{-}}1 {—}{{---}}3
               {“}{{``}}1 {”}{{''}}1 {‘}{{`}}1 {’}{{'}}1
               {…}{{\ldots}}1 {×}{{$\times$}}1 {≤}{{$\le$}}1 {≥}{{$\ge$}}1
               {π}{{$\pi$}}1 {✓}{{\checkmark}}1 {✗}{{\texttimes}}1 {→}{{$\rightarrow$}}1
               {<}{{\textless}}1 {>}{{\textgreater}}1
    },
    #1
  }
\fi
\newcommand{\lt}{\textless}
\newcommand{\gt}{\textgreater}
\newcommand{\tagtt}[1]{\texttt{\lt#1\gt}}
\newcommand{\etagtt}[1]{\texttt{\lt/#1\gt}}

\usepackage{enumitem}
\setlist[enumerate]{leftmargin=*, itemsep=0pt, topsep=0pt}

\usepackage[T1]{fontenc}
\usepackage[utf8]{inputenc}
\usepackage{microtype}
\usepackage{inconsolata}

\makeatletter
\newif\iflatexml
\IfFileExists{latexml.sty}{\usepackage{latexml}}{}
\@ifundefined{iflatexml}{%
  \newif\iflatexml
  \latexmlfalse
  \@ifundefined{LaTeXML}{}{ \latexmltrue }
  \@ifundefined{latexml}{}{ \latexmltrue }
  \@ifundefined{latexmlversion}{}{ \latexmltrue }
}{}
\makeatother

\newcommand{\hficon}{\raisebox{-0.1em}{\includegraphics[height=1.1em]{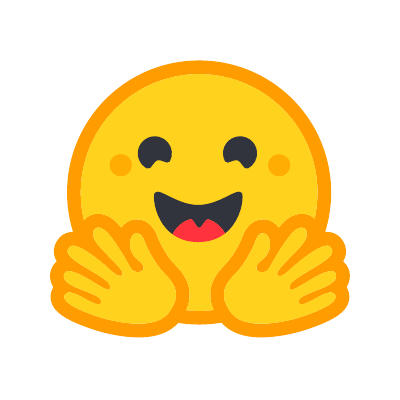}}}

\title{The Illusion of Insight in Reasoning Models}

\author{
  Liv G.~d'Aliberti\\
  Princeton University\\
  Department of Computer Science\\
  Princeton, NJ, USA\\
  \texttt{liv.daliberti@princeton.edu}
  \And
  Manoel Horta Ribeiro\\
  Princeton University\\
  Department of Computer Science\\
  Princeton, NJ, USA\\
  \texttt{manoel@cs.princeton.edu}
}

\iflatexml
  \renewcommand{\hficon}{Hugging Face}
  \newcommand{\codeanddata}{%
    Code: \url{https://github.com/humans-and-machines/Illusion-of-Reasoning}\quad
    Data: \url{https://huggingface.co/datasets/od2961/illusion-of-reasoning-main-traces}%
  }
\else
  \newcommand{\codeanddata}{%
    \href{https://github.com/humans-and-machines/Illusion-of-Reasoning}{\faGithub\ Code},%
    \href{https://huggingface.co/datasets/od2961/illusion-of-reasoning-main-traces}{\hficon\ Data}%
  }
\fi

\newcommand\blfootnote[1]{%
  \begingroup
  \renewcommand\thefootnote{}\footnote{#1}%
  \addtocounter{footnote}{-1}%
  \endgroup
}

\begin{document}

\iflatexml\else\maketitle\fi  

\begin{abstract}
Do reasoning models have ``Aha!'' moments?
Prior work suggests that models like DeepSeek-R1-Zero undergo sudden mid-trace realizations that lead to accurate outputs, implying an \emph{intrinsic} capacity for self-correction. 
Yet, it remains unclear whether such intrinsic shifts in reasoning strategy actually improve performance.
Here, we study mid-reasoning shifts and instrument training runs to detect them. 
Our analysis spans $1$M+ reasoning traces, hundreds of training checkpoints, three reasoning domains, and multiple decoding temperatures and model architectures.
We find that reasoning shifts are rare, do not become more frequent with training, and seldom improve accuracy, indicating that they do not correspond to prior perceptions of model insight. 
However, their effect varies with model uncertainty. Building on this finding, we show that artificially triggering \emph{extrinsic} shifts under high entropy reliably improves accuracy. 
Our results show that mid-reasoning shifts are symptoms of unstable inference behavior rather than an intrinsic mechanism for self-correction.
\end{abstract}

\section{Introduction}

\begin{figure}
    \centering
    \includegraphics[width=\linewidth]{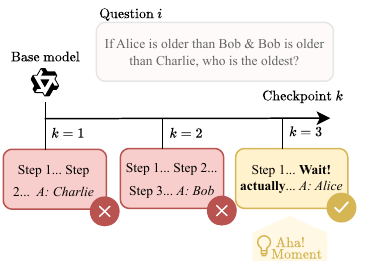}
    \caption{\textbf{Anatomy of an ``Aha!'' Moment.}
    We illustrate an ``Aha!'' moment as described in \citet{deepseekai2025deepseekr1incentivizingreasoningcapability}: within a single chain-of-thought, a cue such as ``Wait... let's re-evaluate'' marks a shift from an initially failing strategy ($k \in \{1,2\}$) to one that yields a correct answer (when $k=3$).
    The figure also anticipates our methodology: we study ``Aha!'' moments by systematically GRPO-tuning and annotating the reasoning traces of Qwen2.5 and Llama models.
    }
    \label{fig:motivation}
    \vspace{-3mm}
\end{figure}

Anecdotal evidence suggests that language models fine-tuned with reinforcement learning exhibit ``Aha!'' moments---episodes of apparent insight reminiscent of human problem-solving. For example, \citet{deepseekai2025deepseekr1incentivizingreasoningcapability} highlight mid-trace cues such as \textit{``Wait... let's re-evaluate step-by-step,''} which sometimes accompany correct answers. Yet, the nature, frequency, and impact of these events (Fig.~\ref{fig:motivation}) remain unclear \citep{yang2025understandingahamomentsexternal}.
\blfootnote{\noindent \codeanddata}

The existence of ``Aha!'' moments is linked to whether reasoning models can \textit{intrinsically} self-correct, i.e., revise their reasoning mid-response without external feedback. Model improvements often arise from \textit{extrinsic} mechanisms like verifiers, reward queries, prompting techniques, or external tools \citep{lightman2024letsverifystepstep, li2024selfpromptinglargelanguagemodels, zhang-etal-2024-small}. 
In contrast, intrinsic self-improvement must be inferred from reasoning traces and is arguably more safety-relevant, as it implies that a model can reorganize its reasoning from internal state alone~\citep{tsui2025selfcorrectionbenchrevealingaddressing, liu2025there}.

Studying the effect of reasoning shifts is challenging. First, these events may occur (and affect performance) \emph{during} training, yet evaluations are typically conducted only post-training \citep{zeng2024mrbenmetareasoningbenchmarkevaluating, xia2025reasoneval}. Second, reasoning models rarely release mid-training checkpoints, limiting longitudinal analyses across the training lifecycle. Third, even when shifts are observed, attributing correctness to a mid-trace change (rather than to general competence or memorization) requires systematically controlled comparisons.
\textit{This gap motivates the need for a systematic investigation of whether reasoning shifts reflect genuine insight.}

\vspace{2mm}
\noindent
\textbf{Present work.}
Here, we investigate whether mid-trace reasoning shifts (e.g., ``Wait... let’s re-evaluate'') signal intrinsic self-correction in reasoning models. Our study is guided by three questions:

\vspace{3mm}
\noindent
\textbf{RQ1}: Do reasoning shifts raise model accuracy?

\vspace{2mm}
\noindent
\textbf{RQ2}: How does the effect of reasoning shifts vary with training stage and decoding temperature?

\vspace{2mm}
\noindent
\textbf{RQ3}: Are reasoning shifts more effective when reasoning models are uncertain?
\vspace{3mm}

To answer these, we (i) formalize ``Aha!'' moments as measurable mid-trace shifts in reasoning that improve performance on problems that were previously unsolved by the model \citep{yang2025understandingahamomentsexternal, zhou2025safekey, hu2025beyond} (Fig.~\ref{fig:aha-moment}; \S\ref{subsec:aha-moment-def}); (ii) curate a diverse evaluation suite (\S\ref{sec:data}) spanning cryptic crosswords \citep{efrat-etal-2021-cryptonite}, mathematical problem-solving (\textsc{MATH-500}) \citep{lightman2024letsverifystepstep}, and Rush Hour puzzles \citep{fogleman2018rushhour}; and (iii) GRPO-tune and annotate the reasoning traces of Qwen2.5 and Llama models (\S\ref{sec:methods}).

Our analysis spans $1$M+ annotated reasoning traces across hundreds of checkpoint evaluations (10--20 per model/run), 3 domains, 4 temperatures, 2 model sizes, and 2 model architectures, providing a longitudinal view of how mid-trace reasoning evolves during RL fine-tuning. With this setup, we connect shift behavior to both correctness and token-level uncertainty signals \citep{ton2025infotheory}.

Our results show that reasoning shifts are rare (overall $\sim$6.31\% of traces) and generally do not improve model accuracy (\textbf{RQ1}). We further find that their impact on accuracy does not reliably flip sign across training stages, but varies substantially with decoding temperature (\textbf{RQ2}). Finally, we find that spontaneously occurring shifts do not become reliably helpful under high uncertainty; however, \emph{externally triggered} reconsideration under high entropy improves accuracy across benchmarks, including a \textbf{+8.41pp} improvement on \textsc{MATH-500} (and smaller gains on crosswords and Rush Hour) (\textbf{RQ3}). Our results are robust across datasets, prompts, and model families.

\vspace{2mm}
\noindent
\textbf{Contributions.}
We make three key contributions:
\vspace{1mm}
\begin{enumerate}
    \item \textbf{Definition \& framework.} We formalize ``Aha!'' moments as measurable mid-trace shifts and introduce an experimental framework for studying intrinsic self-correction during RL fine-tuning.
    \item \textbf{Empirical characterization at scale.} Across $1$M+ traces spanning domains, temperatures, training stages, and model families, we show that reasoning shifts are rare and typically coincide with \emph{lower} accuracy, challenging the view that they reflect genuine insight.
    \item \textbf{Intervention.} We develop an entropy-gated intervention that \emph{induces} reconsideration when models are uncertain, yielding measurable accuracy gains.
\end{enumerate}

\section{Related Work}
\label{sec:related}

\textbf{Emergent Capabilities.}
Large language models often \emph{appear} to acquire new abilities abruptly with scale—such as multi-step reasoning or planning \citep{wei2022emergent, berti2025emergent}—but it remains debated whether these shifts reflect intrinsic cognitive change or artifacts of evaluation~\citep{schaeffer2023mirage, shojaee2025illusionthinkingunderstandingstrengths}. Many behaviors labeled as ``emergent'' arise only under \textit{extrinsic} scaffolds. Structured prompts—e.g., Chain-of-Thought \citep{wei2022chainofthought}, the zero-shot cue ``Let’s think step by step'' \citep{kojima2022large}, or Least-to-Most prompting \citep{zhou2023leasttomostpromptingenablescomplex}—elicit intermediate reasoning that models rarely produce on their own. Optimization methods such as SFT \citep{wolfe2023sft}, RLHF \citep{ouyang2022training}, and GRPO \citep{shao2024deepseekmathpushinglimitsmathematical} reinforce these externally induced behaviors, potentially amplifying the appearance of intrinsic ability gains.

\vspace{2mm}
\noindent
\textbf{Self-Correction and ``Aha!'' Moments.}
Self-correction in reasoning models can arise through \textit{extrinsic} mechanisms—such as verifier models or tool calls \citep{lightman2024letsverifystepstep, li2024selfpromptinglargelanguagemodels}—or through \textit{intrinsic} shifts that occur without any external intervention \citep{liu2024largelanguagemodelsintrinsic}. Recent work has examined these dynamics, including frameworks for trained self-correction \citep{kumar2024traininglanguagemodelsselfcorrect} and benchmarks for iterative refinement \citep{madaan2023selfrefine, tsui2025selfcorrectionbenchrevealingaddressing}, and analyses of mid-inference adjustments \citep{wu-etal-2024-large}. Studies of models such as DeepSeek-R1 \citep{deepseekai2025deepseekr1incentivizingreasoningcapability} suggest that reward optimization can induce \emph{intrinsic} reflection-like artifacts. However, other works have raised doubts about whether observed reasoning shifts reflect genuine insight or superficial self-reflection \citep{liu2025there, ton2025infotheory}. Yet, there has been no systematic evaluation of whether RL-trained models exhibit true intrinsic ``Aha!''-style self-correction \emph{throughout RL fine-tuning}, nor whether such shifts reliably improve correctness when tracked across checkpoints and decoding regimes.

\vspace{2mm}
\noindent
\textbf{Insight Characterization.}
In cognitive psychology, insight is classically defined as an abrupt restructuring of the problem space, exemplified by \citet{kohler1921intelligenzpruefungen}'s chimpanzees stacking boxes to reach bananas. Recent work seeks analogous phenomena in reasoning models: mid-trace uncertainty spikes—sometimes described as ``Gestalt re-centering''—have been associated with shifts in reasoning strategy \citep{ton2025infotheory, yang2025understandingahamomentsexternal}. Metrics such as RASM aim to identify linguistic or uncertainty-based signatures of genuine insight \citep{yang2025understandingahamomentsexternal}, yet existing approaches misclassify superficial hesitations as insight at high rates in some settings (up to 30\%) \citep{zheng-etal-2023-chain, xia2025reasoneval}. These limitations highlight the need for rigorous criteria to distinguish genuine restructurings from superficial reflection.

\vspace{2mm}
\noindent
\textbf{Safety, Faithfulness, and Alignment.}
Transparent reasoning traces are central to alignment and faithfulness, as they allow human oversight of not only a model’s outputs but the process that produces them \citep{uesato2022solving, openai2023processsupervision}. When self-corrections occur without oversight, they raise concerns about hidden objective shifts or deceptive rationales that can mislead users \citep{su2025surveyautonomyinducedsecurityrisks, baker2025monitoring, lanham2023measuringfaithfulness, zhang-etal-2025-understanding}. Process supervision—rewarding intermediate reasoning steps rather than only final answers—has been shown to improve both performance and interpretability in math reasoning tasks \citep{uesato2022solving, openai2023processsupervision}. Complementing this, uncertainty-aware methods help models detect and respond to unreliable reasoning (e.g., via abstention or filtering when uncertainty is high), improving robustness and trustworthiness~\citep{zhou2025safekey, skaf2025largelanguagemodelslearn}. Understanding whether mid-trace shifts reflect genuine correction or uncertainty-driven artifacts is therefore directly relevant to evaluating the safety and reliability of reasoning models.

\section{Formalizing ``Aha!'' Moments}
\label{subsec:aha-moment-def}

We define an ``Aha!'' moment as a discrete point within a model's chain-of-thought where the model abandons its initial reasoning strategy and adopts a qualitatively different one that improves performance.
We formalize this notion below.

Let $\{f_{\theta_k}\}_{k=0}^K$ denote a sequence of checkpointed reasoning models.
At checkpoint $k$, the model defines a policy $\pi_{\theta_k}(a_t \mid a_{<t}, q)$ over token actions $a_t \in \mathcal{V}$. A reasoning trace is a trajectory $\tau = (a_1,\ldots,a_T)$ whose quality is measured by its correctness $R(\tau)$. For a question $q_j$, let
\[
P_{\theta_k}(\checkmark \mid q_j) = \mathbb{E}_{\tau \sim \pi_{\theta_k}}[R(\tau)]
\]
denote expected correctness. Let $S_{q_j,k}(\tau) \in \{0,1\}$ indicate whether a mid-trace shift occurs in a sampled trajectory $\tau$ at checkpoint $k$. This binary label is produced by our shift-detection pipeline, which identifies lexical and structural changes in reasoning strategy (detailed in App.~\ref{sec:app-algorithm}). We write $P(S_{q_j,k}=1)$ for the probability (under $\tau \sim \pi_{\theta_k}$) that a sampled trace contains a detected shift.

\begin{definition}[``Aha!'' Moment]\label{def:aha-moment-lrms}
Let $\delta_1,\delta_2,\delta_3 \in [0,1]$ be thresholds for prior failure, prior stability, and required performance gains. An ``Aha!'' moment occurs for $(q_j,k)$ iff:
\begin{enumerate}[leftmargin=*]
  \item $\forall i < k,\; P_{\theta_i}(\checkmark \mid q_j) < \delta_1$ \quad (\textit{Prior failures}),
  \item $\forall i < k,\; P(S_{q_j,i}=1) < \delta_2$ \quad (\textit{Prior stability}),
  \item $P_{\theta_k}(\checkmark \mid q_j, S_{q_j,k}=1) - P_{\theta_k}(\checkmark \mid q_j) > \delta_3$ \quad (\textit{Performance gain}).
\end{enumerate}
\end{definition}

\begin{figure}[t]
    \centering
    \includegraphics[width=\linewidth]{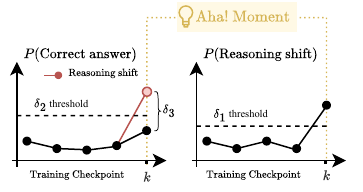}
    \caption{\textbf{Schematic of our operational ``Aha!'' definition.}
    For a fixed problem $q_j$ (horizontal axis: checkpoint index $i$), the figure visualizes the three criteria in Def.~\ref{def:aha-moment-lrms}.
    (1) \textit{Prior failures}: empirical correctness $\hat P_{\theta_i}(\checkmark \mid q_j)$ remains below $\delta_1$ at all checkpoints $i<k$.
    (2) \textit{Prior stability}: the shift rate $\hat\pi_i = \Pr[S_{q_j,i}=1]$ stays below $\delta_2$ for all $i<k$.
    (3) \textit{Performance gain}: at checkpoint $k$, correctness on traces \emph{with} a detected shift (red) exceeds correctness over \emph{all} traces (black) by more than $\delta_3$.}
    \label{fig:aha-moment}
    \vspace{-3mm}
\end{figure}

In plain terms, a checkpoint $k$ qualifies as an ``Aha!'' moment for $q_j$ if: (1) all earlier checkpoints consistently fail on the problem (\textit{prior failures}); (2) earlier checkpoints show little evidence of mid-trace shifts (\textit{prior stability}); and (3) at checkpoint $k$, traces containing a detected shift yield a strictly higher correctness rate than traces overall (\textit{performance gain}).\footnote{Formal ``Aha!'' events are defined over problem--checkpoint pairs \((q_j,k)\) (i.e., a checkpoint-level comparison for a fixed problem), not over individual sampled traces.} Together, these conditions ensure that a detected shift is both \emph{novel} and \emph{beneficial}, preventing superficial or noisy variations from being counted as insight-like events. Figure~\ref{fig:aha-moment} illustrates this behavior. Algorithm~\ref{alg:aha-moment} in App.~\ref{sec:app-algorithm} formalizes the detection procedure.

\begin{figure*}
    \centering
    \includegraphics[width=\linewidth]{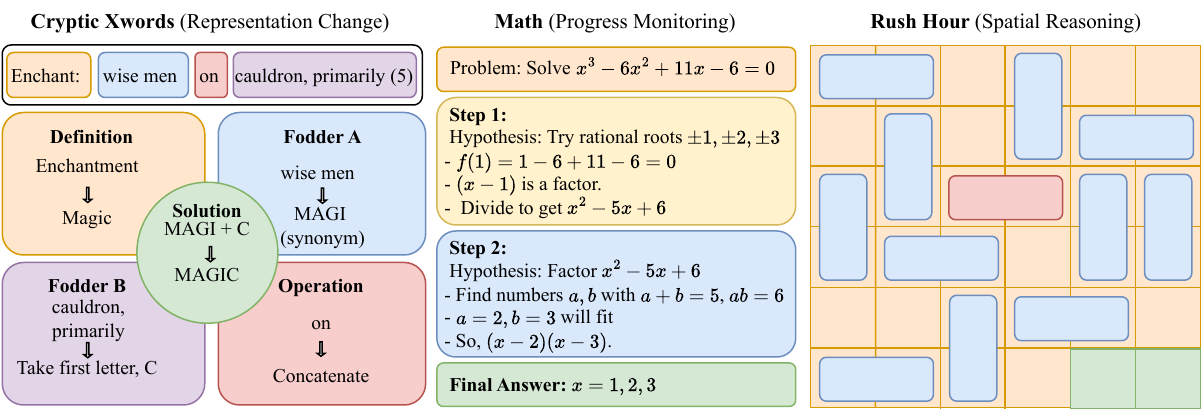}
    \caption{\textbf{Three reasoning lenses and example instances.}
    Each row illustrates one evaluation domain and how it instantiates the three ``reasoning lenses’’ introduced in \S\ref{sec:data}.
    \emph{Left (representation change):} a cryptic Xwords clue with definition and wordplay; shifts correspond to re-parsing the clue (e.g., switching from anagram to charade or hidden-word).
    \emph{Center (progress monitoring):} a math problem with explicit chain-of-thought and checks; shifts occur when the model abandons an inconsistent derivation and restarts with a new method.
    These domains form complementary testbeds for studying when mid-trace shifts (our ``Aha!'' events; Def.~\ref{def:aha-moment-lrms}) co-occur with changes in uncertainty and accuracy.
    \emph{Right (spatial manipulation):} a RHour puzzle requiring a planned sequence of legal moves; mid-trace shifts reflect abandoning one move plan for another.
    }
    \label{fig:triad-overview}
    \vspace{-3mm}
\end{figure*}

The thresholds $(\delta_1,\delta_2,\delta_3)$ act as tunable criteria: stricter values prioritize precision by requiring consistent prior failure and rare prior shifts, while looser values increase recall. In our experiments, we select these thresholds on a held-out development slab and validate robustness using bootstrap confidence intervals (App.~\ref{app:formal-threshold-search-q15b}). In all cases, probabilities such as $P_{\theta_k}(\checkmark \mid q_j)$ and $P_{\theta_k}(\checkmark \mid q_j, S_{q_j,k}=1)$ are estimated from finitely many sampled traces per $(q_j,k)$.

This definition parallels the classical cognitive characterization of insight: a sudden restructuring of the problem space that enables solution \citep{jones2003insight, kohler1921intelligenzpruefungen, duncker1945problem, kaplan1959dunkercandleproblem}. Hallmarks of such shifts include explicit self-reflective cues (e.g., ``wait,'' ``let’s reconsider’’) and an observable pivot in strategy \citep{deepseekai2025deepseekr1incentivizingreasoningcapability}.
Theoretical accounts such as representational change theory \citep{o1987insightproblemrepresentationalchangetheory}, progress monitoring theory \citep{macgregor2001progress}, and Gestalt perspectives on problem-solving \citep{metcalfe1987insight} provide complementary lenses for interpreting analogous shifts in reasoning models.

\section{Data}
\label{sec:data}

Our evaluation suite spans three complementary reasoning lenses (Fig.~\ref{fig:triad-overview}): representational change in cryptic Xwords (\textit{left}), quantitative problem solving (\textit{center}), and spatial reasoning in RHour--style puzzles (\textit{right}). Each domain offers automatic correctness checks, natural opportunities for mid-trace verification, and structured signals of strategy. All data are in English; dataset sizes and splits are summarized in Table~\ref{tab:data-stats}, and additional filtering and scoring details are provided in App.~\ref{app:data-details}. Throughout, we score answers by \emph{normalized exact match} (canonicalizing case, whitespace, and punctuation before exact comparison; App.~\ref{app:data-details}).

\vspace{1.5mm}\noindent
\textbf{Cryptic Xwords.}
Cryptic Xwords clues hide a wordplay instruction (e.g., anagram, abbreviation, homophone) beneath a misleading surface reading, requiring representational shifts to solve. We train on natural clues from \textsc{Cryptonite}~\citep{efrat-etal-2021-cryptonite} and evaluate on a synthetic test set with device-balanced templates (App.~\ref{app:data-details}), scoring by normalized exact match.

\vspace{1.5mm}\noindent
\textbf{Math.}
Math word problems test symbolic manipulation and multi-step deduction, with reasoning progress naturally expressed step-by-step. We train on \texttt{openR1 Math-220k}~\citep{openr1} and evaluate on \textsc{MATH-500}~\citep{lightman2024letsverifystepstep}, ensuring no train/eval leakage (App.~\ref{app:data-details}). Answers are scored by normalized exact match.

\vspace{1.5mm}\noindent
\textbf{RHour.}
We synthetically generate \textsc{RHour} sliding-block puzzles, where the goal is to free a target car from a crowded grid by moving obstructing vehicles. We generate balanced $4{\times}4$, $5{\times}5$, and $6{\times}6$ boards and evaluate on $6{\times}6$ only. Boards are solved optimally via BFS with per-size node caps, discarding timeouts~\citep{fogleman2018rushhour} (App.~\ref{app:data-details}). We filter trivial cases and stratify remaining instances into easy ($<$4 moves), medium ($<$6), and hard ($\geq$6) buckets by solution length.

\section{Methods} 
\label{sec:methods} 

We fine-tune reasoning models with GRPO across evaluation domains (\S\ref{ss:mat}); collect and annotate reasoning traces during training (\S\ref{ss:rtc}); and estimate model uncertainty to trigger entropy-based interventions (\S\ref{ss:unc}).

\subsection{Models and Training}
\label{ss:mat}

Motivated by claims of mid-trace ``Aha!'' behavior in DeepSeek-R1~\citep{deepseekai2025deepseekr1incentivizingreasoningcapability}, we adopt Group Relative Policy Optimization (GRPO)~\citep{shao2024deepseekmathpushinglimitsmathematical} as our fine-tuning method. GRPO is an RLHF-style algorithm~\citep{ouyang2022training} that optimizes chain-of-thought generation by comparing groups of sampled completions and extends PPO~\cite{schulman2017ppo} with group-normalized advantages and KL regularization to a frozen reference policy. Full implementation details appear in App.~\ref{app:grpo-setup}.

We fine-tune Qwen2.5~\citep{qwen2p5} and Llama~3.1~\citep{dubey2024llama3herd} models on each domain for up to 1{,}000 steps. Our primary experiments use Qwen2.5-1.5B trained for 1{,}000 steps ($\approx$2.5--3 epochs per domain), while larger models (Qwen2.5-7B and Llama 3.1-8B) are evaluated at 500 steps due to compute constraints. To verify that models improve during training, we evaluate at multiple checkpoints and report accuracy at initialization (Step~0) and at the final evaluated checkpoint (Step~950 for the 1.5B runs; Step~500 for 7B/8B). Table~\ref{tab:models-tasks} summarizes coverage and accuracy gains.

We use lightweight, task-specific prompts that structure reasoning into a \tagtt{think} block and a concise final answer in \tagtt{answer}, with domain-level checks that invite reconsideration (App.~\ref{app:system-level-prompts}). Informed by established strategies—zero-shot CoT, self-consistency, and reflection routines \citep{kojima2022large,wang2023selfconsistency,madaan2023selfrefine,shinn2023reflexion}—these prompts standardize mid-trace events as reasoning shifts (Def.~\ref{def:aha-moment-lrms}; Alg.~\ref{alg:aha-moment}), enabling consistent comparison across models, tasks, and checkpoints.

\begin{table}[t]
\centering
\small
\setlength{\tabcolsep}{3pt}
\renewcommand{\arraystretch}{0.95}
\begin{tabular*}{\columnwidth}{@{\extracolsep{\fill}} ll r r c r@{}}
\toprule
\textbf{Model} & \textbf{Domain} & \textbf{Step 0} & \textbf{After} & \textbf{Step} & $\boldsymbol{\Delta}$ \\
\midrule
Qwen2.5-1.5B  & Xwords     & 7.69 & 10.00 & 950 & +2.31 \\
Qwen2.5-1.5B  & Math       & 31.00 & 35.00 & 950 & +4.00 \\
Qwen2.5-1.5B  & RHour      & 0.00 & 0.01 & 950 & +0.01 \\
Qwen2.5-7B    & Math       & 61.60 & 66.40 & 500 & +4.80 \\
Llama\,3.1-8B & Math       & 40.20 & 48.36 & 500 & +8.16 \\
\bottomrule
\end{tabular*}
\caption{\textbf{Model coverage and learning progress.}
Accuracy at initialization (Step~0) and at the final training checkpoint, along with the absolute gain ($\Delta$).
All results are 1-shot evaluations at temperature $0$ on the fixed test sets described in \S\ref{sec:data}.
}
\label{tab:models-tasks}
\vspace{-3mm}

\end{table}

\subsection{Trace Collection and Annotation}
\label{ss:rtc}

We evaluate each model at a fixed cadence of \emph{every 50 training steps} from initialization (Step~0) to Step~950 \emph{inclusive} (i.e., checkpoints $k\in\{0,50,\ldots,950\}$), yielding \emph{20 checkpoints per run}. At each checkpoint, we generate \textbf{$G{=}8$} completions per problem using a fixed decoding policy (temperature $\{0,0.05,0.3,0.7\}$, top-$p{=}0.95$). Each completion follows the tag-structured output contract in App.~\ref{app:system-level-prompts}, with private reasoning in \tagtt{think} and a machine-checkable final response in \tagtt{answer}; token budgets and stop criteria are domain-specific and held fixed across checkpoints.

Evaluation sets are held fixed across checkpoints: \textbf{500} problems for \textsc{MATH-500}, \textbf{130} synthetic clues for Xwords, and \textbf{500} $6{\times}6$ RHour boards. For our Qwen2.5-1.5B models, because each item is evaluated at all 20 checkpoints across $T{=}4$ temperatures with $G{=}8$ samples, each run yields 320{,}000 Math traces, 83{,}200 Xwords traces, and 320{,}000 RHour traces. This longitudinal structure allows us to track how mid-trace behavior evolves during RL fine-tuning. We additionally produce 160{,}000 Qwen2.5-7B traces and 160{,}000 Llama3.1-8B traces for \textsc{MATH-500} across 10 checkpoints (Step~0 to Step~450 every 50 steps) to investigate behavior across architecture and model size. Details about our training and evaluation setup appear in App.~\ref{app:data-details}.

To identify reasoning shifts at scale, we use GPT-4o as an LLM-as-judge. Following evidence that rubric-prompted LLMs approximate human evaluation~\citep{zheng2023mtbench, liu2023geval, fu2023gptscore}, we supply a fixed rubric that scores each trajectory for (i) correctness, (ii) presence of a mid-trace reasoning shift, and (iii) whether the shift improved correctness.

To reduce known sources of judge bias---position, length, and model identity effects~\citep{Wang2024FairEval,shi2024positionbias,li-etal-2024-split}---we randomize sample order, use split--merge aggregation, enforce structured JSON outputs, and ensemble across prompt variants. We also adopt a conservative error-handling policy: we assign \textbf{no shift} when the cue prefilter fails or when the judge output is invalid or low-confidence (App.~\ref{app:detecting-aha}). Agreement is high: on \textsc{MATH-500}, GPT-4o achieves $\kappa\!\approx\!0.726$ across prompt variants and $\kappa=0.79$ relative to human majority vote, comparable to expert--expert reliability \citep{Artstein2008IAA}. For additional details, see App.~\ref{app:kappa-agreement}. For qualitative examples, see App.~\ref{app:qualitative-formal-aha}.

\subsection{Uncertainty Measure and Intervention}
\label{ss:unc}

To relate reasoning shifts to model uncertainty, we measure token-level entropy throughout each response. At generation step $t$, with next-token distribution $\mathbf{p}_t$, we compute Shannon entropy $H_t = -\sum_v p_t(v)\log p_t(v)$. For each completion, we summarize uncertainty by averaging entropy over the \tagtt{think} and \tagtt{answer} segments (e.g., $\bar H_{\text{think}}$ and $\bar H_{\text{ans}}$), and use these sequence-level scores in downstream analyses.

We also study whether uncertainty can be exploited to improve performance via \emph{artificially triggered} reflection. In the main runs we use the standard cue ``Wait, we need to reconsider. Let's think this through step by step.'', further details covering this setup are discussed in App. ~\ref{app:triggered-reconsideration}. In a follow-up experiment, we test three semantically similar but lexically distinct reconsideration cues (C1--C3), for example: \emph{(C3) ``Wait, something is not right, we need to reconsider. Let’s think this through step by step.''} For each cue, we first obtain the model's baseline completion (Pass~1), then re-query the model with the same decoding parameters while appending the reconsideration cue (Pass~2). Cue-specific results and regressions are reported in App.~\ref{app:uncertainty-interventions}. We evaluate gains both overall and under an entropy gate: we split instances into \emph{high-entropy} (top 20\% within domain) and \emph{low-entropy} (bottom 80\%) buckets based on Pass~1 sequence entropy, and compare Pass~2 accuracy across buckets.

\section{Results}

We show that spontaneous reasoning shifts are rare and generally harmful to accuracy, and that formal ``Aha'' events are vanishingly rare (\textbf{RQ1}; \S\ref{sec:results-rq1}); that this negative effect remains stable across training stages but varies systematically with decoding temperature (\textbf{RQ2}; \S\ref{sec:results-rq2}); and that extrinsically triggered shifts reliably improve performance, especially on high-entropy instances (\textbf{RQ3}; \S\ref{sec:rq3-uncertainty}).
\subsection{RQ1: Reasoning Shifts \& Model Accuracy}
\label{sec:results-rq1}
\noindent \textbf{Do reasoning shifts improve accuracy?}
Before analyzing formal ``Aha!'' moments, we first consider the broader class of \emph{reasoning shifts}---any mid-trace pivot detected by our annotator, irrespective of whether it satisfies the stricter criteria in Def.~\ref{def:aha-moment-lrms}. If such shifts reflected genuine insight, traces containing them should be \emph{more} accurate than those without them.

Across domains, temperatures, and checkpoints for Qwen2.5--1.5B, reasoning shifts remain uncommon (approximately $7.6\%$ of samples; pooling all models/domains yields 6.31\%) and are associated with substantially \emph{lower} accuracy: $2.57\%$ for shifted traces versus $16.44\%$ for non-shifted traces, $N{=}723{,}200$. A pooled logistic regression of correctness on a shift indicator confirms that this difference is highly significant (p $< 10^{-1198}$).\footnote{In R-style notation,
\(
\texttt{correct} \sim \texttt{shift}.
\)
\texttt{correct} is a binary outcome, and \texttt{shift} is a binary indicator for an annotator-labeled reasoning shift. The pooled regression aggregates all test-set traces across Crossword, Math, and RHour.}

To test whether this pattern is specific to our GRPO-tuned models, we evaluate DeepSeek\textendash R1 and GPT\textendash 4o under matched prompting and shift-detection conditions on \textsc{MATH\textendash 500}, sweeping decoding temperature $T\in\{0,0.05,0.3,0.7,1\}$. As shown in Table~\ref{tab:external-models-all}, \emph{canonical} shifts remain a low-base-rate event for both models across temperatures (DeepSeek\textendash R1: 0.00--0.25\%; GPT\textendash 4o: 0.88--2.21\%). Moreover, conditional accuracy given a detected shift does not show a consistent improvement: for GPT\textendash 4o, $P(\checkmark\mid S{=}1)$ is lower than $P(\checkmark\mid S{=}0)$ at every temperature, while for DeepSeek\textendash R1 the shift count is extremely small and the conditional estimates are correspondingly unstable. These results suggest that the low frequency and non-beneficial association of mid-trace shifts is not unique to our training setup, but also appears in external, high-capability reasoning models under the same operational definition. Moreover, these results characterize only the ``raw'' behavioral signature of mid-trace shifts, independent of any stricter ``Aha!'' interpretation.

\begin{table}[t]
\footnotesize
\setlength{\tabcolsep}{4pt}
\begin{tabular*}{\linewidth}{@{\extracolsep{\fill}} l l r r r @{}}
\toprule
{\textbf{Model}} &
{\textbf{Domain}} &
{\(\%{S_{i,j}}\)} &
{\scriptsize \(P(\checkmark\mid S_{i,j}\!=\!0)\)} &
{\scriptsize \(P(\checkmark\mid S_{i,j}\!=\!1)\)} \\
\midrule
\includegraphics[width=1em]{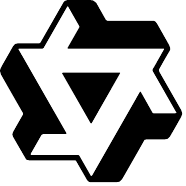}-1.5B
  & Xwords      & 1.22 & 0.096 & 0.201 \\
  & Math        & 2.65 & 0.327 & 0.144 \\
  & RHour       & 14.32 & 0.000 & 0.000 \\
\midrule
\includegraphics[width=1em]{icons/qwen.pdf}-7B
  & Math        & 1.50 & 0.661 & 0.282 \\
\midrule
\includegraphics[width=1em]{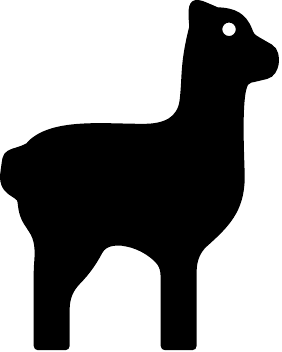}-8B
  & Math        & 5.04 & 0.457 & 0.282 \\
\midrule
\includegraphics[width=1em]{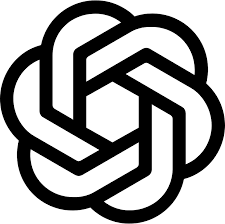}-4o
  & Math        & 2.51 & 0.658 & 0.338 \\
\midrule
\includegraphics[width=1em]{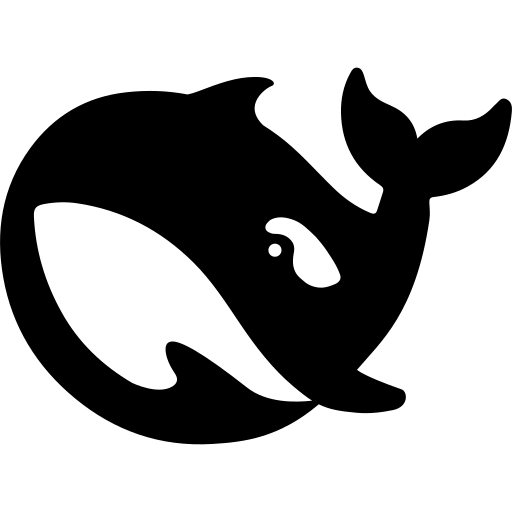}-R1
  & Math        & 4.96 & 0.543 & 0.204 \\
\bottomrule
\end{tabular*}
\caption{\textbf{Shift prevalence and conditional accuracy (RQ1).}
\(\%S_{i,j}\) gives the fraction of traces labeled as containing a reasoning shift.
\(P(\checkmark \mid S_{i,j}{=}0)\) and \(P(\checkmark \mid S_{i,j}{=}1)\) report accuracy without vs.\ with a detected shift, pooled across all problems, temperatures $\{0, 0.05, 0.3, 0.7\}$, checkpoints, and samples using count-weighted (not simple) averages.
Across models and domains, shifted traces are consistently less accurate.
\includegraphics[width=.7em]{icons/qwen.pdf} = Qwen\,2.5;
\includegraphics[width=.7em]{icons/llamma.pdf} = Llama\,3.1;
\includegraphics[width=.7em]{icons/openai.png} = GPT\,4o;
\includegraphics[width=.7em]{icons/deepseek.png} = DeepSeek\,R1.}
\label{tab:shift-accuracy}
\vspace{-3mm}
\end{table}

\begin{figure}[t]
  \centering
  \includegraphics[width=\linewidth]{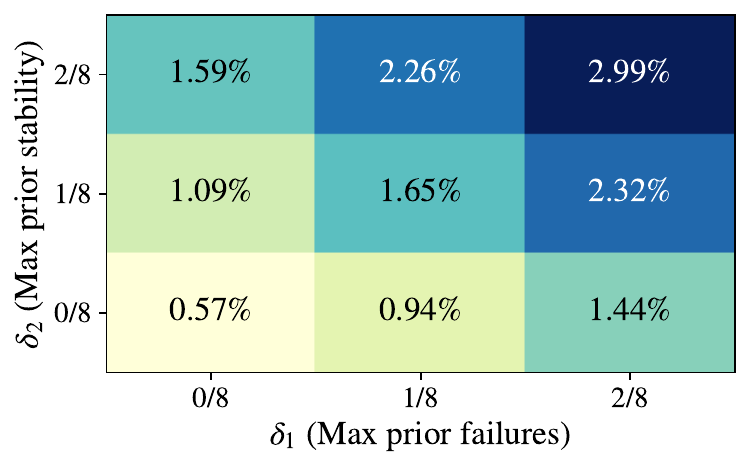}
  \caption{\textbf{Prevalence of formal ``Aha!'' events for Qwen2.5--1.5B (all domains, T{=}0.7).} Each cell shows the fraction (and count) of problem--checkpoint pairs
  \((q_j,k)\) that satisfy Def.~\ref{def:aha-moment-lrms} under varying
  thresholds for prior failures ($\delta_1$) and prior stability ($\delta_2$),
  with $\delta_3=\epsilon>0$. Even under lenient settings, formal ``Aha!''
  events are exceedingly rare. A guide to understanding heatmap calculations in more detail can be found in App.~\ref{app:aha-prevelance-descriptions}.
  See App.~\ref{sec:app-formal-aha-temp} for per-domain and per-temperature breakdowns.}
  \label{fig:aha-heatmap-overall}
  \vspace{-3mm}
\end{figure}

\vspace{1mm}
\noindent \textbf{How frequent are formal ``Aha!'' moments?}
We now restrict attention to the much smaller subset of events that satisfy \emph{all three} criteria in Def.~\ref{subsec:aha-moment-def}. In Fig.~\ref{fig:aha-heatmap-overall}, by varying $\delta_1,\delta_2\in\{0,\,1/8,\,2/8\}$ and fixing $\delta_3=\epsilon>0$, we find that formal ``Aha!'' moments are extremely rare, even with relatively lax constraints.
Similar patterns hold for Qwen2.5--7B and Llama3.1--8B (App.~\ref{sec:app-additional-results}). Pooling every Crossword/Math/RHour checkpoint and temperature, the formal detector fires on only $1.79\%$ of samples.

\vspace{1mm}
\noindent
\textbf{Robustness checks.}
As surface cues such as ``wait'' or ``actually'' often fail to track genuine strategy changes \citep{zheng-etal-2023-chain, xia2025reasoneval}, and LLM-judge labels may pick up prompt- or position-induced biases \citep{Wang2024FairEval,shi2024positionbias,li-etal-2024-split}, we replicate RQ1 using three detector variants (formal, GPT-based, lexical). All yield the same conclusion; see App.~\ref{app:shift-detector-rationale}.

\vspace{1mm}
\noindent
\textbf{Takeaway.}
Reasoning shifts are infrequent and generally harmful to accuracy. Further, \emph{formal} ``Aha!'' moments, which additionally require a performance gain at the pivot, are vanishingly rare. Neither the general phenomenon (reasoning shifts) nor its idealized form (``Aha!'' moments) appears to drive problem-solving performance of reasoning models.

\subsection{RQ2: Training Stage \& Temperature}
\label{sec:results-rq2}

RQ1 establishes two constraints on ``insight-like'' behavior: broad reasoning shifts are uncommon and tend to coincide with worse outcomes, while \emph{formal} ``Aha!'' events are so rare that they contribute little to overall model performance.
This raises a natural question: are we simply averaging over regimes where shifts sometimes help and sometimes hurt?
We test two plausible sources of heterogeneity:
(i) shifts might become more (or less) effective at different \emph{stages} of training; and
(ii) their impact might depend on the \emph{decoding temperature} (and thus sampling entropy).

\begin{figure}[t]
\centering

\begin{subfigure}[t]{\linewidth}
  \centering
  \includegraphics[width=\linewidth]{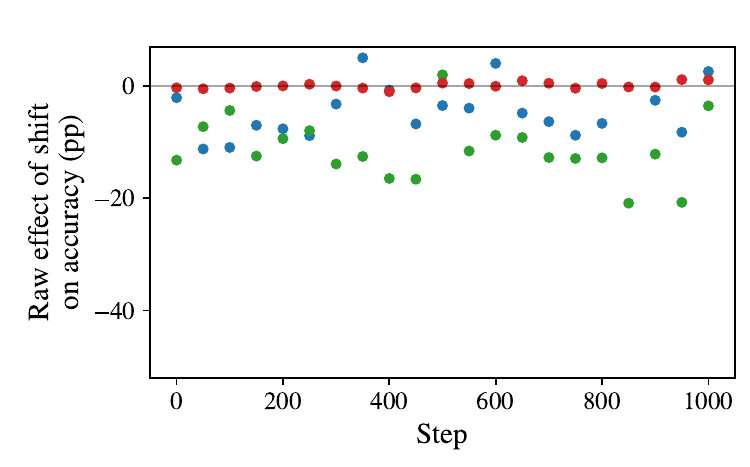}
  \caption{Raw effect of reasoning shifts over training for Qwen2.5-1.5B finetuning across domains (same evaluation at every step).}
  \label{fig:raw-effect-overlay:a}
\end{subfigure}

\vspace{4pt}

\begin{subfigure}[t]{\linewidth}
  \centering
  \includegraphics[width=\linewidth]{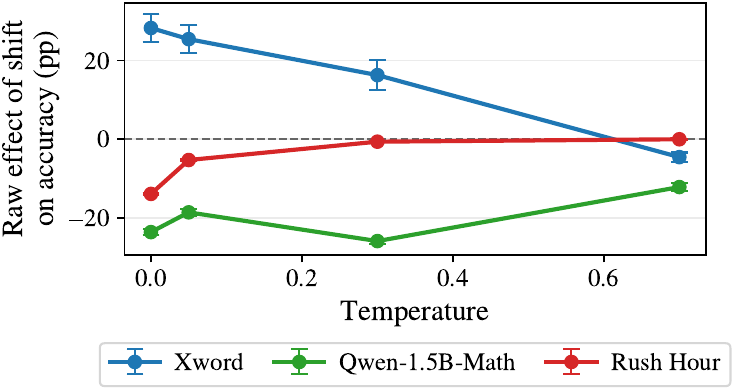}
  \caption{Raw effect of reasoning shifts over Qwen2.5-1.5B finetuning across domains (same evaluation at every temperature).}
  \label{fig:temp-raw-effect}
\end{subfigure}

\caption{\textbf{Reasoning shifts across training and temperature (Qwen2.5-1.5B).}
We plot the raw accuracy gap \(\widehat{\Delta}=\widehat{p}_{Y\mid S=1}-\widehat{p}_{Y\mid S=0}\) (pp).
(a)~At fixed \(T=0.7\), \(\widehat{\Delta}\) stays near zero or negative across training.
(b)~Across \(T\), shifts align with correction on \emph{Xword} at lower \(T\), remain harmful on \emph{Math}, and are near-zero on \emph{RHour}.}
\label{fig:raw-effect-overlay}
\vspace{-3mm}
\end{figure}

\vspace{2mm}
\noindent
\textbf{How does the effect of reasoning shifts vary across training?}
To test whether the shift--accuracy relationship changes as training progresses,
we regress correctness on problem fixed effects, standardized training step, and the shift indicator.
We report average marginal effects (AME) with cluster--robust SEs at the problem level.%
\footnote{In R-style notation:
\(
\texttt{correct} \sim \texttt{C(problem)} + \texttt{step\_std} + \texttt{shift}.
\)
\texttt{correct} is a binary outcome;
\texttt{C(problem)} are problem fixed effects;
\texttt{step\_std} is the standardized checkpoint index.}

At \(T{=}0.7\), we find no evidence that shifts become beneficial later in training.
In \emph{Xwords} and \emph{Math} shifts are uncommon (\(\%S{=}2.433\); \(\%S{=}2.166\)) and are mildly harmful (\(\mathrm{AME}{=}{-}0.0311\), \(p{=}0.02742\); \(\mathrm{AME}{=}{-}0.0615\), \(p{=}1.55\times10^{-4}\)).

In \emph{RHour}, shifts are comparatively frequent (\(\%S{=}11.449\)) but have no measurable practical effect on accuracy (\(\mathrm{AME}{\approx}0.0001\), \(p{\ll}10^{-6}\)).
Analogous results for \(T\in\{0.0,0.05,0.3\}\) are reported in Appendix~\ref{sec:app-rs-temp}.
Figure~\ref{fig:raw-effect-overlay}a echoes this pattern: across checkpoints, shifted traces are not systematically more accurate than non-shifted ones.
We repeat robustness checks using alternative detector variants across \(T\in\{0,0.05,0.3,0.7\}\) in App.~\ref{spp:ablations}.
We observe the same qualitative pattern with the stricter \emph{formal} ``Aha!'' detector (Appendix~\ref{sec:app-formal-aha-temp}), but because it fires on only \(\approx 10^{-3}\) of traces at \(T{=}0.7\), estimates are underpowered for fine-grained stage-by-stage heterogeneity; critically, we do not see a consistent late-training transition to positive effects.

\vspace{1mm}
\noindent
\textbf{How does the effect of reasoning shifts vary with decoding temperature?}
We next ask whether temperature modulates the relationship between shifts and correctness.
We regress correctness on problem fixed effects, standardized temperature, and the shift indicator, aggregating across training steps.%
\footnote{R-style notation:
\(
\texttt{correct} \sim \texttt{C(problem)} + \texttt{temp\_std} + \texttt{shift}.
\)
\texttt{temp\_std} is the standardized decoding temperature.}

Table~\ref{tab:rs} summarizes the average association between shifts and correctness while controlling for standardized decoding temperature (via \texttt{temp\_std}).
Figure~\ref{fig:temp-raw-effect} shows the corresponding per-\(T\) raw pattern.
On \emph{Xwords}, the coefficient is positive but not statistically distinguishable from zero (\(\mathrm{AME}{=}0.0326\), \(p{=}0.2595\)), despite a positive raw contrast \(\Delta{=}{+}10.54\)pp.
On \emph{Math}, shifts are strongly harmful (\(\mathrm{AME}{=}{-}0.0831\), \(p{=}2.68\times10^{-8}\); \(\Delta{=}{-}18.35\)pp).
On \emph{RHour}, shifts are frequent (\(\%S{=}14.32\)) but correctness is extremely low overall; accordingly, the estimated effect is statistically detectable yet numerically negligible (\(\mathrm{AME}{\approx}{-}0.0003\), \(p{=}2.72\times10^{-7}\); \(\Delta{\approx}{-}0.02\)pp).

Raw per-temperature contrasts (Fig.~\ref{fig:temp-raw-effect}) sharpen the interpretation:
on \emph{Xwords}, shifts can coincide with productive correction at low \(T\), but the benefit weakens and may reverse by \(T{=}0.7\).
In \emph{Math}, shifts remain harmful across temperatures, though the raw penalty attenuates as \(T\) increases.
In \emph{RHour}, the curve stays close to zero in magnitude, reflecting the near-zero accuracy regime.

\vspace{1mm}
\noindent
\textbf{Takeaway.}
We find that reasoning shifts do not reliably yield higher accuracy across specific training phases or at particular temperatures.

\begin{table}[t]
\centering
\small
\begin{tabular}{lrrr}
\toprule
\multicolumn{4}{c}{\textbf{(a) Training stage}} \\
\midrule
\textbf{Metric} & \textbf{Xword} & \textbf{Math} & \textbf{RHour} \\
\midrule
$N$                   & 20{,}800 & 80{,}000 & 80{,}000 \\
$\%S$                 & 2.433    & 2.166    & 11.449  \\
$\hat{p}_{Y\mid S=1}$ & 0.0731   & 0.1691   & 0.0001  \\
$\Delta\mathrm{pp}$   & $-4.52$  & $-11.83$ & $+0.00$ \\
$\mathrm{AME}$        & $-0.0311$& $-0.0615$& $0.0001$ \\
$p$                   & $0.02742$ & $1.55\times10^{-4}$ & $\ll 10^{-6}$ \\
\midrule
\multicolumn{4}{c}{ \textbf{(b) Temperature}} \\
\midrule
\textbf{Metric} & \textbf{Xword} & \textbf{Math} & \textbf{RHour} \\
\midrule
$N$                       & 83{,}200  & 320{,}000 & 320{,}000 \\
\(\%S\)                   & 1.220     & 2.646     & 14.318 \\
$\hat{p}_{Y\mid S=1}$     & 0.2010    & 0.1435    & 0.0000 \\
$\Delta\mathrm{pp}$       & $+10.54$  & $-18.35$  & $-0.02$ \\
$\mathrm{AME}$            & 0.0326    & $-0.0831$ & $-0.0003$ \\
$p$                       & $0.2595$ & $2.68\times10^{-8}$ & $2.72\times10^{-7}$ \\
\bottomrule
\end{tabular}
\caption{\textbf{Effect of detected reasoning shifts on accuracy (Qwen2.5-1.5B).}
For each domain, \(\%S\) is the share of samples where the LLM-as-judge detects a shift (\(S_{i,j}=1\)); \(\hat{p}_{Y\mid S=1}\) is the empirical accuracy among shifted traces; and \(\Delta\mathrm{pp}\) is the raw accuracy difference (in percentage points) between shifted and non-shifted traces.
\emph{(a)} controls for training step (standardized) at fixed training decoding temperature \(T=0.7\); \emph{(b)} controls for decoding temperature \(T\in\{0.0,0.05,0.3,0.7\}\) while aggregating over steps.
\(\mathrm{AME}\) is the average marginal effect of a shift from a logistic regression with problem fixed effects and cluster-robust SEs; negative values mean that, holding problem and step/temperature fixed, traces with shifts are less likely to be correct.
See \S\ref{sec:results-rq2} for the full regression specification.}
\label{tab:rs}
\vspace{-5mm}
\end{table}

\subsection{RQ3: Reasoning Shifts \& Uncertainty}
\label{sec:rq3-uncertainty}

The results above (particularly \textit{Xwords}, see Fig.~\ref{fig:temp-raw-effect}) suggest that decoding temperature may modulate the effect of reasoning shifts: at low $T$ they sometimes align with productive corrections, while at higher $T$ they resemble noise. Because temperature primarily alters sampling entropy rather than the model's underlying reasoning process~\citep{hinton2015distilling,holtzman2019degeneration}, this points to a link between shifts and internal uncertainty. We thus ask whether, under high-uncertainty regimens, reasoning shifts are more frequent or become more helpful.

\vspace{1mm}
\noindent
\textbf{Are reasoning shifts more likely under high uncertainty?}
To test whether shifts preferentially occur when the model is uncertain, we relate each trace's reasoning shift indicator to its sequence-level entropy. We pool traces across all decoding temperatures and training checkpoints, and fit a logistic regression of shift prevalence on standardized entropy with problem fixed effects and cluster-robust SEs (clustered by problem).%
\footnote{In R-style notation:
\(
\texttt{shift} \sim \texttt{C(problem)} + \texttt{std\_entropy}.
\)
Here \texttt{shift} is a binary indicator for a reasoning shift, \texttt{C(problem)} denotes problem fixed effects, and \texttt{std\_entropy} is the within-domain $z$-scored pass-1 sequence entropy. We estimate a Binomial(logit) GLM with cluster-robust SEs at the problem level.}

Pooling all traces across domains (\textit{Xwords}, \textit{Math}, \textit{RHour}), we find weak evidence that higher entropy is associated with \emph{fewer} detected shifts on average (OR$\approx 0.77\times$, $\beta=-0.258$, $\mathrm{SE}=0.143$, $p=0.070$; 95\% CI OR $\in[0.58,1.02]$; $N=723{,}200$). This aggregate pattern masks domain heterogeneity: the entropy--shift association is positive in \textit{Xwords} (OR$\approx 2.05\times$) and \textit{RHour} (OR$\approx 1.19\times$), but negative in \textit{Math} (OR$\approx 0.58\times$). One possible interpretation is that in \textit{Math}, high-entropy generations more often reflect diffuse exploration or verbose ``flailing'' rather than a discrete mid-trace pivot, so the rare, rubric-qualified shifts concentrate in comparatively lower-entropy traces.

\vspace{1.5mm}
\noindent
\textbf{Do reasoning shifts improve performance under high uncertainty?}
A natural hypothesis is that when the model is uncertain, a mid-trace pivot might reflect productive self-correction. We test this by stratifying traces into \emph{high-entropy} instances (top 20\% within domain) and \emph{low-entropy} instances (bottom 80\%), using a fixed entropy threshold per domain. Within each stratum, we estimate the effect of a shift on correctness using a logistic regression with problem fixed effects and controls for continuous entropy and training step, and report the shift coefficient alongside the raw accuracy difference between shifted and non-shifted traces.%
\footnote{Within each domain, we split at the 80th percentile of sequence entropy and fit a Binomial(logit) GLM predicting \texttt{correct} from \texttt{shift} with problem fixed effects and covariates. We report both regression and raw contrasts for interpretability.}

Table~\ref{tab:rq3-shift-entropy-strata} shows that shifts do \emph{not} become reliably beneficial in the high-entropy regime. In \emph{Math}, shifts remain harmful even among high-entropy traces (raw $\Delta=-7.40$pp) and are substantially more harmful in the low-entropy majority (raw $\Delta=-22.88$pp). In \emph{Xwords}, the point estimate in the high-entropy stratum is near zero (raw $\Delta=+0.63$pp), but shifts are rare and estimates are noisy. In \emph{RHour}, accuracy is near-zero throughout, so estimated effects are statistically detectable due to sample size but negligible in magnitude.

\begin{table}[t]
  \centering
  \small
  \setlength{\tabcolsep}{4pt}
  \renewcommand{\arraystretch}{1.05}
  \begin{tabular*}{\columnwidth}{@{\extracolsep{\fill}} l r r r @{}}
    \toprule
    \textbf{Metric} & \textbf{Xword} & \textbf{Math} & \textbf{RHour} \\
    \midrule
    \multicolumn{4}{c}{\textbf{All traces}} \\
    \midrule
    $N$                       & 83{,}200  & 320{,}000 & 320{,}000 \\
    $\Delta$ (pp)             & $-6.24$   & $-19.78$  & $-0.02$   \\
    $\mathrm{coef(shift)}$    & $-1.49$   & $-1.11$   & $-22.76$  \\
    $p$                       & $0.123$   & $2.25\times10^{-7}$ & $\approx 0$ \\
    \midrule
    \multicolumn{4}{c}{\textbf{High entropy (top 20\%)}} \\
    \midrule
    $N$                       & 16{,}640  & 64{,}000  & 64{,}000 \\
    $\Delta$ (pp)             & $+0.63$   & $-7.40$   & $-0.03$  \\
    $\mathrm{coef(shift)}$    & $-0.04$   & $-0.28$   & $-22.48$ \\
    $p$                       & $0.904$   & $0.739$   & $\approx 0$ \\
    \midrule
    \multicolumn{4}{c}{\textbf{Low entropy (bottom 80\%)}} \\
    \midrule
    $N$                       & 66{,}560  & 256{,}000 & 256{,}000 \\
    $\Delta$ (pp)             & $-10.00$  & $-22.88$  & $-0.02$   \\
    $\mathrm{coef(shift)}$    & $-28.83$  & $-1.14$   & $-22.90$  \\
    $p$                       & $1.33\times10^{-46249}$ & $4.96\times10^{-7}$ & $\approx 0$ \\
    \bottomrule
  \end{tabular*}
  \caption{\textbf{Do spontaneous reasoning shifts help under high uncertainty?}
  We stratify traces within each domain by sequence entropy (high = top 20\% at the within-domain 80th percentile; low = bottom 80\%), and compare shifted vs.\ non-shifted traces. $\Delta$ (pp) is the raw accuracy difference
  $\hat p(\checkmark\!\mid\!S{=}1) - \hat p(\checkmark\!\mid\!S{=}0)$. $\mathrm{coef(shift)}$ and $p$ report the shift coefficient and $p$-value from a logistic regression with problem fixed effects and covariates. Across domains, shifts do not become reliably beneficial in the high-entropy regime.}
  \label{tab:rq3-shift-entropy-strata}
\end{table}

\vspace{1mm}
\noindent
\textbf{Can artificially triggered reasoning shifts improve performance?}
The negative results above suggest that \emph{spontaneous} shifts are not a dependable self-correction mechanism, even when the model is uncertain. High entropy does not cause more spontaneous pivots; rather, it identifies instances where a second-pass reconsideration has higher marginal value. We therefore test an \emph{extrinsically triggered} ``forced Aha'' intervention: for each prompt we generate a baseline completion (Pass~1), then re-query the model under identical decoding settings while appending a reconsideration cue (Pass~2), and compare paired correctness outcomes. Pass~2 uses the same cue across all domains; see App.~\ref{app:uncertainty-interventions} for the exact wording and additional analyses.

Table~\ref{tab:rq3-forced-aha} reports paired results aggregated across checkpoints and decoding temperatures. Triggered reconsideration yields a large gain on \emph{Math} ($0.322\!\rightarrow\!0.406$; $+8.41$pp) and a small gain on \emph{Xwords} ($+0.45$pp), while remaining negligible in absolute terms on \emph{RHour} ($+0.01$pp) due to its near-zero base rate. The paired ``win'' counts show that improvements dominate backslides in \emph{Math} (50{,}574 wrong$\rightarrow$right vs.\ 23{,}500 right$\rightarrow$wrong), indicating that the effect is not merely random flipping. In contrast, \emph{Xwords} shows near-balanced wins and losses (5{,}380 vs.\ 5{,}004), consistent with a much smaller net gain.

Finally, consistent with uncertainty serving as a useful gate for reflection, Appendix~\ref{app:uncertainty-interventions} shows that these gains are amplified on high-entropy instances (Table~\ref{tab:pass2-entropy-gate}), with a complementary regression analysis reported in Table~\ref{tab:pass2-entropy-regression}.

\begin{table}[t]
  \centering
  \small
  \setlength{\tabcolsep}{4pt}
  \renewcommand{\arraystretch}{1.05}
  \begin{tabular*}{\columnwidth}{@{\extracolsep{\fill}} l r r r r r r @{}}
    \toprule
    \textbf{Metric} & \textbf{Xword} & \textbf{Math} & \textbf{RHour} \\
    \midrule
    $N$ (paired samples)      & 83{,}200  & 320{,}000  & 320{,}000 \\
    $\hat p_{\text{P1}}$      & 0.0970    & 0.3221    & 0.000233 \\
    $\hat p_{\text{P2}}$      & 0.1015    & 0.4062    & 0.000363 \\
    $\Delta$ (pp)             & $+0.45$   & $+8.41$   & $+0.01$ \\
    wins (P2 $\uparrow$)      & 5{,}380   & 50{,}574  & 100 \\
    wins (P1 $\uparrow$)      & 5{,}004   & 23{,}500  & 58 \\
    \bottomrule
  \end{tabular*}
  \caption{\textbf{Forced ``Aha'' (triggered reconsideration), sample-level results.}
  We compare paired outcomes between a baseline generation (Pass~1) and a second generation with an appended reconsideration cue (Pass~2).
  $\hat p_{\text{P1}}$ and $\hat p_{\text{P2}}$ denote accuracies in each pass; $\Delta$ (pp) is the percentage-point gain.}
  \label{tab:rq3-forced-aha}
  \vspace{-5mm}
\end{table}

\vspace{1mm}
\noindent
\textbf{Takeaway.} Reasoning shifts are a low-base-rate event whose association with entropy varies by domain, and conditioning on uncertainty does not reveal a ``hidden regime'' where spontaneous shifts reliably help. In contrast, artificially triggering reconsideration yields consistent gains, especially for \emph{Math} and especially in the high-entropy tail (App.~\ref{app:pass2-entropy}, Table~\ref{tab:pass2-entropy-gate}).

\section{Discussion and Future Work}
\label{Discussion}

We formalize and empirically test the notion of intrinsic ``Aha!'' moments, mid-trace reasoning shifts that appear to reflect sudden insight. We find that they are vanishingly rare and that mid-trace reasoning shifts are typically unhelpful, even in states of high uncertainty. However, by intervening to trigger reconsideration under high-entropy conditions, we demonstrate that uncertainty can be converted into productive reflection, resulting in measurable accuracy gains.

This reframes reasoning shifts not as an emergent cognitive ability, but as a mechanistic behavior---a byproduct of the model's inference dynamics that can nonetheless be harnessed and controlled.
Rather than asking whether models have insight, it may be more useful to ask how and when they can be made to simulate it. This shift in perspective bridges recent work on uncertainty-aware decoding~\cite{ton2025infotheory, zhou2025safekey}, process supervision~\cite{uesato2022solving, openai2023processsupervision}, and self-correction~\citep{madaan2023selfrefine, kumar2024traininglanguagemodelsselfcorrect, tsui2025selfcorrectionbenchrevealingaddressing}, positioning mid-trace reasoning as a manipulable mechanism for improving reliability rather than genuine insight.

Our findings open several directions for further investigation.
First, the link we uncover between uncertainty and the usefulness of mid-reasoning shifts invites new forms of process-level supervision that explicitly condition reflection on entropy or confidence estimates \citep{uesato2022solving, openai2023processsupervision}.
Second, future work should examine whether RL-based objectives that reward models for revising earlier answers truly improve reasoning or merely reinforce uncertainty-responsive heuristics.
While recent approaches such as \citet{kumar2024traininglanguagemodelsselfcorrect} demonstrate that self-correction can be trained, our results highlight the need for  analyses that disentangle the learning of reflection-like language from genuine representational changes. It would be valuable to investigate what the observed dynamics between uncertainty and mid-reasoning shifts reveal about human insight—whether uncertainty-driven reconsideration in models mirrors metacognitive signals in people, or whether the resemblance is purely linguistic. Bridging computational and cognitive accounts of ``Aha!'' phenomena could help identify which internal mechanisms, if any, correspond to genuine insight. Finally, we hope that this piece inspires more fundamental research into the impact of RL post-training on model performance: why do algorithms like GRPO lead to a performance shift if not from improved reasoning?

\section{Limitations}
While our study offers the first systematic analysis of “Aha!” phenomena in reasoning models, it has several limitations. 
First, our detection of reasoning shifts relies on \textit{explicit linguistic cues} (e.g., “wait,” “actually”) and measurable plan changes. This makes our estimates conservative: models may undergo unlexicalized representational changes that our detector misses, while some detected shifts may instead reflect superficial hedges. Future work could incorporate hidden-state dynamics or token-level embeddings to better identify implicit restructurings. 
Second, our evaluation spans three reasoning domains but remains limited to tasks with well-defined correctness signals (math, Xwords, spatial puzzles). Whether similar patterns hold for open-ended reasoning or multi-turn interaction remains an open question. 
Third, our intervention experiments manipulate model behavior via prompt-level cues rather than modifying training objectives. Thus, while we demonstrate that uncertainty-gated reconsideration can improve accuracy, this does not establish a causal mechanism of internal insight. Extending our analysis to training-time interventions or process supervision would help clarify how reflection-like behaviors emerge and generalize.
Finally, as with most large-model studies, our results depend on a small set of families (Qwen, Llama) and inference hyperparameters (e.g., temperature, sampling policy). Broader replications across architectures, decoding methods, larger sizes, and reinforcement-learning setups are necessary to test the generality of our conclusions.

\section{Ethical Considerations}

Our study analyzes the internal reasoning behavior of large language models and does not involve human subjects or personally identifiable data. 
All datasets used---\textsc{MATH-500}~\citep{lightman2024letsverifystepstep}, \textsc{Cryptonite}~\citep{efrat-etal-2021-cryptonite}, and synthetic \textsc{RHour} puzzles---are publicly available and contain no sensitive content. 
We follow the terms of use for each dataset and model.

Because our work involves interpreting and modifying model reasoning traces, it carries two potential ethical implications. 
First, methods that manipulate mid-trace behavior could be misused to steer reasoning models toward undesirable or deceptive outputs if deployed irresponsibly. 
Our interventions are limited to controlled research settings and designed to study model uncertainty, not to conceal reasoning or produce persuasive content. 
Second, interpretability claims about “insight’’ or “self-correction’’ risk overstating model understanding. 
We therefore emphasize that our findings concern statistical behavior, not human-like cognition or consciousness.

Generative AI tools were used to enhance the search for related works and to refine the writing and formatting of this manuscript. 
We followed the recommendations of~\citet{schroeder2025llmsqual}, who provide guidance for legitimate uses of AI in research while safeguarding qualitative sensemaking. 
Specifically, Claude, ChatGPT, and Elicit were used to identify relevant research papers for the Related Work and Discussion sections (alongside non-generative tools such as Google Scholar and Zotero). 
After the Discussion had been written, ChatGPT was used to streamline and refine the prose, which was then manually edited by the authors. 
Claude and ChatGPT were additionally used for formatting tasks, such as generating table templates and translating supplementary materials to \LaTeX. 
Where generative AI was used, the authors certify that they have reviewed, adapted, and corrected all text and stand fully behind the final content.

All model runs, including training and inference, were conducted on NVIDIA A100 GPUs or NVIDIA A6000 GPUs, with resource management, access controls, and energy considerations in place. We estimate the total carbon footprint of all experiments at approximately~110~kg~CO\textsubscript{2}e, following the methodology of~\citet{luccioni2019carbon}.

\section*{Acknowledgments}
This work was supported by a First-Year Fellowship from the Princeton University Graduate School. We are grateful for computational resources provided by the Beowulf cluster, and we thank the ML Theory Group for generously sharing additional compute. We also acknowledge the support of the Center for Information Technology Policy (CITP) and the Department of Computer Science at Princeton University. We thank our volunteer annotators and the broader Princeton community. Special thanks to Cannoli, our lab dog, and to the musical artist, Doechii.

\newpage

\bibliography{curated_bib}

\newpage

\appendix

\section{Experimental Setup and Data}
\label{sec:appendix}

This first part of the appendix collects the \emph{reproducibility scaffolding} for our experiments: what data we train and evaluate on, what prompts and output contracts we impose, and how GRPO training is configured. 
We begin with dataset sizes and domain-specific preprocessing details (\S\ref{app:data-details}). 
We then provide the verbatim system-level prompts used in each domain, including the shared \tagtt{think}/\tagtt{answer} formatting requirements and domain-specific guardrails (\S\ref{app:system-level-prompts}). 
Finally, we summarize the GRPO training setup and per-domain hyperparameters (\S\ref{app:grpo-setup}; App.~A.3). 
All of these components are held fixed across checkpoints unless explicitly noted, so that differences reported in the main text reflect changes in model state rather than instruction drift or evaluation artifacts.

\subsection{Dataset Details}
\label{app:data-details}

Table~\ref{tab:data-stats} summarizes dataset sizes, splits, and evaluation coverage for all three reasoning domains. We include additional details here for reproducibility.

\paragraph{Cryptic Xwords.}
We use the \textsc{Cryptonite} corpus for training \citep{efrat-etal-2021-cryptonite} and generate synthetic evaluation clues using device-balanced templates. All answers are normalized (uppercase, punctuation-stripped) before exact-match scoring.

\paragraph{Math.}
The training distribution is \texttt{openR1 Math-220k} \citep{openr1}; evaluation is on the \textsc{MATH-500} benchmark \citep{lightman2024letsverifystepstep}. Normalization removes \LaTeX\ wrappers, whitespace, and trivial formatting differences (e.g., `$1/2$' vs.\ `$\frac{1}{2}$') before exact match.

\paragraph{RHour.}
We generate balanced $4{\times}4$, $5{\times}5$, and $6{\times}6$ puzzles and evaluate on $6{\times}6$ only. Puzzles are solved optimally via BFS with per-size node caps; unsolved or degenerate boards are removed \citep{fogleman2018rushhour}. Solutions are canonicalized before comparison.

\paragraph{Data release.}
Code to regenerate the synthetic Cryptic Xwords evaluation set and the selected RHour puzzles is included in our repository under \texttt{data/}. We also release the exact evaluation subsets on Hugging Face: \texttt{od2961/rush4-5-6-balanced} and \texttt{od2961/Guardian-cryptonite-official-split}.

\subsection{System-Level Prompts}
\label{app:system-level-prompts}

\begin{figure*}[t]
  \centering
  \begin{promptbox}[width=\textwidth]{System Prompt — Cryptic Xwords}
  You are an expert cryptic-xword solver. Do this (repeat until fully consistent):
  A) DEVICE TRIAGE
    • List plausible devices from {anagram, container, reversal, hidden, charade, deletion, homophone, double def, &lit, letter selection, substitution, abbr}.
    • For each, quote the indicator word(s). Reject with a reason.
  B) PARSE
    • Mark the **definition** (start or end).
    • Mark the **wordplay** with exact fodder + operations.
  C) CHECKS
    • Enumeration must match exactly.
    • Letter accounting must be exact (anagram multiset or stepwise build).
  D) DECIDE
    • Pick the candidate best matching definition, indicator(s), and enumeration.
    • Do NOT assume anagram without a clear anagrind and fully used fodder.
  E) RECONSIDER (if any check fails)
    • Begin the next \tagtt{think} with: "Wait, we need to reconsider. Let's think this through step by step."
    • Say why it failed, then re-run A–D with an alternative device/parse.
  FORMAT (no deviations):
    • Reasoning only in \tagtt{think}$\cdots$\etagtt{think}
    • Final entry ONLY (UPPER-CASE) in \tagtt{answer} $\cdots$ \etagtt{answer}
  Clue: Close, as seen in plaNET EARly (4)
  \tagtt{think}Device: HIDDEN; indicator "as seen in".
  Def: "Close". Wordplay: hidden in "plaNET EARly" -> NEAR.
  Enumeration: (4) OK.\etagtt{think}
  \tagtt{answer} NEAR \etagtt{answer}
  Clue: Mix TEA for a hot drink (3)
  \tagtt{think}Device: ANAGRAM; indicator "Mix". Fodder TEA -> TEA.
  Def: "a hot drink". Accounting exact; (3) OK.\etagtt{think}
  \tagtt{answer} TEA \etagtt{answer}
  Clue: Shoe liner at home on fish (6)
  \tagtt{think}Device triage: {hidden ? ("on" is not a hidden indicator), anagram ✗ (no anagrind),
  charade ✓ ("at home"=IN, "on"=next to), homophone ✗, ...}
  Attempt (HIDDEN) rejected: no indicator; also hidden spans don't give (6).
  Candidate attempt (wrong path): — fails enumeration/indicator, so we must rethink.
  Re-evaluate as CHARADES: IN ("at home") + SOLE ("fish") -> INSOLE.
  Accounting: INSOLE letters: I N S O L E (6). Definition "Shoe liner" fits. Enumeration (6) OK.\etagtt{think}
  \tagtt{answer}INSOLE\etagtt{answer}
  \end{promptbox}
  \caption{\textbf{System Prompt — Cryptic Xword.}
  Verbatim system-level prompt used for the Xwords domain.}
  \label{fig:prompt-xword}
\vspace{-1mm}
\end{figure*}

\paragraph{Design goals.}
Our system prompts serve two purposes: (i) \emph{scaffold} domain-appropriate reasoning with verifiable intermediate structure, and (ii) \emph{standardize} outputs so they are machine-checkable and comparable across checkpoints. Across all domains we therefore (a) separate private reasoning from the final response with explicit tags (\tagtt{think}\,/\,\etagtt{think} and \tagtt{answer}\,/\,\etagtt{answer}), (b) enforce domain-specific \emph{guardrails} (e.g., enumeration and letter accounting for cryptics; canonical forms for mathematics; regex-constrained action sequences for RHour), and (c) build in a light-weight \emph{self-correction loop} that triggers targeted reconsideration when a check fails. The prompts below were held fixed across checkpoints and temperatures (unless noted), ensuring that any changes we observe arise from the model state rather than instruction drift.

\begin{figure*}[t]
  \centering
  \begin{promptbox}[width=\textwidth]{System Prompt — Math}
You are an expert *mathematics problem-solver.
  Every time you receive a problem you must:
  • Analyze it thoroughly.
    – Pinpoint the **goal** (what quantity/set/form is requested).
    – Pinpoint the **givens/constraints** (domains, integrality, non-negativity, geometric conditions).
    – Choose the **methods** to apply (algebraic manipulation, factorization, inequalities, counting, modular arithmetic, geometry, calculus, etc.).
    – Write out the full derivation that leads to the final result.
  • Check that the result satisfies all original constraints (no extraneous roots, correct domain, simplified form, exact arithmetic).
  • Respond in **exactly** the tag-based format shown below – no greeting, no commentary outside the tags.
    – The final answer goes inside \tagtt{answer} **only**.
    – Use **exact** math (fractions, radicals, π, e). Avoid unnecessary decimals.
    – Canonical forms: integers as plain numbers; reduced fractions a/b with b>0; simplified radicals; rationalized denominators; sets/tuples with standard notation; intervals in standard notation.
    – If there is **no solution**, write NO SOLUTION. If the problem is **underdetermined**, write I DON'T KNOW.
  • You have a hard cap of **750 output tokens**. Be concise but complete.
  TAG TEMPLATE (copy this shape for every problem)
  \tagtt{think}
  YOUR reasoning process goes here:
  1. quote the relevant bits of the problem
  2. name the mathematical tool(s) you apply
  3. show each intermediate step until the result is reached
  If you spot an error or an unmet constraint, iterate, repeating steps 1–3 as many
  times as necessary until you are confident in your result. Finish by verifying the
  result satisfies the original conditions exactly (substitution/checks).
  \etagtt{think}
  \tagtt{answer}
  THEANSWER
  \etagtt{answer}
  \end{promptbox}
  \caption{\textbf{System Prompt — Math.}
  Verbatim system-level prompt used for math.}
  \label{fig:prompt-math}
\end{figure*}

\paragraph{Common scaffolding (all domains).}
We ask models to reason entirely inside \tagtt{think} and to place the \emph{final} object to be graded inside \tagtt{answer} \emph{only}. Tag separation lets us (1) compute reasoning-shift features on the private trace without leaking them into the final output, and (2) apply exact validators to \tagtt{answer}. To avoid verbosity that can mask errors, prompts specify concise but complete derivations, a token budget, and deterministic formatting. 

\begin{table}[t]
  \centering
  \small
  \setlength{\tabcolsep}{6pt}
  \renewcommand{\arraystretch}{1.05}
  \begin{tabular*}{0.92\linewidth}{@{\extracolsep{\fill}} l r r @{}}
    \toprule
    \textbf{Domain} & \textbf{Train ($N$)} & \textbf{Eval ($N$)} \\
    \midrule
    Cryptic Xwords & 50{,}000 & 130 \\
    Math          & 220{,}000 & 500 \\
    RHour         & 180{,}000 & 500 \\
    \bottomrule
  \end{tabular*}
  \caption{\textbf{Dataset sizes.} Training instances are natural clues (Xwords), problems (Math), and boards (RHour); evaluation uses synthetic clues for Xwords.}
  \vspace{-5mm}
  \label{tab:data-stats}
\end{table}

\paragraph{Cryptic Xwords.}
The Xwords prompt encodes established solving practice: \emph{device triage} (anagram, container, reversal, hidden, \emph{etc.}) with quoted indicators, followed by a two-part parse (\textbf{definition} and \textbf{wordplay}) and two hard checks: exact enumeration and exact letter accounting. This combination suppresses common failure modes such as defaulting to anagrams without a bona fide anagrind or silently dropping letters in charades. The reconsideration loop is narrow: it requires explaining \emph{why} the current attempt fails before proposing an alternative device/parse. We found this prevents thrashing while still eliciting genuine mid-trace pivots when a better device is available. Examples in the prompt illustrate (i) a hidden, (ii) an anagram, and (iii) a charade—covering the most frequent device classes in our corpus.

\paragraph{Math.}
The math prompt stresses (i) \emph{goal/givens/methods} triage, (ii) exact, symbolic manipulation with canonical forms (fractions, radicals, \(\pi\), \(e\)), and (iii) end-of-proof \emph{validation} (domain, extraneous roots, simplification). We explicitly specify what to output when a problem is infeasible (``\texttt{NO SOLUTION}'') or underdetermined (``\texttt{I DON'T KNOW}''), which reduces hallucinated specificity. The tag split is enforced more strictly here to prevent the final answer from appearing in \tagtt{think} and to keep \tagtt{answer} parsable for grading and correctness metrics. The 750-token cap preserves headroom for multi-step derivations while discouraging digressions that add entropy without improving validity.

\begin{figure*}[t]
  \centering
  \begin{promptbox}[width=\textwidth]{System Prompt — RHour}
  You are an expert RHour ({N}×{N}) solver.
  TASK
  • Input fields are provided in the user message:
    - Board (row-major string with 'o', 'A', 'B'..'Z', optional 'x')
    - Board size (e.g., 4x4 or 5x5 or 6x6)
    - Minimal moves to solve (ground-truth optimum), shown for training
  • OUTPUT exactly ONE optimal solution as a comma-separated list of moves.
    - Move token = <PIECE><DIR><STEPS> (e.g., A>2,B<1,Cv3)
    - Directions: '<' left, '>' right, '^' up, 'v' down
    - No spaces, no prose, no extra punctuation or lines.
  GOAL
  • The right end of 'A' must reach the right edge.
  OPTIMALITY & TIE-BREAKS
  • Your list must have the minimal possible number of moves.
  • If multiple optimal sequences exist, return the lexicographically smallest
    comma-separated sequence (ASCII order) after normalizing tokens.
  VALIDATION
  • Tokens must match: ^[A-Z][<>^v]\d+(,[A-Z][<>^v]\d+)*\$
  • Each move respects vehicle axes and avoids overlaps with walls/pieces.
  • Applying the full sequence reaches the goal with exactly {moves} moves.
  IF INCORRECT / UNVALIDATED
  • Repeat reasoning process, iterating until correct.
  FORMAT
  • Answer in the following way:
  \tagtt{think}
  Your reasoning
  \etagtt{think}
  \tagtt{answer}
  A>2,B<1,Cv3
  \etagtt{answer}
  \end{promptbox}
  \vspace{-3mm}
  \caption{\textbf{System Prompt — RHour.}
  Verbatim system-level prompt used for RHour puzzles.}
  \label{fig:prompt-rush}
\vspace{-1mm}

\end{figure*}

\paragraph{RHour.}
For RHour puzzles, the prompt formalizes the interface between natural-language reasoning and a discrete planner. Inputs are normalized (\(N\times N\) board, row-major encoding), and \tagtt{answer} must match a regular expression of move tokens (\verb|^[A-Z][<>^v]\d+(,[A-Z][<>^v]\d+)*$|). We add two verifiability clauses: (i) the sequence must be \emph{optimal} (minimum length), with lexicographic tie-breaks to canonicalize multiple optimal plans; and (ii) applying the sequence must achieve the goal (\(A\) exits) in exactly the declared number of moves. These guardrails allow us to reject superficially plausible but illegal or suboptimal sequences and to attribute improvements to better internal search rather than looser grading.

\paragraph{Configs and model release.}
The full configs, exactly as used for training, are available in our repository under \texttt{recipes/}. We also release all trained models (including checkpoints) on Hugging Face, listed in Table~\ref{tab:hf-models}.

\begin{table*}[t]
  \centering
  \small
  \setlength{\tabcolsep}{3.5pt}
  \renewcommand{\arraystretch}{1.05}
  \begin{tabular*}{\textwidth}{@{} l l p{0.70\textwidth} @{}}
    \toprule
    \textbf{Model} & \textbf{Domain} & \textbf{Hugging Face repository} \\
    \midrule
    Llama\,3.1--8B     & Math  &
    \texttt{https://huggingface.co/od2961/Llama-8B-Open-R1-GRPO-math-v1} \\
    Qwen\,2.5--1.5B    & Xwords &
    \texttt{https://huggingface.co/od2961/Qwen2.5-1.5B-Open-R1-GRPO-Crosswords-v03} \\
    Qwen\,2.5--1.5B    & RHour &
    \texttt{https://huggingface.co/od2961/Qwen2.5-1.5B-Open-R1-GRPO-carpark-v1} \\
    Qwen\,2.5--7B      & Math  &
    \texttt{https://huggingface.co/od2961/Qwen2.5-7B-Open-R1-GRPO-math-7b} \\
    Qwen\,2.5--1.5B    & Math  &
    \texttt{https://huggingface.co/od2961/Qwen2.5-1.5B-Open-R1-GRPO-math-v1} \\
    \bottomrule
  \end{tabular*}
  \caption{\textbf{Released GRPO-trained checkpoints.} Public Hugging Face repositories containing trained models and intermediate checkpoints used in this work.}
  \label{tab:hf-models}
\end{table*}

\subsection{Prompt Robustness \& Evaluation}
\label{app:prompt-robustness}

\paragraph{Robustness to system-prompt wording.}
To probe how sensitive performance is to \emph{system-prompt wording}, we evaluated \(K{=}5\) paraphrased system prompts for \texttt{Qwen2.5-1.5B} (Open-R1 GRPO, trained on \texttt{Math220k}) on \texttt{MATH-500} at decoding temperature \(T{=}0\), using randomized item order and a short prefilter on input length (350 characters; Table~\ref{tab:judge-reliability}). For each epoch and prompt variant, we compute standard test accuracy and then summarize the distribution across the five prompts. Across variants, accuracy changes only modestly, and the qualitative conclusions reported in the main text are unchanged. We therefore report main results using the canonical system prompt shown above, and use the prompt ensemble only to quantify prompt-induced variance.

\begin{table*}[t]
\centering
\footnotesize
\setlength{\tabcolsep}{4pt}
\renewcommand{\arraystretch}{0.95}

\begin{tabular*}{\textwidth}{@{\extracolsep{\fill}} l c c c @{}}
\toprule
Epoch & Mean accuracy & Std.\ across prompts & Range (min, max) \\
\midrule
0 (pre) & 31.8 & 0.8 & (30.8, 33.2) \\
1 (ckpt 500) & 38.8 & 1.2 & (36.6, 40.0) \\
2 (ckpt 1000) & 38.3 & 1.0 & (36.8, 39.8) \\
3 (final) & 40.2 & 0.7 & (39.5, 41.4) \\
\bottomrule
\end{tabular*}
\caption{Accuracy stability across system-prompt paraphrases for \textsc{MATH-500} (Qwen-1.5B). Each row summarizes accuracy over \(K{=}5\) system prompts at temperature \(0\), using the same test problems and a prefilter window of 350 input characters.}
\label{tab:judge-reliability}
\end{table*}

\paragraph{Reproducibility and evaluation.}
Prompts are released verbatim below. We use the same decoding policy across checkpoints (temperature, top-\(p\), stop criteria), cache RNG seeds, and reject outputs that violate the format contracts before computing task-specific rewards. This protocol ensures that improvements in correctness or in ``Aha!'' prevalence reflect changes in the model’s internal state rather than changes in instructions or graders. Figures \ref{fig:prompt-xword}, \ref{fig:prompt-math}, \ref{fig:prompt-rush} show our verbatim system level prompts.

\subsection{Model Training (GRPO Setup)}
\label{app:grpo-setup}

\begin{table*}[t]
\centering
\small
\setlength{\tabcolsep}{4pt}
\renewcommand{\arraystretch}{1.05}
\begin{tabular}{l l l}
\toprule
\textbf{Task / Model} & \textbf{Dataset} & \textbf{Key GRPO settings} \\
\midrule
\makecell[l]{\textbf{Math} (1.5B Qwen)} 
 & OpenR1--Math--220k 
 & \makecell[l]{LR $5\!\times\!10^{-6}$; bs/dev=8; \\ grad\_acc=64; epochs=3; \\ num\_gens=4; max\_prompt=512; \\ max\_completion=750; reward=\texttt{pure\_accuracy\_math}; \\ KL target 0.07, init\_KL 3.0} \\
\midrule
\makecell[l]{\textbf{Math} (7B Qwen)}
 & OpenR1--Math--220k
 & \makecell[l]{LR $5\!\times\!10^{-6}$; bs/dev=2; \\ grad\_acc=32; epochs=3; \\ num\_gens=4; max\_prompt=450; \\ max\_completion=750; reward=\texttt{pure\_accuracy\_math}} \\
\midrule
\makecell[l]{\textbf{Math} (8B Llama\,3.1)}
 & OpenR1--Math--220k
 & \makecell[l]{LR $5\!\times\!10^{-6}$; bs/dev=2; \\ grad\_acc=8; epochs=3; \\ num\_gens=4; max\_prompt=450; \\ max\_completion=750; reward=\texttt{pure\_accuracy\_math}; \\ PPO clip 0.10} \\
\midrule
\makecell[l]{\textbf{Xword} (1.5B Qwen)}
 & Guardian--Cryptonite (official split)
 & \makecell[l]{LR $1\!\times\!10^{-5}$; bs/dev=4; \\ grad\_acc=256; epochs=3; \\ num\_gens=8; \texttt{return\_reason}=true; \\ max\_reason=275; max\_completion=320; \\ reward=\texttt{pure\_accuracy} (0/1 + shaping)} \\
\midrule
\makecell[l]{\textbf{RHour} (1.5B Qwen)}
 & Rush\,4/5/6--balanced
 & \makecell[l]{LR $5\!\times\!10^{-6}$; bs/dev=4; \\ grad\_acc=64; epochs=3; \\ num\_gens=4; \texttt{return\_reason}=true; \\ max\_prompt=3000; max\_completion=300; \\ reward=\texttt{rush\_solution\_shaped}} \\
\bottomrule
\end{tabular}
\caption{Per-domain GRPO run settings. Values shown are the \emph{run-time} choices from the YAML configs; optimizer, KL control, horizons, and logging match the overview text.}
\label{tab:grpo-configs}
\end{table*}

\paragraph{Overview.}
We fine-tune instruction models (Qwen\,2.5--1.5B, Qwen\,2.5--7B, Llama\,3.1--8B) with Group Relative Policy Optimization (GRPO) \cite{shao2024deepseekmathpushinglimitsmathematical}, using task-specific, tag-constrained prompts that place private reasoning in \tagtt{think} and a single, machine-checkable response in \tagtt{answer} (See App.~\S\ref{app:system-level-prompts}).

\paragraph{Rollout + training architecture.}
We use the OpenR1 GRPO trainer \citep{openr1} with a vLLM inference server for on-policy rollouts and \texttt{accelerate}+DeepSpeed ZeRO-3 for training. A dedicated GPU hosts vLLM; the remaining GPUs run GRPO. Mixed precision is \texttt{bf16} for training; vLLM runs \texttt{fp16}. DeepSpeed is configured with ZeRO-3, CPU offload for parameters/optimizer, and overlap-comm; the \texttt{accelerate} configuration uses four to seven processes depending on available devices.

\paragraph{Domain--specific reward functions.}
All rewards are per-sample and clipped to $[0,1]$.
\begin{itemize}[leftmargin=*,itemsep=2pt]
  \item \textbf{Xwords.} Exact match on the inner \tagtt{answer} (strict 0/1) plus two shaping signals: (i) a tiny “contains as a standalone word” bonus, scaled by a \emph{tag factor} (fraction of \{\tagtt{think}, \etagtt{think}, \tagtt{answer}, \etagtt{answer}\} present), and (ii) a “Xwords accuracy” term that linearly ramps with \tagtt{think} length and is multiplied by the same tag factor; optional enumeration checks reject length mismatches.
  \item \textbf{Math.} Requires the full tag template; the gold and predicted \tagtt{answer} are canonicalized (LaTeX/math normalization) and compared for exact equality (0/1).
  \item \textbf{RHour.} Composite score combining \emph{exact} (canonical token sequence), \emph{prefix} (longest common prefix vs.\ gold), \emph{solve} (shorter optimal solutions $\uparrow$), and a planning heuristic $\Phi$ (distance/blockers decrease $\uparrow$); when a board is provided we legality-check and simulate moves, otherwise a gold-only variant supplies solve/prefix shaping. Defaults (used here): $w_\text{exact}{=}0.65$, $w_\text{solve}{=}0.20$, $w_\text{prefix}{=}0.10$, $w_\Phi{=}0.05$. 
\end{itemize}

\paragraph{Optimization and KL control.}
Across runs we use cosine LR schedules with warmup, clipped advantages/values, and KL control targeting $\text{KL}_\text{target}\!\approx\!0.07$ via an adaptive coefficient ($\beta$) with horizon $50$k and step size $0.15$; value loss weight $0.25$; PPO/GRPO clip ranges $0.05$--$0.10$ depending on run; horizon $1024$; $\gamma{=}0.99$, $\lambda_\text{GAE}{=}0.95$.

\paragraph{Prompt templates and budgets.}
We use fixed system prompts per domain that enforce exact formatting (no deviation), encourage compact reasoning, and cap \tagtt{think}/\tagtt{answer} token budgets. This standardization lets the rewards remain reliable and comparable across checkpoints and temperatures.

\paragraph{Per-domain GRPO configurations.}
Table~\ref{tab:grpo-configs} summarizes only the run-level choices that differ by domain/model; all other defaults follow the overview above.

\paragraph{DeepSpeed/Accelerate settings.}
We train with ZeRO-3 (stage 3), CPU offload for parameters and optimizer, and the standard single-node launcher; the provided \texttt{accelerate} config sets \texttt{bf16} mixed precision and \texttt{num\_processes} according to available training GPUs (vLLM occupies a dedicated device).

\paragraph{Operational notes.}
Jobs were launched under Slurm on 8-GPU nodes, using a mix of NVIDIA A100 and RTX A6000 GPUs. For the \textbf{Qwen2.5-1.5B} runs, we reserve \textbf{one} GPU for vLLM rollouts and use the remaining GPUs for \texttt{accelerate} training. For the larger \textbf{Qwen2.5-7B} and \textbf{Llama\,3.1-8B} runs, we reserve \textbf{two} GPUs for vLLM to support higher-throughput rollouts, with the remaining GPUs used for training. Per-run environment/caching settings and health-checks follow the batch script. The trainer logs per-step KL, policy/critic losses, and gradient norms; checkpoints are saved every 50 steps and pushed locally/HF Hub per config.


\section{Shift Detection, ``Aha!'' Detection, and Annotation}
\label{sec:app-detection-and-annotation}

This appendix details the pipeline used to (i) label mid-trace reasoning shifts in individual generations and (ii) operationalize \emph{formal} ``Aha!'' events as a checkpoint-level phenomenon.
We first present our formal ``Aha!'' detector, which combines prior-failure, prior-stability, and conditional-gain criteria (App.~\S\ref{sec:app-algorithm}).
We then describe the trace-level shift annotation protocol used throughout the paper (App.~\S\ref{app:detecting-aha}).
Then, we document the LLM-as-a-judge reliability protocol and the human-labeling template used for validation (App.~\S\ref{app:kappa-agreement} and App.~\S\ref{app:human-annotators-template}). Finally, we describe our triggered-reconsideration intervention (the fixed ``Wait~\ldots'' cue used to demarcate Pass~2 pivots) and explain how it avoids circularity with shift detection (App.~\S\ref{app:triggered-reconsideration}).

\subsection{Algorithm: Detecting an ``Aha!'' Moment}
\label{sec:app-algorithm}

\paragraph{Overview.}
Alg.~\ref{alg:aha-moment} operationalizes Def.~\ref{def:aha-moment-lrms} and Fig.~\ref{fig:aha-moment} via three checks:
\begin{enumerate}[label=(\roman*), leftmargin=*]
    \item \textbf{Prior failures:} for $q_j$, all checkpoints $i<k$ remain below a correctness ceiling.
    \item \textbf{Prior stability:} mid-trace shifts are rare for $i<k$.
    \item \textbf{Conditional gain at $k$:} when a mid-trace shift occurs at $k$, expected correctness increases by a prescribed margin.
\end{enumerate}

\paragraph{Estimating expected correctness.}
For each pair $(q_j,k)$ we draw $M$ independent traces $\tau^{(m)} \sim \pi_{\theta_k}(\cdot \mid q_j)$ under a fixed decoding policy (temperature $\tau$, top-$p$, and stop conditions held constant across $k$):
{\small
\[
\hat P_{\theta_k}(\checkmark \mid q_j)
\;=\;
\frac{1}{M}\sum_{m=1}^M R\!\big(\tau^{(m)}\big),
\qquad
R(\tau)\in\{0,1\}.
\]
}
For the conditional estimate we average only \emph{shifted} traces at $k$:
{\small
\[
\hat P_{\theta_k}(\checkmark \mid q_j,\; S_{q_j,k}{=}1)
\;=\;
\frac{\sum_{m=1}^M R(\tau^{(m)})\,\mathbb{1}[S(\tau^{(m)}){=}1]}
     {\sum_{m=1}^M \mathbb{1}[S(\tau^{(m)}){=}1] + \epsilon},
\]
}
with a tiny $\epsilon$ (e.g., $10^{-6}$) to avoid division by zero. If the denominator is $0$, Step~3 is inconclusive and the procedure returns \False.

\paragraph{Detecting mid-trace shifts ($S(\tau){=}1$).}
We mark a generation as \emph{shifted} if it contains \emph{both}:
(i) a lexical cue of reconsideration (e.g., ``wait'', ``recheck'', ``let's try'', ``this fails because~\dots''), and
(ii) a \emph{material} revision of the preceding plan (rejects/corrects an earlier hypothesis, switches method or candidate, or resolves a contradiction).
We implement this with a conservative two-stage detector (lexical cue prefilter + rubric-guided adjudication) described in App.~\S\ref{app:detecting-aha}.
To calibrate superficial hedges and edge cases, we tuned thresholds for the cue matcher and clamping on a small, human-verified set (App.~\S\ref{app:human-annotators-template}).

\paragraph{Prior stability (Step 2).}
For each $i<k$,
\[
\widehat{\Pr}[S_{q_j,i}{=}1]
\;=\;
\frac{1}{M}\sum_{m=1}^M \mathbb{1}\!\big[S(\tau^{(m)}_{i}){=}1\big],
\]
and we require $\widehat{\Pr}[S_{q_j,i}{=}1] < \delta_2$ for all $i<k$.

\paragraph{Thresholds and statistical test (Step 3).}
We set $(\delta_1,\delta_2,\delta_3)$ on a held-out development slab by maximizing F1 for \textsc{Aha} vs.\ non-\textsc{Aha} against human labels.
Unless stated otherwise, we use $\delta_1{=}0.125$ (prior correctness ceiling), $\delta_2{=}0.125$ (shift-rate ceiling), and $\delta_3{=}0.125$ (minimum gain).
To guard against Monte Carlo noise in $\hat P$, we further require the one-sided bootstrap CI (2000 resamples over traces) for
$\hat P_{\theta_k}(\checkmark \mid q_j, S{=}1) - \hat P_{\theta_k}(\checkmark \mid q_j)$
to exceed $0$ at level $\alpha{=}0.05$.
If this test fails, Step~3 returns \False.

\begin{algorithm}[t]
\small
\caption{Detecting an ``Aha!'' Moment for question $q_j$ at checkpoint $k$}
\label{alg:aha-moment}
\DontPrintSemicolon
\KwIn{Checkpoints $\{f_{\theta_i}\}_{i=0}^K$, question $q_j$, thresholds $\delta_1,\delta_2,\delta_3$.}
\KwOut{Boolean flag \texttt{Aha}[j,k].}

\textbf{Step 1: Prior failures}\;
\For{$i \gets 0$ \KwTo $k{-}1$}{
  $p_i \gets P_{\theta_i}(\text{correct}\mid q_j)$\;
  \If{$p_i \ge \delta_1$}{\Return \False \tcp*{Prior success breaks failure condition}}
}

\textbf{Step 2: Prior stability}\;
\For{$i \gets 0$ \KwTo $k{-}1$}{
  $s_i \gets \Pr[S_{q_j,i}{=}1]$\;
  \If{$s_i \ge \delta_2$}{\Return \False \tcp*{Too many prior shifts}}
}

\textbf{Step 3: Performance gain}\;
$p_k \gets P_{\theta_k}(\text{correct}\mid q_j)$\;
$p^{\text{shift}}_{k} \gets P_{\theta_k}(\text{correct}\mid q_j,\; S_{q_j,k}{=}1)$\;
\If{$p^{\text{shift}}_{k} - p_k > \delta_3$}{\Return \True}
\Else{\Return \False \tcp*{No significant gain}}
\end{algorithm}

\paragraph{Decoding protocol.}
Unless noted otherwise, we use $M{=}8$ samples per $(q_j,k)$, top-$p{=}0.95$, temperature $\tau{=}0.7$,
and truncate at the first full solution (math), full entry parse (xword), or solved board state (RHour).
We cache RNG seeds so cross-checkpoint comparisons differ only by $\theta_k$.

\paragraph{Complexity and caching.}
The detector uses $O(JKM)$ forward passes (plus inexpensive prefiltering and adjudication), where $J$ is the number of items and $K$ the number of checkpoints.
We cache $\{\tau^{(m)}, R(\tau^{(m)}), S(\tau^{(m)})\}$ per $(q_j,k)$ for reuse in ablations (temperature sweeps, entropy bins).

\paragraph{Edge cases and fallbacks.}
(i) If any $i<k$ violates prior failure ($\hat P_{\theta_i}{\ge}\delta_1$), return \False.
(ii) If no shifted traces occur at $k$, Step~3 is inconclusive (\False).
(iii) Extremely small $M$ inflates variance; we mark detections as ``\emph{provisional}'' if the bootstrap half-width of either probability exceeds $0.08$ and exclude them from aggregates.

\paragraph{Diagnostics.}
For each $(q_j,k)$ we log:
the prior-failure margin $\delta_1-\max_{i<k}\hat P_{\theta_i}$,
the stability margin $\delta_2-\max_{i<k}\widehat{\Pr}[S{=}1]$,
the gain $\hat\Delta=\hat P_{\theta_k}(\checkmark \mid S{=}1)-\hat P_{\theta_k}(\checkmark)$ with its CI,
and short excerpts around the first cue marker for audits (App.~\S\ref{app:detecting-aha}).

\paragraph{Limitations.}
Our shift detector may miss unlexicalized representational changes (false negatives) and can be triggered by surface hedges if the adjudicator fails (false positives).
The bootstrap addresses variance within checkpoints but not dataset shift across checkpoints; we therefore hold decoding hyperparameters fixed across~$k$.

\subsection{Detecting Reasoning Shifts in Traces}
\label{app:detecting-aha}

We flag a binary \emph{shift in reasoning} inside the \tagtt{think} block. A trace is labeled \textbf{TRUE} only if it exhibits both:
(A) an explicit lexical cue of reconsideration, and
(B) a \emph{material revision} of the preceding plan (rejects/corrects an earlier hypothesis, switches method or candidate, or resolves a contradiction).
Otherwise the label is \textbf{FALSE}.

\begin{algorithm}[h]
\scriptsize
\caption{Detecting a reasoning shift in a single trace $\tau$}
\label{alg:detect-shift}
\KwIn{Trace $\tau$; cue whitelist $\mathcal{W}$; judge $\mathcal{J}$ with strict JSON schema.}
\KwOut{$S(\tau)\in\{\True,\False\}$.}

Extract $t \leftarrow \tau.\tagtt{think}$ (clamp to 4{,}096 characters)\;
$c \leftarrow \textsc{PrefilterCues}(t;\mathcal{W})$\;
\If{$c$ is empty}{\Return \False}
$v \leftarrow \mathcal{J}(t, c)$ \tcp*{rubric-guided verdict in JSON}
\If{$v$ is invalid JSON}{\Return \False}
\If{$v.\texttt{shift\_in\_reasoning}=\True$}{\Return \True}
\Else{\Return \False}
\end{algorithm}

\paragraph{Annotation pipeline.}
Given checkpointed JSONL outputs, we annotate each trace in four steps:
\begin{enumerate}[leftmargin=*]
  \item \textbf{Parse.} Extract \tagtt{think} and \tagtt{answer} with a robust regex; clamp \tagtt{think} to 4{,}096 characters.
  \item \textbf{Cue prefilter (A).} Search \tagtt{think} for any cue from a whitelist (Table~\ref{tab:shift-cue-whitelist}). If none is present, assign \textbf{FALSE}.
  \item \textbf{Material revision check (B).} For prefilter hits, query an LLM judge (GPT--4o) with a rubric that restates (A)+(B) and requests a strict JSON verdict plus short before/after excerpts around the first cue. If the verdict is uncertain or invalid, assign \textbf{FALSE}.
  \item \textbf{Record.} Write the Boolean label and minimal diagnostics (markers, first-cue offset, excerpts) back to the record; processing order is randomized with a fixed seed. The procedure is idempotent---existing labels are left unchanged.
\end{enumerate}
This conservative policy (requiring both an explicit cue and a substantiated revision, and defaulting to \textbf{FALSE} on uncertainty) keeps false positives low and yields conservative prevalence estimates.

\begin{figure*}[t]
  \centering
  \begin{promptbox}[width=\textwidth]{LLM-as-a-Judge — System Prompt (Shift in Reasoning)}
You are a careful annotator of single-pass reasoning transcripts.
Your task is to judge whether the writer makes a CLEAR, EXPLICIT "shift in reasoning"
within \tagtt{think}...\etagtt{think}.

A TRUE label requires BOTH:
(A) an explicit cue (e.g., "wait", "hold on", "scratch that", "contradiction"),
AND (B) a material revision of the earlier idea (reject/correct an initial hypothesis,
pick a new candidate, fix a contradiction, or change device/method).

Do NOT mark TRUE for rhetorical transitions, hedging, or generic connectives
without an actual correction. Judge ONLY the content inside \tagtt{think}.
Be conservative; these events are rare.
  \end{promptbox}
  \caption{\textbf{LLM-as-a-Judge (system prompt).} One instruction template used to adjudicate whether a \tagtt{think} trace contains a bona fide reasoning shift (explicit cue \emph{and} material revision).}

  \centering
  \begin{promptbox}[width=\textwidth]{LLM-as-a-Judge — User Template (filled per example)}
Problem/Clue (if available):
{problem}

PASS-1 \tagtt{think} (truncated if long):
{think}

Heuristic cue candidates (may be empty): {cues}
first_marker_pos: {pos}

Return ONLY a compact JSON object with keys:
- shift_in_reasoning: true|false
- confidence: "low"|"medium"|"high"
- markers_found: string[]       (verbatim lexical cues you relied on)
- first_marker_index: integer   (character offset into \tagtt{think}, -1 if absent)
- before_excerpt: string        (<=120 chars ending right before the first marker)
- after_excerpt: string         (<=140 chars starting at the first marker)
- explanation_short: string     (<=140 chars justification)
  \end{promptbox}
  \caption{\textbf{LLM-as-a-Judge (user template).} Per-example payload including the clamped \tagtt{think} text, problem/clue, and whitelist-prefiltered cue markers/position, with a strict JSON schema for the verdict.}
  \label{fig:judge-user-template}
\end{figure*}

\paragraph{Whitelist (lexical cues).}
To bias the LLM-as-a-judge toward \emph{explicit} reconsideration, we pre-filter traces using a hand-crafted list of lexical cues.
Concretely, we match case-insensitive regex patterns over the \tagtt{think} text (Table~\ref{tab:shift-cue-whitelist}), covering common morphology and light paraphrase (e.g., ``wait'', ``hold on'', ``scratch that'', ``I was wrong'', ``misread'', ``re-check'', etc.).
Cues are grouped semantically in the implementation (e.g., \texttt{src/annotate/core/prefilter.py}).
A positive shift label is only accepted when at least one explicit cue is present---either from the prefilter or from cues the judge itself extracts.

\begin{table*}[t]
\centering
\footnotesize
\setlength{\tabcolsep}{5pt}
\renewcommand{\arraystretch}{1.02}
\caption{\textbf{Cue list} (lexical indicators for reconsideration). Each row denotes a family of regex triggers; variants and minor orthographic differences are included.}
\label{tab:shift-cue-whitelist}
\begin{tabularx}{\textwidth}{@{} l Y @{}}
\toprule
\textbf{Category} & \textbf{Representative cues (lemmas/phrases)} \\
\midrule
Pauses \& self--interruptions &
\emph{wait}, \emph{hold on/up}, \emph{hang on}, \emph{one/just a second}, \emph{give me a moment}, \emph{pause}, \emph{on second/further thought}, \emph{reconsider}, \emph{rethink} \\
Explicit pivots/corrections &
\emph{actually}, \emph{in fact}, \emph{rather}, \emph{instead (of)}, \emph{let’s fix/correct that}, \emph{correction:}, \emph{to correct}, \emph{change/switch to}, \emph{replace with}, \emph{try/consider instead}, \emph{alternate/alternative}, \emph{new candidate/answer/approach} \\
Immediate reversals &
\emph{no, that/this/it ...}, \emph{never mind/nvm}, \emph{disregard/ignore that}, \emph{scratch/strike/forget that}, \emph{I retract/take it back}, \emph{I stand corrected}, \emph{not X but/rather Y} \\
Error admissions &
\emph{I was wrong / that’s wrong / incorrect}, \emph{(my) mistake}, \emph{my bad}, \emph{oops/whoops}, \emph{apologies}, \emph{erroneous} \\
``Mis-*'' failures &
\emph{misread}, \emph{miscount}, \emph{miscalculate / calculation error}, \emph{misapply}, \emph{misparse}, \emph{misspell}, \emph{misindex}, \emph{misuse}, \emph{conflated}, \emph{typo}, \emph{off-by-one} \\
Constraint/length mismatches (xword) &
\emph{doesn’t fit/match (length/pattern)}, \emph{letters don’t fit}, \emph{pattern/length mismatch}, \emph{too many/few letters}, \emph{wrong length}, \emph{violates enumeration}, \emph{doesn’t parse}, \emph{definition mismatch}, \emph{not an anagram of}, \emph{fodder mismatch} \\
Contradictions/impossibility &
\emph{contradiction}, \emph{inconsistent}, \emph{can’t/cannot be}, \emph{impossible}, \emph{doesn’t make sense / add up}, \emph{cannot both}, \emph{leads to a contradiction} \\
Re--check / backtrack &
\emph{recheck / double--check / check again}, \emph{re--evaluate / re--examine / upon review/reflection}, \emph{backtrack}, \emph{start over/restart/reset/from scratch} \\
``Prev X, but …'' templates &
\emph{I (initially/originally) thought ... but/however}, \emph{previously ... but/however}, \emph{earlier ... but/however}, \emph{however ... correct/fix/instead/rather/change} \\
Omission/oversight &
\emph{I forgot/missed/overlooked/ignored}, \emph{didn’t notice}, \emph{misremembered/misheard} \\
Directional swaps &
\emph{reversed / backwards}, \emph{swapped}, \emph{mixed up} \\
Realization formulas &
\emph{turns out}, \emph{I (now) realize}, \emph{on reflection}, \emph{after all} \\
Failure templates &
\emph{fails because}, \emph{won’t work / not working}, \emph{dead end} \\
\bottomrule
\end{tabularx}
\end{table*}

\paragraph{Blacklist (negatives \& exclusions).}
We reject as insufficient evidence:
(i) bare discourse markers without correction (\emph{but}, \emph{however}, \emph{therefore}, \emph{also});
(ii) hedges or meta-verbosity (\emph{maybe}, \emph{perhaps}, \emph{I think}, \emph{let’s be careful}) without an explicit pivot;
(iii) formatting or notational fixes only;
(iv) device/method names listed without rejecting a prior attempt; and
(v) cues appearing \emph{outside} \tagtt{think}.
The judge prompt enforces these, and our implementation forces \textbf{FALSE} when no explicit cue is present.

\paragraph{Material-revision test (B).}
The judge must justify that the post-cue span negates or corrects a prior claim, selects a different candidate, changes the solving device/method, or resolves a contradiction.
We store short \texttt{before/after} excerpts around the first cue to aid audits, and we only accept a \textbf{TRUE} label when the judge's JSON is parseable and consistent with the excerpts. Otherwise we default to \textbf{FALSE}.

\paragraph{Error handling, privacy, and rate limits.}
If the judge call fails or returns invalid JSON, we save the prompt to a local log file, stamp \textbf{FALSE}, and continue.
We clamp long \tagtt{think} segments before sending to the judge.
Optional jitter (default $\le 0.25$s) randomizes inter-call delays.

\paragraph{Reproducibility.}
We fix a shuffle seed for candidate order, sort files by natural \texttt{stepNNN} and path, and perform atomic rewrites.
The detector is content-idempotent: re-running will skip annotated lines and only fill missing fields.

\paragraph{Limitations.}
The whitelist privileges \emph{explicit} cues and may miss unlexicalized pivots (false negatives).
Conversely, some cues can appear in non-revisional discourse; the material-revision test mitigates but does not eliminate such false positives.
Because we default to \textbf{FALSE} on uncertainty, prevalence estimates are conservative.

\subsection{LLM-as-a-Judge Protocol and Reliability}
\label{app:kappa-agreement}

\paragraph{Bias mitigation.}
We use GPT--4o as a scalable surrogate for shift annotation and address known judge biases---position, length, and model-identity---with a three-part protocol \citep{Wang2024FairEval, shi2024positionbias, li-etal-2024-split}:
\begin{enumerate}[leftmargin=*]
  \item \textbf{Order randomization.} We randomly permute items and (when applicable) apply split--merge aggregation to neutralize position effects \citep{shi2024positionbias}.
  \item \textbf{Rubric-anchored scoring.} GPT--4o completes a structured JSON rubric, following G-Eval-style guidance \citep{liu2023geval}.
  \item \textbf{Prompt-variant stability.} We re-query with $K{=}5$ judge-prompt variants at judge temperature $0$ and report inter-prompt agreement.
\end{enumerate}
Table~\ref{alt-prompts} lists the five judge prompt variants; Table~\ref{tab:judge-reliability-ckpt} summarizes inter-prompt agreement on a fixed evaluation set.

\begin{table}[t]
\centering
\footnotesize
\setlength{\tabcolsep}{6pt}
\renewcommand{\arraystretch}{0.95}
\begin{tabular}{l p{0.78\linewidth}}
\toprule
Variant & System prompt summary \\ \midrule
v1 & Baseline strict judge: explicit cue (e.g., ``wait'', ``hold on'', ``scratch that'', ``contradiction'') AND a material revision; ignore hedging; judge only the \tagtt{think} span. \\
v2 & Audit \tagtt{think} for change of course: cue + substantive revision required; ignore rhetorical connectives; conservative. \\
v3 & ``Corrects themselves mid-thought'': needs an explicit reconsideration cue and a replacement/fix of prior approach; ignore small edits/hedges. \\
v4 & ``Quality control'': cue + meaningful course change; minor tweaks/hedging are not shifts; judge only the \tagtt{think} span. \\
v5 & ``Spot explicit change of mind'': cue + real update (reject/swap/repair); true shifts are rare. \\
\bottomrule
\end{tabular}
\caption{Judge prompt variants v1--v5 used for shift-in-reasoning annotation.}
\label{alt-prompts}
\end{table}

\paragraph{Logged annotations.}
For each trajectory we record:
(i) graded correctness,
(ii) shift/no-shift label,
(iii) whether a shift improved correctness,
(iv) GPT--4o's confidence (low/med/high), and
(v) auxiliary statistics (e.g., entropy).
This separation supports analyses of shift \emph{prevalence} versus shift \emph{efficacy}.

\paragraph{Prefiltering.}
Before judging, we apply a cue-based prefilter (App.~\S\ref{app:detecting-aha}).
Empirically, responses without cue words almost never contain qualifying shifts (human annotation of 100 such responses from Qwen-1.5B on \textsc{MATH-500} found none).

\paragraph{Reliability (inter-prompt agreement).}
We evaluated Qwen-1.5B (GRPO) on \textsc{MATH-500}, using five paraphrased judge prompts (v1--v5), randomized item order, and judge temperature $0$.
Table~\ref{tab:judge-reliability-ckpt} reports percent agreement (PO), mean pairwise Cohen's~$\kappa$, and 95\% bootstrap CIs.

\begin{table*}[t]
\centering
\footnotesize
\setlength{\tabcolsep}{4pt}
\renewcommand{\arraystretch}{0.95}
\begin{tabular*}{\textwidth}{@{\extracolsep{\fill}} l c c c @{}}
\toprule
Epoch & Judged $N$ & Mean PO & Mean $\kappa$ \; [95\% CI] \\
\midrule
0 (pre)       & 500 & 0.983 & 0.655 \; [0.507, 0.775] \\
1 (ckpt 500)  & 500 & 0.986 & 0.759 \; [0.606, 0.863] \\
2 (ckpt 1000) & 500 & 0.988 & 0.770 \; [0.631, 0.868] \\
3 (final)     & 500 & 0.988 & 0.719 \; [0.526, 0.848] \\
\bottomrule
\end{tabular*}
\caption{Inter-prompt agreement on binary reasoning-shift labels for \textsc{MATH-500} (Qwen-1.5B). Settings: $K{=}5$ judge-prompt variants, model decoding temperature $0.7$ (for the generated traces), judge temperature $0$, cue prefilter window 350 characters, randomized item order. PO = percent agreement.}
\label{tab:judge-reliability-ckpt}
\end{table*}

\paragraph{Human validation.}
Relative to a human majority-vote reference on 20 examples, GPT--4o achieved Cohen's $\kappa=0.794$ with PO$=0.900$.
Mean human--human agreement was lower (PO$=0.703$, mean pairwise $\kappa=0.42$), and mean LLM--human agreement was intermediate (PO$=0.758$, mean pairwise $\kappa=0.51$).
Table~\ref{tab:human-validation} summarizes these comparisons.

\begin{table}[t]
  \centering
  \small
  \setlength{\tabcolsep}{5pt}
  \renewcommand{\arraystretch}{1.05}
  \begin{tabular}{l r r r}
    \toprule
    \textbf{Comparison} & \textbf{$N$} & \textbf{PO} & \textbf{Cohen's $\kappa$} \\
    \midrule
GPT--4o vs.\ human majority vote & 20 & 0.900 & 0.794 \\
Mean human--human (pairwise)     & 20 & 0.703 & 0.42 \\
Mean LLM--human (pairwise)       & 20 & 0.758 & 0.51 \\
    \bottomrule
  \end{tabular}
  \caption{\textbf{Human validation of shift labels.}
  We compare GPT--4o shift judgments against a human majority-vote reference on a 20-item validation set (PO = percent agreement). We also report mean pairwise Cohen's $\kappa$ among human annotators and between GPT--4o and individual humans on the same items.}
  \label{tab:human-validation}
  \vspace{-3mm}
\end{table}

\paragraph{Reproducibility.}
We include the full rubric and sample items from our human annotation survey in App.~\S\ref{app:human-annotators-template}.

\subsection{Human Annotators Template}
\label{app:human-annotators-template}

\paragraph{Annotator pool \& consent.}
We used $6$ \textit{volunteer} adult annotators (unpaid), recruited from the authors' academic networks. Participants gave informed consent on the task page and could withdraw at any time. No sensitive personal information was requested.

\paragraph{IRB status.}
This activity consisted solely of judgments about model-generated text and did not involve collection of sensitive data or interventions with human participants. Under our institutional guidelines, it does not constitute human-subjects research; consequently, no IRB review was sought.

\paragraph{Presentation \& blinding.}
Items were shown in randomized order. Annotators saw the original \emph{Question asked} and the verbatim \tagtt{think} trace (with tags preserved; traces clamped to 4096 characters). Model family, size, checkpoint, temperature, and correctness signals were withheld.

\paragraph{Labels \& rubric.}
Primary label: \textbf{Yes/No} (shift present). Optional fields: confidence (low/med/high), first cue index (character offset), and a one-sentence rationale. Edge cases defaulted to \textbf{No} unless a method switch (e.g., completing-the-square $\to$ factoring; permutations $\to$ stars-and-bars; prime factorization $\to$ Euclidean algorithm) was evident.

\paragraph{Calibration \& quality.}
Annotators completed a short calibration set (including Examples A--H) with immediate feedback. During labeling we interleaved hidden gold items and monitored time-on-item; submissions failing pre-registered thresholds were flagged for review.

\paragraph{Agreement \& adjudication.}
Each item received independent labels. We report Cohen's $\kappa$ with 95\% bootstrap CIs.

\paragraph{Data handling.}
We did not collect sensitive demographics. Released artifacts include prompts, anonymized traces (with \tagtt{think} clamps), labels, and aggregation scripts; any operational contact data (if present) were excluded from the release.

\begingroup\itshape
\paragraph*{\textit{Task.}}
Read the model's \tagtt{think} trace for a math problem and answer:
\textit{``Does this \tagtt{think} trace include a change in thinking?''}
Choices: \textbf{Yes / No}.

\paragraph*{\textit{When to mark \textbf{Yes}.}}
(1) The model clearly \textit{switches strategies} mid-trace. \;
(2) It abandons one method after noticing a contradiction, dead end, or mistake, and adopts a different method. \;
(3) This is a real strategy pivot, not a small fix.

\paragraph*{\textit{When to mark \textbf{No}.}}
(1) The model keeps using the same method throughout. \;
(2) It only makes minor arithmetic/algebra fixes. \;
(3) It adds detail or notation without changing approach. \;
\textit{Important:} cue words alone (``wait'', ``recheck'', etc.) do \textit{not} count; look for an actual method switch.

\paragraph*{\textit{Quick checklist.}}
Identify the \textit{initial method}. \;
Look for a \textit{pivot}: does the model drop that plan and adopt a different method? \;
Ignore small fixes. \;
Answer \textbf{Yes} only with a clear pivot; otherwise \textbf{No}.
\endgroup

\subsection*{Worked Examples (Gold-Labeled)}
\label{app:survey-worked-examples}

\paragraph{Example A — YES}
\textit{Question.} How many sides would there be in a convex polygon if the sum of all but one of its interior angles is \(1070^\circ\)?
\begin{promptbox}{Model \tagtt{think} trace}
\tagtt{think}… computes with a wrong assumption, gets \(360=90\) (contradiction), then re-evaluates and sets up
\(\theta=(n-2)\cdot 180^\circ-1070^\circ\) and solves under \(0^\circ<\theta<180^\circ\) …\etagtt{think}
\end{promptbox}
\textit{Correct answer:} \textbf{Yes.}\quad
\textit{Why:} Notices a contradiction and switches approach.

\paragraph{Example B — NO}
\textit{Question.} Simplify \(3/\sqrt{27}\).
\begin{promptbox}{Model \tagtt{think} trace}
\tagtt{think}… \(\sqrt{27}=3\sqrt{3}\)\(\;\to\;\) \(3/(3\sqrt{3})=1/\sqrt{3}\)\(\;\to\;\) rationalize \(\to\) \(\sqrt{3}/3\) …\etagtt{think}
\end{promptbox}
\textit{Correct answer:} \textbf{No.}\quad
\textit{Why:} One method throughout (simplify radical \(\to\) rationalize).

\paragraph{Example C — YES}
\textit{Question.} Solve \(x^{2}-5x-14=0\).
\begin{promptbox}{Model \tagtt{think} trace}
\tagtt{think}… tries completing the square, finds it awkward, then switches to factoring \((x-7)(x+2)\) …\etagtt{think}
\end{promptbox}
\textit{Correct answer:} \textbf{Yes.}\quad
\textit{Why:} Switch from completing the square to factoring.

\paragraph{Example D — NO}
\textit{Question.} Compute \(\dfrac{d}{dx}\left(\dfrac{x^{2}+3x+2}{x+1}\right)\).
\begin{promptbox}{Model \tagtt{think} trace}
\tagtt{think}… uses the quotient rule; minor sign fix; simplify …\etagtt{think}
\end{promptbox}
\textit{Correct answer:} \textbf{No.}\quad
\textit{Why:} Stays with quotient rule; no pivot.

\paragraph{Example E — YES}
\textit{Question.} How many nonnegative integer solutions \((x,y)\) satisfy \(x+y=20\)?
\begin{promptbox}{Model \tagtt{think} trace}
\tagtt{think}… first frames as permutations, then switches to stars and bars: \(\binom{21}{1}=21\) …\etagtt{think}
\end{promptbox}
\textit{Correct answer:} \textbf{Yes.}\quad
\textit{Why:} Pivot from permutations framing to stars-and-bars.

\paragraph{Example F — NO}
\textit{Question.} What is the sum of the first \(50\) odd integers?
\begin{promptbox}{Model \tagtt{think} trace}
\tagtt{think}… recalls \(n^{2}\) \(\to\) \(50^{2}=2500\) …\etagtt{think}
\end{promptbox}
\textit{Correct answer:} \textbf{No.}\quad
\textit{Why:} Single recalled formula; no change in method.

\paragraph{Example G — YES}
\textit{Question.} Find \(\gcd(252,198)\).
\begin{promptbox}{Model \tagtt{think} trace}
\tagtt{think}… starts prime factoring, then switches to the Euclidean algorithm …\etagtt{think}
\end{promptbox}
\textit{Correct answer:} \textbf{Yes.}\quad
\textit{Why:} Abandons prime factorization for Euclidean algorithm.

\paragraph{Example H — NO}
\textit{Question.} Solve the system \(x+y=10,\; xy=21\).
\begin{promptbox}{Model \tagtt{think} trace}
\tagtt{think}… uses Vieta’s formulas \(\to\) factor \(\to\) \(\{7,3\}\) …\etagtt{think}
\end{promptbox}
\textit{Correct answer:} \textbf{No.}\quad
\textit{Why:} One consistent quadratic/Vieta approach.

\textbf{Annotation Questions, Verbatim.} Then, our annotators are asked to provide Yes/No answers as to whether a shift in reasoning occurred for the following $20$ questions. We have included one sample question here, to provide an example of the task:


\paragraph{Question 1: Does this \tagtt{think} trace include a change in thinking? (Yes/No)}
\textbf{Question Asked:} If $a$ and $b$ are positive integers such that $\gcd(a,b)=210$, $\mathop{\text{lcm}}[a,b]=210^3$, and $a<b$, how many possible values are there for $a$?

\textbf{Model Response:}
\tagtt{think}The greatest common divisor (GCD) of \(a\) and \(b\) is given as \(210\), and their least common multiple (LCM) is \(210^3\). This implies that both \(a\) and \(b\) share the prime factors of \(210\), which are \(2\), \(3\), and \(5\).

Let's express \(a\) and \(b\) in terms of their prime factorizations:
\[ a = 2^x \cdot 3^y \cdot 5^z \]
\[ b = 2^w \cdot 3^v \cdot 5^t \]

Given that \(\gcd(a,b) = 210 = 2^1 \cdot 3^1 \cdot 5^1\), we get the following equations from comparing exponents:

For the GCD:
\[ x = 1, \quad y = 1, \quad z = 1 \]

For the LCM:
\[ w = x, \quad v = y, \quad t = z \]

Thus,
\[ w = 1, \quad v = 1, \quad t = 1 \]

Since \(\mathop{\text{lcm}}[a,b] = 210^3\), substituting the values gives us:
\[ w + v + t = 3 \]

This simplifies our expressions for \(w\), \(v\), and \(t\) to:
\[ 1 + 1 + 1 = 3 \]

Therefore, the only combination satisfying these conditions is:
\[ w = 1, \quad v = 1, \quad t = 1 \]

Substituting back into the expression for \(a\) yields:
\[ a = 2^1 \cdot 3^1 \cdot 5^1 = 210 \]

However, since \(a < b\), this alone does not provide any new information because \(a = b\).

To find all pairs \((a, b)\) where \(a < b\) and \(\gcd(a, b) = 210\):
- We need to consider all combinations of \(a\) and \(b\) such that:
\[ 2^x \cdot 3^y \cdot 5^z < 2^{x'} \cdot 3^{y'} \cdot 5^{z'} \]

Considering \(x > x'\), \(y > y'\), and \(z > z'\):

- For \(x > x'\): Since \(x\) cannot be greater than 1 (as \(a\) must be less than \(b\)), the only possibility is \(x' = 0\) (i.e., \(a\) can take on any power of 2 except 0, making \(a = 210k\) where \(k\) is some integer).

- Similarly, considering \(y > y'\) and \(z > z'\), the same logic applies for powers of 3 and 5 respectively.

Thus, \(a\) could be \(210 \cdot 3^m \cdot 5^n\) for various combinations of \(m\) and \(n\).

Each choice of \(m\) and \(n\) uniquely determines \(a\), resulting in infinitely many possible values of \(a\) because there are infinitely many integers \(m\) and \(n\).\etagtt{think}

\textbf{Fill-in-the-bubble response}. Choices: Yes / No (no alternative / third choice was provided).

\textbf{LLM-as-a-Judge assessed answer}. "Yes".

\paragraph{Post Assessment} Post-assessment, we reveal the LLM-as-a-Judge answer to participants and encourage them to invite others to participate. Each individual's score was weighted equally, and we analyzed annotator agreement as described in \ref{app:kappa-agreement}. The complete assessment is made available as part of our codebase.

\subsection{Triggered Reconsideration Intervention}
\label{app:triggered-reconsideration}

In \S\ref{sec:rq3-uncertainty}, we study an \emph{extrinsically triggered} reconsideration mechanism (``forced Aha''): for each instance we first generate a baseline completion (Pass~1) and then re-query the model under the same decoding settings while appending a short reconsideration clause (Pass~2).
The intervention is not intended to \emph{define} a shift, but to create a consistent opportunity for deliberate re-evaluation.

\paragraph{Fixed pivot marker (deterministic demarcation).}
For Pass~2 we prepend a fixed sentence that begins the reconsideration clause (e.g., \emph{``Wait, we need to reconsider. Let's think this through step by step.''}).
This invariant prefix serves a purely operational role: it deterministically marks the intended pivot point across models, checkpoints, and temperatures, making the intervention location unambiguous in downstream analyses (e.g., excerpting or aligning text around the prompted reconsideration).

\paragraph{Avoiding circularity with shift detection.}
Because the intervention includes an explicit reconsideration cue, a lexical-only detector would trivially fire on Pass~2. We therefore emphasize that our canonical shift detector is \emph{not} a cue detector: it requires both (i) an explicit cue \emph{and} (ii) a \emph{material revision} of the plan within the \tagtt{think} trace (reject/correct a prior hypothesis, switch method/candidate, or resolve a contradiction with a substantively different approach; App.~\S\ref{app:detecting-aha}). Thus, the fixed ``Wait~\ldots'' prefix provides a standardized \emph{timestamp} for reconsideration, while the detector separately verifies whether the content exhibits a genuine strategy pivot rather than superficial cueing.

\paragraph{Implementation details.}
In all Pass~2 runs we hold decoding parameters fixed relative to Pass~1 (temperature, top-$p$, stop criteria, and token budgets) and vary \emph{only} the appended reconsideration text.
We also reuse the same answer-format contract (\tagtt{think}/\tagtt{answer}) so that validators and scoring remain comparable across passes.

\paragraph{Cue robustness and variants.}
The main paper reports results using a single canonical reconsideration clause for Pass~2.
To test whether gains depend on the specific lexical phrasing (rather than the act of prompting reconsideration), we repeat the intervention with multiple semantically similar but lexically distinct cues (C1--C3) and show that the qualitative conclusions persist; see App.~\S\ref{app:uncertainty-interventions}.

\section{Additional Results and Robustness Checks}
\label{sec:app-additional-results}

This appendix collects supplementary analyses that extend and stress-test the main results.
We first report the prevalence of \emph{formal} ``Aha!'' events under threshold grids and summarize cross-domain patterns (App.~\S\ref{app:aha-prevelance-descriptions}).
We then replicate key regressions and uncertainty analyses on larger model families (Qwen2.5--7B and Llama\,3.1--8B) to verify that the shift effects generalize beyond Qwen2.5--1.5B (App.~\S\ref{app:qwen-llama}).
Next, we test whether entropy-gated, \emph{extrinsically triggered} reconsideration is robust to the specific cue wording (App.~\S\ref{app:uncertainty-interventions}).
Finally, we evaluate external models (DeepSeek\textendash R1 and GPT\textendash 4o) under the same shift-detection protocol and compare alternative shift detectors (App.~\S\ref{app:external-models} and App.~\S\ref{app:shift-detector-rationale}).
Together, these checks show that our qualitative conclusions are stable across domains, model families/sizes, prompt variants, and detector choices.

\subsection{``Aha!'' Moment Prevalence}
\label{app:aha-prevelance-descriptions}

\paragraph{How to read the heatmaps.}
Each panel reports the share of problem--checkpoint pairs $(q_j,k)$ that satisfy our operational definition of an ``Aha!'' moment (Def.~\ref{def:aha-moment-lrms}) under a grid of thresholds:
$\delta_1\!\in\!\{0,\tfrac{1}{8},\tfrac{2}{8}\}$ (maximum prior accuracy),
$\delta_2\!\in\!\{0,\tfrac{1}{8},\tfrac{2}{8}\}$ (maximum prior shift rate),
and, unless noted, $\delta_3=\epsilon>0$ (any non-zero gain at $k$).
Cells show the percentage and the raw counts \((\#\text{events}/\#\text{pairs})\).
We aggregate over checkpoints $\leq\!1000$ with $G{=}8$ samples per item, and we use the conservative detector described in App.~\ref{app:detecting-aha} (lexical cue \emph{and} material revision; default to \textbf{FALSE} on uncertainty).

\paragraph{Cross-domain patterns.}
Three robust trends emerge across \emph{Xword}, \emph{Math}, and \emph{RHour} and across model families/sizes.
\begin{itemize}[leftmargin=*, itemsep=2pt]
  \item \textbf{Rarity.} Even under the lenient gain criterion $\delta_3\!=\!\epsilon$, ``Aha!'' events occupy a very small fraction of problem--checkpoint pairs. Most cells are near zero; none approach a large fraction. This mirrors the main-text finding that mid-trace shifts seldom coincide with measurable improvements.
  \item \textbf{Sensitivity to prior instability and prior accuracy.} Relaxing either prerequisite increases counts but remains small in magnitude. In particular, moving to higher $\delta_2$ (allowing more prior shifts, i.e., lower prior stability) and higher $\delta_1$ (allowing occasional prior solves) produces the visually “warmest’’ cells---consistent with the intuition that ``Aha!'' detections concentrate where traces have shown some volatility and the item is not maximally hard.
  \item \textbf{Domain/model differences.} \emph{RHour} exhibits a higher \emph{raw} shift rate (App.~\S\ref{sec:results-rq2}), but the ``Aha!'' filter (requiring a gain at $k$) prunes most cases; the absolute prevalence remains low. \emph{Xword} shows small pockets of higher prevalence when $\delta_1,\delta_2\!\geq\!\tfrac{1}{8}$, whereas \emph{Math} is uniformly sparse. Scaling from Qwen\,1.5B to 7B or switching to Llama\,3.1--8B does not materially increase prevalence.
\end{itemize}

\paragraph{Stricter gain thresholds.}
Replacing $\delta_3{=}\epsilon$ with a minimal absolute lift (e.g., at least one of the $G{=}8$ samples flips from incorrect to correct at $k$) further reduces counts but preserves the qualitative ordering across domains and models.

\paragraph{Takeaway.}
Across all settings, formal ``Aha!'' moments---requiring \emph{both} a mid-trace reasoning pivot and a contemporaneous performance gain---are \emph{vanishingly uncommon}.
The sparse, threshold-stable patterns in Figs.~\ref{fig:app-aha-grid-q15b-bytemp}--\ref{fig:app-aha-grid-q7b-l8b-bytemp} show this finding across temperatures, domains, and models.

\subsection{Formal Threshold Search}
\label{app:formal-threshold-search-q15b}

To make our threshold-selection procedure concrete, we ran the grid/bootstrapped threshold search across the stored Qwen2.5--1.5B evaluation outputs for each domain and temperature.

For each domain$\times$temperature root, we searched a small grid
$\delta_1,\delta_2\in\{0,1/8,2/8\}$ and $\delta_3\in\{\text{None},0,0.05,0.125\}$ (with \texttt{min\_prior\_steps=2}),
and selected the ``best'' configuration according to the script's default ranking
(maximize the bootstrap lower CI bound for the mean gain; ties broken by prevalence and mean gain).
Table~\ref{tab:formal-threshold-search-q15b} reports the best row per root. We report mean gain as
$100\cdot\mathbb{E}[\hat P(\checkmark\mid S{=}1)-\hat P(\checkmark)]$ in percentage points (pp), with a 95\% bootstrap CI over flagged pairs;
entries are ``--'' when no events are found or when $N$ is too small to form a stable CI.

\begin{table*}[t]
\centering
\small
\setlength{\tabcolsep}{5pt}
\renewcommand{\arraystretch}{1.05}
\begin{tabular}{l r c c c r r r l}
\toprule
\textbf{Domain} & \textbf{$T$} & $\boldsymbol{\delta_1}$ & $\boldsymbol{\delta_2}$ & $\boldsymbol{\delta_3}$ &
\textbf{events/pairs} & \textbf{prev (\%)} & \textbf{gain (pp)} & \textbf{CI (pp)} \\
\midrule
Math  & 0    & 1/8 & 1/8 & $\epsilon$ & 8/16000  & 0.05 & +0.00 & {[}+0.00, +0.00{]} \\
Math  & 0.05 & 2/8 & 2/8 & $\epsilon$  & 41/16000 & 0.26 & -2.74 & {[}-5.79, -0.30{]} \\
Math  & 0.3  & 2/8 & 2/8 & $\epsilon$  & 43/10000 & 0.43 & -2.62 & {[}-4.65, -1.16{]} \\
Math  & 0.7  & 2/8 & 2/8 & $\epsilon$  & 92/16000 & 0.57 & +1.22 & {[}-1.77, +4.76{]} \\
\midrule
Xwords & 0    & 0   & 0   & $\epsilon$  & 0/3120   & 0.00 & --    & -- \\
Xwords & 0.05 & 1/8 & 1/8 & $\epsilon$  & 3/3120   & 0.10 & +0.00 & {[}+0.00, +0.00{]} \\
Xwords & 0.3  & 1/8 & 1/8 & $\epsilon$  & 7/3120   & 0.22 & +0.00 & {[}+0.00, +0.00{]} \\
Xwords & 0.7  & 2/8 & 2/8 & $\epsilon$  & 18/3120  & 0.58 & +0.00 & {[}+0.00, +0.00{]} \\
\midrule
RHour & 0    & 1/8 & 1/8 & $\epsilon$ & 1/503    & 0.20 & +0.00 & -- \\
RHour & 0.05 & 1/8 & 1/8 & $\epsilon$  & 1/498    & 0.20 & -0.07 & -- \\
RHour & 0.3  & 1/8 & 1/8 & $\epsilon$  & 7/498    & 1.41 & -0.01 & {[}-0.01, +0.00{]} \\
RHour & 0.7  & 1/8 & 1/8 & $\epsilon$  & 18/513   & 3.51 & -0.01 & {[}-0.02, -0.00{]} \\
\bottomrule
\end{tabular}
\caption{\textbf{Grid/bootstrapped threshold search on Qwen2.5--1.5B stored evaluation outputs (best row per root).}
Each row summarizes the top-ranked threshold setting for the corresponding domain$\times$temperature root when running with 500 bootstrap draws.
``events/pairs'' counts flagged formal pairs out of all (problem, step) pairs in that root, and ``prev'' is the corresponding percentage.
``gain'' is the mean gain at shifted traces (pp) over flagged pairs, and ``CI'' is the 95\% bootstrap percentile interval.
}
\label{tab:formal-threshold-search-q15b}
\vspace{-4mm}
\end{table*}

\paragraph{Takeaway.}
Across these stored outputs, the selected thresholds yield extremely low event prevalence, and gains are generally small, unstable, or negative.
In particular, for Math at $T\in\{0.05,0.3\}$ the best available configurations (under this search) have \emph{negative} bootstrap lower bounds, indicating no robust evidence that shifted traces outperform the baseline on the flagged pairs.

\subsection{Qwen-7B and Llama-8B Regressions}
\label{app:qwen-llama}

We extend the main-text analysis to probe the role of model \emph{family} and \emph{size}.
Replicating the raw-effect analyses for \textbf{Qwen2.5--7B} and \textbf{Llama\,3.1--8B} on \textsc{Math}, we observe the same qualitative pattern reported for \textbf{Qwen2.5--1.5B}: mid-trace reasoning shifts are consistently detrimental across training steps and remain negative across decoding temperatures (magnitudes vary, not the sign), matching Fig.~\ref{fig:raw-effect-overlay-7b8b} and Table~\ref{tab:rs-7b8b}.

\begin{figure}[t]
  \centering
  \includegraphics[width=\linewidth]{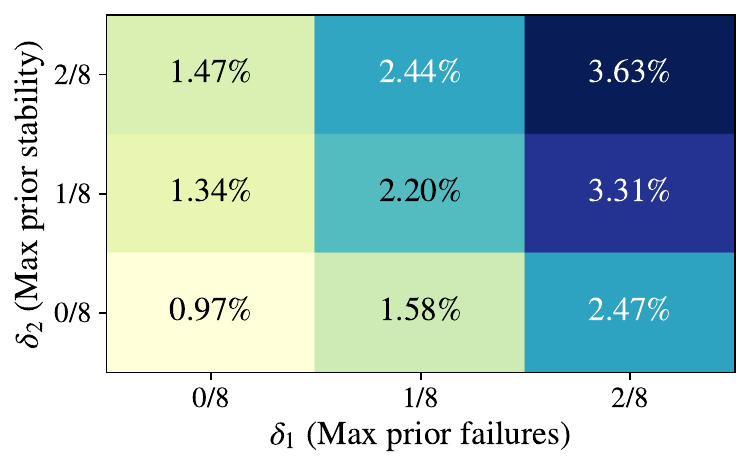}
  \caption{\textbf{Prevalence of formal ``Aha!'' events for Qwen--7B and Llama--8B (Math, $T{=}0.7$).}
  Each cell shows the fraction (and count) of problem--checkpoint pairs $(q_j,k)$ that satisfy Def.~\ref{def:aha-moment-lrms} under varying
  thresholds for prior failures ($\delta_1$) and prior stability ($\delta_2$), with $\delta_3=\epsilon>0$. Even under lenient settings,
  formal ``Aha!'' events are exceedingly rare. Per-temperature breakdowns appear in App.~\ref{sec:app-formal-aha-temp}.}
  \label{fig:aha-heatmap-overall-7b8b}
  \vspace{-5mm}
\end{figure}

We begin by checking whether the core RQ1 finding about \emph{rarity} generalizes across model family and size.
Using the same formal detector (Def.~\ref{def:aha-moment-lrms}) and threshold grid as in the main text, we compute the fraction of problem--checkpoint pairs that qualify as ``Aha!'' events.
Fig.~\ref{fig:aha-heatmap-overall-7b8b} shows that these events remain extremely sparse for both Qwen2.5--7B and Llama\,3.1--8B (\textsc{Math}, $T{=}0.7$).

\subsubsection{Step and Temperature Analysis}

\begin{figure}[t]
\centering
\begin{subfigure}[t]{\linewidth}
  \centering
  \includegraphics[width=\linewidth]{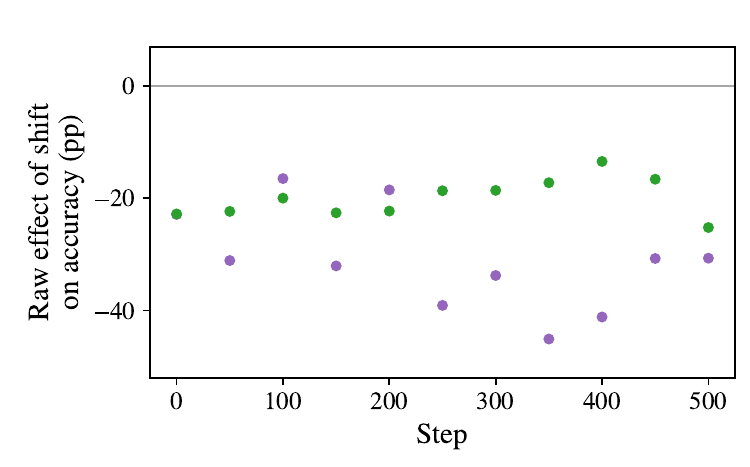}
  \caption{Raw effect by training step at $T{=}0.7$ (\textsc{Math}).}
  \label{fig:raw-effect-overlay-7b8b:a}
\end{subfigure}

\vspace{4pt}

\begin{subfigure}[t]{\linewidth}
  \centering
  \includegraphics[width=\linewidth]{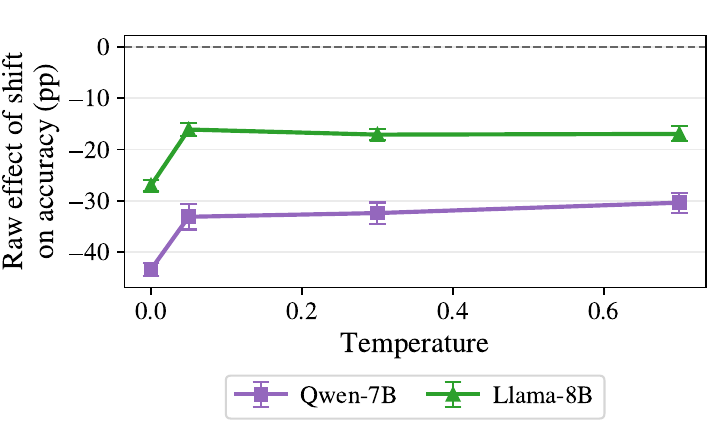}
  \caption{Raw effect vs.\ decoding temperature (\textsc{Math}).}
  \label{fig:raw-effect-overlay-7b8b:b}
\end{subfigure}

\caption{\textbf{Qwen2.5--7B vs.\ Llama\,3.1--8B on \textsc{Math}.}
Raw accuracy difference $\Delta=\widehat{p}_{Y\mid S=1}-\widehat{p}_{Y\mid S=0}$.
(a) Across training steps at $T{=}0.7$, the effect is stable and negative for both models.
(b) Across temperatures $T\in\{0.0, 0.05, 0.3, 0.7\}$ the effect remains negative; Llama\,3.1--8B exhibits a smaller penalty than Qwen2.5--7B.}
\label{fig:raw-effect-overlay-7b8b}
\vspace{-5mm}
\end{figure}

We then repeat the regression analysis from Table~\ref{tab:rs} for these models.
Figure~\ref{fig:raw-effect-overlay-7b8b} visualizes the raw effect across training steps and decoding temperatures.

\begin{table}[h]
\centering
\small
\setlength{\tabcolsep}{4pt}
\renewcommand{\arraystretch}{1.05}
\begin{tabular*}{\columnwidth}{@{\extracolsep{\fill}} l r r r @{}}
\toprule
\multicolumn{4}{c}{\textbf{(a) Training stage (fixed $T=0.7$)}} \\
\midrule
\textbf{Metric} & \textbf{Qwen2.5--7B} & \textbf{Llama\,3.1--8B} & \textbf{Combined} \\
\midrule
$N$                       & 40{,}000 & 40{,}000  & 40{,}000  \\
$\%S$                     & 1.37     & 6.54     & 3.89     \\
$\hat{p}_{Y\mid S=1}$     & 0.3467   & 0.2709   & 0.2846   \\
$\Delta$ (pp)             & $-30.39$ & $-17.68$ & $-26.97$ \\
$\mathrm{AME}$            & $-0.0841$& $-0.0688$& $-0.1706$ \\
$p$                       & $4.38\times10^{-4}$ & $6.7\times10^{-11}$ & $5.93\times10^{-42}$ \\
\midrule
\multicolumn{4}{c}{\textbf{(b) Temperature (temps pooled, steps $\le 450$)}} \\
\midrule
\textbf{Metric} & \textbf{Qwen2.5--7B} & \textbf{Llama\,3.1--8B} & \textbf{Combined} \\
\midrule
$N$                       & 160{,}000 & 160{,}000 & 320{,}000 \\
$\%S$                     & 1.50      & 5.04      & 3.26      \\
$\hat{p}_{Y\mid S=1}$     & 0.2821    & 0.2816    & 0.2818    \\
$\Delta$ (pp)             & $-37.85$  & $-17.56$  & $-27.94$  \\
$\mathrm{AME}$            & $-0.0833$ & $-0.0529$ & $-0.1457$ \\
$p$                       & $4.89\times10^{-6}$ & $2.25\times10^{-5}$ & $2.83\times10^{-22}$ \\
\bottomrule
\end{tabular*}
\caption{\textbf{Effect of detected reasoning shifts on accuracy (Qwen2.5--7B/Llama\,3.1--8B).}
$\%S$ is shift prevalence; $\hat{p}_{Y\mid S=1}$ is accuracy among shifted traces; and $\Delta$ is the raw accuracy gap (pp) relative to non-shifted traces.
$\mathrm{AME}$ and $p$ come from Binomial(logit) regressions with problem fixed effects and cluster-robust SEs (clustered by problem).
Negative AMEs indicate that shifts reduce accuracy.}
\label{tab:rs-7b8b}
\vspace{-5mm}
\end{table}

\subsubsection{Uncertainty Analysis}
\label{app:uncertainty-analysis}

This appendix extends the main-text uncertainty analysis to larger model families on \textsc{Math}, using traces from \textbf{Qwen2.5--7B} and \textbf{Llama\,3.1--8B}.
Our goal is to test a simple hypothesis: if reasoning shifts are primarily an \emph{uncertainty response}, then shifts should become \emph{more likely} as uncertainty rises.
We operationalize uncertainty using each trace’s \emph{sequence-level entropy} and use the same GPT-derived binary shift indicator as in the main text.

\paragraph{Shift prevalence vs.\ entropy.}
For each decoding temperature $T$, we regress the shift indicator on standardized sequence entropy with problem fixed effects and cluster-robust standard errors clustered by problem:
\[
\texttt{shift} \sim \texttt{C(problem)} + \texttt{std\_entropy}.
\]
Across both model families, we again find a \emph{non-positive} association between entropy and shift prevalence.
In particular, at $T{=}0.05$ and $T{=}0.7$, a 1 SD increase in entropy significantly \emph{reduces} the odds of a detected shift (OR$_{1\sigma}{=}0.63$, $p{=}0.001294$; OR$_{1\sigma}{=}0.67$, $p{=}0.002396$), while the estimates at $T{=}0$ and $T{=}0.3$ are not distinguishable from zero.
This mirrors the smaller Qwen2.5--1.5B \textsc{Math} models: shifts are not more common in high-entropy regimes, and when a dependence is detectable, it points in the opposite direction.

\paragraph{Entropy-stratified shift effects on accuracy.}
To complement the prevalence analysis, Table~\ref{tab:shift-entropy-strata-7b8b} stratifies the \emph{raw} shift effect on correctness by entropy (high = top 20\%, low = bottom 80\%), pooling temperatures and restricting to early training steps (steps $\le\!450$).
The qualitative picture is consistent across strata: shifts are associated with \emph{lower} accuracy even within the high-entropy slice.

\begin{table}[t]
  \centering
  \small
  \setlength{\tabcolsep}{4pt}
  \renewcommand{\arraystretch}{1.05}
  \begin{tabular*}{\columnwidth}{@{\extracolsep{\fill}} l r r r @{}}
    \toprule
    \textbf{Metric} & \textbf{Qwen2.5--7B} & \textbf{Llama\,3.1--8B} & \textbf{Combined} \\
    \midrule
    \multicolumn{4}{c}{\textbf{All traces (temps pooled, steps $\le 450$)}} \\
    \midrule
    $N$           & 160{,}000 & 160{,}000 & 320{,}000 \\
    $\Delta$ (pp) & $-44.43$  & $-14.83$  & $-33.69$  \\
    $p$           & $1.32\times10^{-4}$ & $0.6973$ & $0.001725$ \\
    \midrule
    \multicolumn{4}{c}{\textbf{High entropy (top 20\%)}} \\
    \midrule
    $N$           & 32{,}000  & 31{,}757 & 63{,}763 \\
    $\Delta$ (pp) & $-22.03$  & $-8.93$  & $-10.30$ \\
    $p$           & $0.06963$ & $0.7834$ & $0.001017$ \\
    \midrule
    \multicolumn{4}{c}{\textbf{Low entropy (bottom 80\%)}} \\
    \midrule
    $N$           & 128{,}000 & 127{,}027 & 255{,}021 \\
    $\Delta$ (pp) & $-48.87$  & $-14.23$  & $-38.86$ \\
    $p$           & $1.44\times10^{-4}$ & $0.7221$ & $0.01824$ \\
    \bottomrule
  \end{tabular*}
  \caption{Entropy-stratified shift effects (\textsc{Math}, steps $\le\!450$, temps pooled).
  $\Delta$ (pp) is the raw accuracy gap $\hat p(\checkmark\mid S{=}1) - \hat p(\checkmark\mid S{=}0)$.
  $p$ is from logit(correct $\sim$ shift + problem FEs) within each stratum.}
  \label{tab:shift-entropy-strata-7b8b}
\end{table}

\paragraph{Forced reconsideration as a separate mechanism.}
Finally, Table~\ref{tab:forced-aha-math-7b8b} reports paired sample-level results for \emph{triggered reconsideration} (Pass~2).
This manipulation differs from spontaneous shifts: it explicitly prompts the model to re-evaluate.
On \textsc{Math}, forced reconsideration yields a positive gain for \textbf{Qwen2.5--7B} ($+5.97$pp) but a negative gain for \textbf{Llama\,3.1--8B} ($-4.19$pp) in this evaluation slice. We tested only on a subset given the high compute cost.

\begin{table}[t]
  \centering
  \small
  \setlength{\tabcolsep}{4pt}
  \renewcommand{\arraystretch}{1.05}
  \begin{tabular*}{\columnwidth}{@{\extracolsep{\fill}} l r r r @{}}
    \toprule
    \textbf{Metric} & \textbf{Qwen2.5--7B} & \textbf{Llama\,3.1--8B} \\
    \midrule
    $N$ & 14{,}176 & 222{,}658  \\
    $\hat p_{\text{P1}}$ & 0.5509 & 0.4416 \\
    $\hat p_{\text{P2}}$ & 0.6107 & 0.3997 \\
    $\Delta$ (pp)        & $+5.97$ & $-4.19$ \\
    wins (P2 $\uparrow$) & 2{,}156 & 27{,}106  \\
    wins (P1 $\uparrow$) & 1{,}309 & 36{,}439  \\
    \bottomrule
  \end{tabular*}
  \caption{\textbf{Forced ``Aha'' (triggered reconsideration), sample-level results on \textsc{Math}.}
  $\hat p_{\text{P1}}$ and $\hat p_{\text{P2}}$ are accuracies in baseline vs.\ forced pass; $\Delta$ is the percentage-point gain;
  ``wins'' count paired samples where one pass is correct and the other is not.}
  \label{tab:forced-aha-math-7b8b}
\end{table}

\subsection{Entropy-Gated Interventions with Multiple Cues}
\label{app:uncertainty-interventions}

To test whether the effect of artificially triggered reflection depends on the specific reconsideration cue used, we evaluate three semantically similar but lexically distinct prompts:
\begin{itemize}[leftmargin=*]
    \item \textbf{C1:} ``Hold on, this reasoning might be wrong. Let's go back and check each step carefully.''
    \item \textbf{C2:} ``Actually, this approach doesn't look correct. Let's restart and work through the solution more systematically.''
    \item \textbf{C3:} ``Wait, something is not right; we need to reconsider. Let's think this through step by step.''
\end{itemize}

For each cue, we re-run $8\times 500$ Math problems (Qwen2.5--1.5B, final checkpoint) with 1-shot decoding at $T{=}0.1$, obtaining $500$ paired baseline and cued completions per cue.
We then fit a logistic regression for each cue, controlling for baseline correctness and problem identity.\footnote{In R-style notation:
\(
\texttt{correct} \sim \texttt{entropy\_std} + \texttt{baseline\_correct} + \texttt{C(problem)}.
\)
Here \texttt{entropy\_std} is the within-domain standardized sequence-level entropy defined in \S\ref{ss:unc}.}

Across all cues, higher entropy is strongly associated with improved post-intervention accuracy.
Table~\ref{tab:cue-regressions} reports standardized entropy coefficients, unit odds ratios (raw entropy), and odds ratios for a 1 SD increase in entropy.

\begin{table}[t]
\centering
\footnotesize
\setlength{\tabcolsep}{5pt}
\renewcommand{\arraystretch}{1.05}
\begin{tabular*}{0.92\linewidth}{@{\extracolsep{\fill}} l r r r @{}}
\toprule
\textbf{Cue} & \textbf{$\beta$ (std.\ ent.)} & \textbf{OR} & \textbf{OR$_{1\sigma}$} \\
\midrule
C1 & 0.79 & 3.64 & 2.21 \\
   & {\scriptsize[0.59, 1.00]} & {\scriptsize[2.60, 5.09]} & {\scriptsize[1.80, 2.72]} \\
C2 & 0.86 & 4.32 & 2.36 \\
   & {\scriptsize[0.65, 1.07]} & {\scriptsize[3.03, 6.17]} & {\scriptsize[1.92, 2.91]} \\
C3 & 0.91 & 4.09 & 2.49 \\
   & {\scriptsize[0.71, 1.12]} & {\scriptsize[2.98, 5.62]} & {\scriptsize[2.03, 3.06]} \\
\bottomrule
\end{tabular*}
\caption{\textbf{Entropy-gated improvement under three reconsideration cues.}
$\beta$ is the coefficient on standardized entropy from a logistic regression controlling for baseline correctness and problem fixed effects; brackets give 95\% CIs.
OR is the unit odds ratio (raw entropy), and OR$_{1\sigma}$ is the odds ratio for a $1$ SD increase in entropy.}
\label{tab:cue-regressions}
\vspace{-2mm}
\end{table}

All three cues show the same qualitative pattern: a one–standard deviation increase in entropy substantially increases the odds of correctness after the reconsideration cue (2.2$\times$--2.5$\times$ across cues).
C2 yields the strongest effect, but the differences are modest, indicating that the intervention’s success is tied to \emph{uncertainty} rather than to any particular lexical phrasing.

\subsection{Reasoning Shifts at Scale}
\label{app:external-models}

To verify that our findings are not an artifact of the GRPO-tuned models studied in the main paper, we evaluate two widely discussed reasoning models---DeepSeek\textendash R1 and GPT\textendash 4o---under our shift-detection protocol.
These models have been cited as exhibiting frequent ``Aha!'' moments or dramatic mid-trace realizations \citep{deepseekai2025deepseekr1incentivizingreasoningcapability}, making them a natural stress test for our methodology.

\paragraph{Experimental setup.}
We evaluate both models with:
\begin{itemize}[leftmargin=*]
    \item group sampling with $G{=}8$ independent traces per instance (fixed decoding settings per group),
    \item decoding temps $T\in\{0, 0.05, 0.3, 0.7, 1\}$,
    \item identical prompting format (with \tagtt{think} and \tagtt{answer} tags),
    \item no system-level alterations or heuristics.
\end{itemize}
On \textsc{MATH\textendash 500} and \textsc{RHour}, each model produces $G{=}8$ samples for each of 500 instances, yielding $N{=}4{,}000$ traces per model per temperature.
On \textsc{Xwords}, we evaluate 130 instances under the same $G{=}8$ protocol, yielding $N{=}1{,}040$ traces per model per temperature (reflecting completed pass-1 outputs).

\paragraph{Shift detection.}
We use the same annotation protocol as in \S\ref{ss:rtc} and App.~\ref{app:detecting-aha}:
\begin{enumerate}[leftmargin=*]
    \item \textbf{Cue prefilter:} at least one explicit lexical cue of reconsideration (e.g., ``wait'', ``actually'', ``hold on''), using the whitelist in Table~\ref{tab:shift-cue-whitelist}.
    \item \textbf{Material revision:} GPT\textendash 4o judges whether the post-cue reasoning constitutes a genuine plan pivot (rejecting a candidate, switching method, resolving a contradiction), returning a strict JSON verdict.
    \item Cases lacking either (A) lexical cue or (B) structural revision are labeled as \textbf{no shift}.
\end{enumerate}

\paragraph{Results.}
Table~\ref{tab:external-models-all} reports reasoning shift prevalence and conditional accuracy by decoding temperature across three domains.
Under our strict definition (lexical cue + material plan revision), shifts are \emph{non-trivial but still minority} events across both external models.
On \textsc{MATH\textendash 500}, DeepSeek\textendash R1 shifts occur in roughly $4.70$--$5.17\%$ of traces, while GPT\textendash 4o ranges from $1.45$--$4.40\%$ depending on $T$.
On \textsc{Xwords}, canonical shift rates are lower for DeepSeek\textendash R1 ($0.96$--$1.83\%$) and moderate for GPT\textendash 4o ($2.98$--$7.31\%$).
On \textsc{RHour}, GPT\textendash 4o exhibits $1.45$--$3.12\%$ shift rates across temperatures; DeepSeek\textendash R1 \textsc{RHour} results are not yet available.

Crucially, shifts do not yield a consistent accuracy benefit: for most model--domain--temperature settings, $P(\checkmark\mid S{=}1)$ is below $P(\checkmark\mid S{=}0)$.
Even in the closest-to-parity case (GPT\textendash 4o on \textsc{MATH\textendash 500} at $T{=}0.3$), shifted traces remain slightly less accurate ($P(\checkmark\mid S{=}1){=}0.632$ vs.\ $P(\checkmark\mid S{=}0){=}0.649$), and the direction of the gap is not stable across temperatures.
Overall, we do not observe a temperature regime in which shifts reliably improve correctness.

\begin{table*}[t]
\centering
\footnotesize
\setlength{\tabcolsep}{3.5pt}
\begin{tabular*}{\textwidth}{@{\extracolsep{\fill}} l l c r r r r r r @{}}
\toprule
\textbf{Domain} & \textbf{Model} & \textbf{$T$} &
\makecell{\textbf{\#}\\\textbf{Problems}} &
\textbf{$G$} &
\makecell{\textbf{\#}\\\textbf{Traces}} &
\makecell{\textbf{\% Shifts}\\\textbf{(count)}} &
{\scriptsize \(\mathbf{P(\checkmark\mid S{=}0)}\)} &
{\scriptsize \(\mathbf{P(\checkmark\mid S{=}1)}\)} \\
\midrule
\multicolumn{9}{@{}l}{\textsc{MATH\textendash 500}} \\
\midrule
& DeepSeek\textendash R1 & 0    & 500 & 8 & 4{,}000 & 4.70\% (188) & 0.559 & 0.245 \\
& DeepSeek\textendash R1 & 0.05 & 500 & 8 & 4{,}000 & 5.17\% (207) & 0.550 & 0.169 \\
& DeepSeek\textendash R1 & 0.3  & 500 & 8 & 4{,}000 & 5.10\% (204) & 0.534 & 0.211 \\
& DeepSeek\textendash R1 & 0.7  & 500 & 8 & 4{,}000 & 4.85\% (194) & 0.529 & 0.196 \\
& DeepSeek\textendash R1 & 1    & 500 & 8 & 4{,}000 & 5.10\% (204) & 0.532 & 0.172 \\
\addlinespace[1mm]
& GPT\textendash 4o      & 0    & 500 & 8 & 4{,}000 & 4.40\% (176) & 0.656 & 0.267 \\
& GPT\textendash 4o      & 0.05 & 500 & 8 & 4{,}000 & 2.02\% (81)  & 0.688 & 0.210 \\
& GPT\textendash 4o      & 0.3  & 500 & 8 & 4{,}000 & 2.17\% (87)  & 0.649 & 0.632 \\
& GPT\textendash 4o      & 0.7  & 500 & 8 & 4{,}000 & 1.45\% (58)  & 0.639 & 0.293 \\
& GPT\textendash 4o      & 1    & 500 & 8 & 4{,}000 & 3.45\% (138) & 0.643 & 0.304 \\
\midrule
\multicolumn{9}{@{}l}{\textsc{Xwords}} \\
\midrule
& DeepSeek\textendash R1 & 0    & 130 & 8 & 1{,}040 & 1.54\% (16)  & 0.677 & 0.688 \\
& DeepSeek\textendash R1 & 0.05 & 130 & 8 & 1{,}040 & 1.63\% (17)  & 0.703 & 0.235 \\
& DeepSeek\textendash R1 & 0.3  & 130 & 8 & 1{,}040 & 0.96\% (10)  & 0.799 & 0.400 \\
& DeepSeek\textendash R1 & 0.7  & 130 & 8 & 1{,}040 & 1.06\% (11)  & 0.797 & 0.545 \\
& DeepSeek\textendash R1 & 1    & 130 & 8 & 1{,}040 & 1.83\% (19)  & 0.793 & 0.526 \\
\addlinespace[1mm]
& GPT\textendash 4o      & 0    & 130 & 8 & 1{,}040 & 6.63\% (69)  & 0.733 & 0.377 \\
& GPT\textendash 4o      & 0.05 & 130 & 8 & 1{,}040 & 6.35\% (66)  & 0.716 & 0.439 \\
& GPT\textendash 4o      & 0.3  & 130 & 8 & 1{,}040 & 6.44\% (67)  & 0.725 & 0.373 \\
& GPT\textendash 4o      & 0.7  & 130 & 8 & 1{,}040 & 7.31\% (76)  & 0.698 & 0.421 \\
& GPT\textendash 4o      & 1    & 130 & 8 & 1{,}040 & 2.98\% (31)  & 0.650 & 0.581 \\
\midrule
\multicolumn{9}{@{}l}{\textsc{RHour}} \\
\midrule
& GPT\textendash 4o      & 0    & 500 & 8 & 4{,}000 & 3.12\% (125) & 0.004 & 0.000 \\
& GPT\textendash 4o      & 0.05 & 500 & 8 & 4{,}000 & 1.45\% (58)  & 0.005 & 0.000 \\
& GPT\textendash 4o      & 0.3  & 500 & 8 & 4{,}000 & 2.17\% (87)  & 0.008 & 0.000 \\
& GPT\textendash 4o      & 0.7  & 500 & 8 & 4{,}000 & 2.40\% (96)  & 0.007 & 0.000 \\
& GPT\textendash 4o      & 1    & 500 & 8 & 4{,}000 & 2.73\% (109) & 0.005 & 0.000 \\
& DeepSeek\textendash R1 & 0    & 500 & 8 & 4{,}000 & 11.45\% (458) & 0.000 & 0.000 \\
& DeepSeek\textendash R1 & 0.05 & 500 & 8 & 4{,}000 & 11.25\% (450) & 0.000 & 0.000 \\
& DeepSeek\textendash R1 & 0.3  & 500 & 8 & 4{,}000 & 10.53\% (421) & 0.000 & 0.000 \\
& DeepSeek\textendash R1 & 0.7  & 500 & 8 & 4{,}000 & 10.95\% (438) & 0.000 & 0.000 \\
& DeepSeek\textendash R1 & 1    & 500 & 8 & 4{,}000 & 10.82\% (433) & 0.000 & 0.000 \\

\bottomrule
\end{tabular*}
\caption{\textbf{External models across domains.}
Canonical shift rates and conditional accuracy by decoding temperature on \textsc{MATH\textendash 500}, \textsc{Xwords}, and \textsc{RHour}.
Each instance is sampled $G{=}8$ times per temperature; \textsc{MATH\textendash 500} and \textsc{RHour} use 500 instances ($N{=}4{,}000$ traces per model per $T$), while \textsc{Xwords} uses 130 instances ($N{=}1{,}040$ traces per model per $T$), matching completed pass-1 outputs.
We report shift prevalence (\% Shifts), accuracy on non-shifted traces $P(\checkmark\mid S{=}0)$, and accuracy on shifted traces $P(\checkmark\mid S{=}1)$.}
\label{tab:external-models-all}
\end{table*}

\paragraph{Interpretation.}
These results reinforce two conclusions:
\begin{enumerate}[leftmargin=*]
    \item \textbf{Reasoning shifts remain a minority behavior even in heavily discussed ``reasoning'' models.}
    Across three domains and a five-temperature sweep, only a small fraction of traces satisfy our strict definition (lexical cue + material plan revision), with rates ranging from roughly $\sim\!1\%$ on \textsc{Xwords} (DeepSeek\textendash R1) to a few percent on \textsc{MATH\textendash 500} and \textsc{RHour}.
    \item \textbf{Reasoning shifts do not reliably improve accuracy.}
    Conditional accuracy under a shift is typically \emph{lower} than the non-shift baseline across domains and temperatures, and we do not observe a stable regime in which shifts consistently improve correctness.
\end{enumerate}

\paragraph{Data release.}
We release the external-model outputs and shift annotations used in this analysis on Hugging Face, mirroring the per-temperature splits used throughout this appendix (Table~\ref{tab:external-datasets}).
DeepSeek\textendash R1 \textsc{RHour} runs are not yet included.

\begin{table}[t]
  \centering
  \footnotesize
  \setlength{\tabcolsep}{4pt}
  \renewcommand{\arraystretch}{1.05}
  \begin{tabular}{@{}l l l@{}}
    \toprule
    \textbf{Domain} & \textbf{Model} & \textbf{Dataset (Hugging Face)} \\
    \midrule
    \textsc{MATH-500} & GPT\textendash 4o      & \texttt{gpt4o-math500-t0} \\
    \textsc{MATH-500} & GPT\textendash 4o      & \texttt{gpt4o-math500-t005} \\
    \textsc{MATH-500} & GPT\textendash 4o      & \texttt{gpt4o-math500-t03} \\
    \textsc{MATH-500} & GPT\textendash 4o      & \texttt{gpt4o-math500-t07} \\
    \textsc{MATH-500} & GPT\textendash 4o      & \texttt{gpt4o-math500-t1} \\
    \textsc{MATH-500} & DeepSeek\textendash R1 & \texttt{deepseek-r1-math500-t0} \\
    \textsc{MATH-500} & DeepSeek\textendash R1 & \texttt{deepseek-r1-math500-t005} \\
    \textsc{MATH-500} & DeepSeek\textendash R1 & \texttt{deepseek-r1-math500-t03} \\
    \textsc{MATH-500} & DeepSeek\textendash R1 & \texttt{deepseek-r1-math500-t07} \\
    \textsc{MATH-500} & DeepSeek\textendash R1 & \texttt{deepseek-r1-math500-t1} \\
    \midrule
    \textsc{Xwords}   & GPT\textendash 4o      & \texttt{gpt4o-xwords-t0} \\
    \textsc{Xwords}   & GPT\textendash 4o      & \texttt{gpt4o-xwords-t005} \\
    \textsc{Xwords}   & GPT\textendash 4o      & \texttt{gpt4o-xwords-t03} \\
    \textsc{Xwords}   & GPT\textendash 4o      & \texttt{gpt4o-xwords-t07} \\
    \textsc{Xwords}   & GPT\textendash 4o      & \texttt{gpt4o-xwords-t1} \\
    \textsc{Xwords}   & DeepSeek\textendash R1 & \texttt{deepseek-r1-xwords-t0} \\
    \textsc{Xwords}   & DeepSeek\textendash R1 & \texttt{deepseek-r1-xwords-t005} \\
    \textsc{Xwords}   & DeepSeek\textendash R1 & \texttt{deepseek-r1-xwords-t03} \\
    \textsc{Xwords}   & DeepSeek\textendash R1 & \texttt{deepseek-r1-xwords-t07} \\
    \textsc{Xwords}   & DeepSeek\textendash R1 & \texttt{deepseek-r1-xwords-t1} \\
    \midrule
    \textsc{RHour}    & GPT\textendash 4o      & \texttt{gpt4o-rhour-t0} \\
    \textsc{RHour}    & GPT\textendash 4o      & \texttt{gpt4o-rhour-t005} \\
    \textsc{RHour}    & GPT\textendash 4o      & \texttt{gpt4o-rhour-t03} \\
    \textsc{RHour}    & GPT\textendash 4o      & \texttt{gpt4o-rhour-t07} \\
    \textsc{RHour}    & GPT\textendash 4o      & \texttt{gpt4o-rhour-t1} \\
    \textsc{RHour}    & DeepSeek\textendash R1 & \texttt{deepseek-r1-rhour-t0} \\
    \textsc{RHour}    & DeepSeek\textendash R1 & \texttt{deepseek-r1-rhour-t005} \\
    \textsc{RHour}    & DeepSeek\textendash R1 & \texttt{deepseek-r1-rhour-t03} \\
    \textsc{RHour}    & DeepSeek\textendash R1 & \texttt{deepseek-r1-rhour-t07} \\
    \textsc{RHour}    & DeepSeek\textendash R1 & \texttt{deepseek-r1-rhour-t1} \\
    \bottomrule
  \end{tabular}
  \caption{\textbf{Released external-model outputs across domains.} Hugging Face datasets containing the traces used in App.~\S\ref{app:external-models}, for $T\in\{0,0.05,0.3,0.7,1\}$. All dataset IDs are under the namespace \texttt{od2961/}.}
  \label{tab:external-datasets}
  \vspace{-3mm}
\end{table}

\subsection{Alternate Shift Detectors}
\label{app:shift-detector-rationale}

Prior work shows that superficial linguistic markers of hesitation---such as “wait,” “hold on,” or “actually”---are unreliable indicators of genuine cognitive shifts.
Keyword-based detectors misclassify such cues at high rates, often interpreting hedges or verbosity as insight-like events \citep{zheng-etal-2023-chain, xia2025reasoneval}.
Recent analyses of “Aha!”-style behavior in LLMs similarly report that many mid-trace cues reflect shallow self-correction or filler language rather than substantive plan changes \citep{yang2025understandingahamomentsexternal}.

In parallel, LLM-as-a-judge evaluations are known to exhibit position, ordering, and verbosity biases unless structured and controlled \citep{Wang2024FairEval, shi2024positionbias, li-etal-2024-split}.
Because our primary shift detector uses an LLM-as-judge, it is important to verify that conclusions do not depend on the specific annotation mechanism.

\paragraph{Detector variants.}
We replicate the full RQ1 analysis using three detectors:
(i) a strict formal “Aha!” criterion (Def.~\ref{def:aha-moment-lrms}),
(ii) our rubric-guided GPT-based shift detector used in the main text, and
(iii) a permissive lexical-only detector that flags any cue-phrase occurrence.
Table~\ref{tab:alt-shift-detectors} summarizes results for Qwen2.5--1.5B at $T=0.7$.

\begin{enumerate}[leftmargin=*]
    \item \textbf{Formal Aha (\texttt{formal}).} The strict criterion in Def.~\ref{def:aha-moment-lrms}, which requires (i) prior failure, (ii) prior stability, and (iii) a performance gain on traces with a detected shift.
    \item \textbf{GPT-based shifts (\texttt{gpt}).} GPT-4o marks a shift when it observes an explicit cue of reconsideration together with a material change in reasoning strategy (App.~\ref{app:detecting-aha}).
    \item \textbf{Lexical-only shifts (\texttt{words}).} A looser detector that flags a shift whenever the \tagtt{think} trace contains at least one cue phrase from our whitelist, regardless of whether the subsequent reasoning reflects a genuine plan pivot.
\end{enumerate}

\paragraph{Metrics.}
For each detector, domain, and item we compute:
(i) shift prevalence \(\%S\),
(ii) accuracies $\hat p_{Y\mid S=1}$ and $\hat p_{Y\mid S=0}$,
(iii) the raw accuracy difference $\Delta\% = 100\cdot(\hat p_{Y\mid S=1}-\hat p_{Y\mid S=0})$ (percentage points),
and (iv) the average marginal effect (AME) of a shift from a logistic regression with problem fixed effects and cluster-robust SEs (shown with $p$-value).

\begin{table*}[t]
\centering
\small
\setlength{\tabcolsep}{4pt}
\renewcommand{\arraystretch}{1.02}
\begin{tabular*}{\textwidth}{@{\extracolsep{\fill}} ll r r r r r @{}}
\toprule
\textbf{Domain} & \textbf{Detector} & \(\%S\) & $\hat p_{Y\mid S=1}$ & $\hat p_{Y\mid S=0}$ & $\Delta\%$ & $\mathrm{AME}$ \;($p$) \\
\midrule
\textbf{Xword}
 & formal & 0.0008 & 0.0000 & 0.1181 & $-11.81$ & $-0.1181$ \;(0) \\
 & gpt    & 0.0010 & 0.0400 & 0.1181 & $-7.81$  & $-0.0651$ \;(0.05095) \\
 & words  & 0.0013 & 0.0312 & 0.1182 & $-8.69$  & $-0.0712$ \;(0.04761) \\
\midrule
\textbf{Math}
 & formal & 0.0008 & 0.0215 & 0.3006 & $-27.91$ & $+0.0275$ \;(0.8201) \\
 & gpt    & 0.0030 & 0.1622 & 0.3008 & $-13.87$ & $-0.1086$ \;($7.80\times10^{-6}$) \\
 & words  & 0.0120 & 0.2606 & 0.3009 & $-4.03$  & $-0.0469$ \;(0.002153) \\
\midrule
\textbf{RHour}
 & formal & 0.0023 & 0.0000 & 0.0001 & $-0.01$  & $-0.0001$ \;(0) \\
 & gpt    & 0.0026 & 0.0000 & 0.0001 & $-0.01$  & $-0.0001$ \;(0) \\
 & words  & 0.0060 & 0.0000 & 0.0001 & $-0.01$  & $-0.0001$ \;(0) \\
\bottomrule
\end{tabular*}
\caption{\textbf{Alternative shift detectors (Qwen2.5--1.5B, $T{=}0.7$).}
Across all three detectors, shifts are rare and do not yield higher accuracy.}
\label{tab:alt-shift-detectors}
\vspace{-3mm}
\end{table*}

\paragraph{Takeaways.}
Two patterns are consistent across domains:
\begin{enumerate}[leftmargin=*]
    \item \textbf{Shifts are rare under every detector.} Even the most permissive lexical detector (\texttt{words}) identifies shifts in at most $1.2\%$ of Math traces and $0.6\%$ of RHour traces; the formal \texttt{Aha} criterion is stricter still.
    \item \textbf{Shifts are non-beneficial to accuracy.} Raw differences $\Delta\%$ and AMEs are non-positive across domains and detectors, with the only exception being \textsc{Math} under the strict \texttt{formal} detector, where the estimate is small and statistically indistinguishable from zero ($p{=}0.82$). In \textsc{Math}, both the GPT-based and lexical detectors show statistically significant negative AMEs.
\end{enumerate}

Overall, this robustness check confirms that our main RQ1 conclusion does not depend on the specific shift detector: whether we use the strict formal \texttt{Aha} definition, the rubric-guided GPT detector, or a lexical cue heuristic, mid-trace shifts are rare and generally \emph{harm} correctness rather than help it.

\section{Supplementary Figures \& Tables}
\label{app:temp-ablations}

\paragraph{Overview.}
This appendix collects supplementary tables and figures that expand the main-text analyses and document additional aggregations that are referenced in our scripts but not surfaced elsewhere in the paper.
We provide: (i) training-stage regressions at fixed decoding temperatures (beyond the $T{=}0.7$ slice in the main text), (ii) temperature sweeps for the stricter \emph{formal} ``Aha!'' detector, (iii) analogous temperature/stage breakdowns for larger models (Qwen2.5--7B and Llama\,3.1--8B) on \textsc{Math}, and (iv) additional uncertainty-gated intervention summaries, including pooled Qwen-1.5B and 7B/8B entropy-regression results.
All tables use the same conventions as the main text: \(\%S\) is shift prevalence, \(\Delta\mathrm{pp}\) denotes a raw accuracy difference in percentage points, and AMEs/coefficients come from Binomial(logit) models with problem fixed effects and cluster-robust SEs (clustered by problem).

\subsection{Training-stage effects at other decoding temperatures}
\label{sec:app-rs-temp}

Table~\ref{tab:rs_stage_T0005T03} replicates the training-stage analysis from
Table~\ref{tab:rs}, holding the decoding temperature fixed at
\(T\in\{0.0,0.05,0.3\}\).
Across these settings, we again find no evidence that reasoning shifts become beneficial later in training.
In \emph{Math}, shifts are consistently harmful across all temperatures.
In \emph{RHour}, accuracies are near zero for both shifted and non-shifted traces, and the estimated effects are practically negligible.

\begin{table}[t]
\centering
\small
\setlength{\tabcolsep}{4pt}
\renewcommand{\arraystretch}{1.05}
\begin{tabular}{lrrr}
\toprule
\multicolumn{4}{c}{\textbf{Training stage at fixed decoding temperature $T=0.0$}} \\
\midrule
\textbf{Metric} & \textbf{Xword} & \textbf{Math} & \textbf{RHour} \\
\midrule
$N$                   & 20{,}800 & 80{,}000 & 80{,}000 \\
$\%S$                 & 0.947    & 1.866    & 14.679  \\
$\hat{p}_{Y\mid S=1}$ & 0.3655   & 0.0683   & 0.0000  \\
$\Delta\mathrm{pp}$   & $+28.24$ & $-23.64$ & $-0.04$ \\
$\mathrm{AME}$        & 0.0027   & $-0.0044$& $-0.0001$\\
$p$                   & $6.89\times10^{-35}$ & $1.19\times10^{-67}$ & 0.999 \\

\midrule
\multicolumn{4}{c}{\textbf{Training stage at fixed decoding temperature $T=0.05$}} \\
\midrule
\textbf{Metric} & \textbf{Xword} & \textbf{Math} & \textbf{RHour} \\
\midrule
$N$                   & 20{,}800 & 80{,}000 & 80{,}000 \\
$\%S$                 & 0.851    & 1.854    & 15.386  \\
$\hat{p}_{Y\mid S=1}$ & 0.3390   & 0.1382   & 0.0000  \\
$\Delta\mathrm{pp}$   & $+25.41$ & $-18.56$ & $-0.05$ \\
$\mathrm{AME}$        & 0.0022   & $-0.0034$& $-0.0001$\\
$p$                   & $1.94\times10^{-26}$ & $2.0\times10^{-47}$ & 0.999 \\

\midrule
\multicolumn{4}{c}{\textbf{Training stage at fixed decoding temperature $T=0.3$}} \\
\midrule
\textbf{Metric} & \textbf{Xword} & \textbf{Math} & \textbf{RHour} \\
\midrule
$N$                   & 20{,}800 & 80{,}000 & 80{,}000 \\
$\%S$                 & 0.649    & 4.696    & 15.759  \\
$\hat{p}_{Y\mid S=1}$ & 0.2593   & 0.1637   & 0.0000  \\
$\Delta\mathrm{pp}$   & $+16.28$ & $-23.01$ & $-0.01$ \\
$\mathrm{AME}$        & 0.0011   & $-0.0108$& $-0.0000$\\
$p$                   & $1.93\times10^{-9}$ & $1.58\times10^{-158}$ & 0.999 \\
\bottomrule
\end{tabular}
\caption{\textbf{Effect of detected reasoning shifts on accuracy (Qwen2.5-1.5B): training-stage analysis at fixed temperature.}
For each fixed decoding temperature \(T\in\{0.0,0.05,0.3\}\), we report the share of traces with a detected shift (\(\%S\)),
accuracy among shifted traces (\(\hat{p}_{Y\mid S=1}\)), the raw accuracy difference in percentage points (\(\Delta\mathrm{pp}\))
between shifted and non-shifted traces, and the average marginal effect (\(\mathrm{AME}\)) from a logistic regression with
problem fixed effects, a standardized training-step control, and cluster-robust SEs (clustered by problem). Negative AME values
indicate that shifted traces are less likely to be correct holding problem and training stage fixed.}
\label{tab:rs_stage_T0005T03}
\vspace{-3mm}
\end{table}

\subsection{Training-stage effects at other decoding temperatures (Qwen-7B and Llama-8B)}
\label{sec:app-rs-temp-7b8b}

Table~\ref{tab:rs_stage_T0005T03_7b8b} provides the same fixed-temperature, training-stage analysis as Table~\ref{tab:rs_stage_T0005T03}, but for larger models on \textsc{Math} (Qwen2.5--7B and Llama\,3.1--8B), evaluated over steps $\le 450$.
Across temperatures, shifts remain associated with lower accuracy; the magnitude of the raw penalty varies with $T$ and model family, but does not reverse sign.

 \begin{table}[t]
\centering
\small
\setlength{\tabcolsep}{4pt}
\renewcommand{\arraystretch}{1.05}
\begin{tabular}{lrr}
\toprule
\multicolumn{3}{c}{\textbf{Training stage at fixed decoding temperature $T=0$}} \\
\midrule
\textbf{Metric} & \textbf{Qwen2.5-7B} & \textbf{Llama3.1-8B} \\
\midrule
$N$                   & 40{,}000 & 40{,}000 \\
$\%S$                 & 2.538    & 2.418    \\
$\hat{p}_{Y\mid S=1}$ & 0.2039   & 0.1607   \\
$\Delta\mathrm{pp}$   & $-45.10$ & $-27.18$ \\
$\mathrm{AME}$        & $-0.0659$& $-0.0597$\\
$p$                   & $0.03314$& $0.04043$\\
\midrule
\multicolumn{3}{c}{\textbf{Training stage at fixed decoding temperature $T=0.05$}} \\
\midrule
\textbf{Metric} & \textbf{Qwen2.5-7B} & \textbf{Llama3.1-8B} \\
\midrule
$N$                   & 40{,}000 & 40{,}208 \\
$\%S$                 & 0.853    & 5.710    \\
$\hat{p}_{Y\mid S=1}$ & 0.3284   & 0.3319   \\
$\Delta\mathrm{pp}$   & $-34.06$ & $-14.82$ \\
$\mathrm{AME}$        & $-0.0401$& $-0.0436$\\
$p$                   & $0.07879$& $0.007971$\\
\midrule
\multicolumn{3}{c}{\textbf{Training stage at fixed decoding temperature $T=0.3$}} \\
\midrule
\textbf{Metric} & \textbf{Qwen2.5-7B} & \textbf{Llama3.1-8B} \\
\midrule
$N$                   & 40{,}000 & 40{,}192 \\
$\%S$                 & 1.248    & 5.576    \\
$\hat{p}_{Y\mid S=1}$ & 0.3387   & 0.2945   \\
$\Delta\mathrm{pp}$   & $-32.91$ & $-17.44$ \\
$\mathrm{AME}$        & $-0.0788$& $-0.0540$\\
$p$                   & $2.91\times10^{-4}$ & $4.4\times10^{-5}$ \\
\bottomrule
\end{tabular}
\caption{\textbf{Effect of detected reasoning shifts on accuracy: training-stage analysis at fixed temperature.}
For each fixed decoding temperature $T\in\{0.0,0.05,0.3\}$, we report shift prevalence (\%S), accuracy among shifted traces (\(\hat{p}_{Y\mid S=1}\)),
the raw accuracy difference in percentage points (\(\Delta\mathrm{pp}\)), and the average marginal effect (\(\mathrm{AME}\)) from a logistic regression with
problem fixed effects, a standardized training-step control, and cluster-robust SEs. Negative AME values indicate that shifted traces are less likely
to be correct holding problem and training stage fixed.}
\label{tab:rs_stage_T0005T03_7b8b}
\vspace{-3mm}
\end{table}

\subsection{Formal ``Aha!'' moments across decoding temperatures}
\label{sec:app-formal-aha-temp}

We repeat the temperature-sweep analysis using the stricter \emph{formal} ``Aha!'' detector (Def.~3.1),
which requires a mid-trace pivot \emph{and} a contemporaneous performance gain at that checkpoint.
For each decoding temperature \(T\in\{0,0.05,0.3,0.7\}\), we estimate the association between correctness
and the formal-Aha indicator while controlling for problem fixed effects and training stage (standardized step),
reporting average marginal effects (AME) with cluster-robust SEs.
Because the formal detector is extremely sparse in several regimes (and never fires for RHour at \(T\le 0.3\)),
some conditional quantities are undefined; we denote these with ``--''.

\subsection{Formal ``Aha!'' moments across decoding temperatures (Qwen-7B and Llama-8B)}
\label{sec:app-formal-aha-temp-7b8b}

Table~\ref{tab:formal-aha-temp-7b8b} repeats the formal-detector temperature sweep for larger models on \textsc{Math} (Qwen2.5--7B and Llama\,3.1--8B), evaluated over steps $\le 450$.
As in the 1.5B setting, formal ``Aha!'' detections remain extremely sparse across temperatures, and conditional estimates can be unstable.

\begin{table}[t]
\centering
\small
\setlength{\tabcolsep}{4pt}
\renewcommand{\arraystretch}{1.05}
\begin{tabular}{lrr}
\toprule
\textbf{Metric} & \textbf{Qwen2.5-7B} & \textbf{Llama3.1-8B} \\
\midrule
\multicolumn{3}{c}{\textbf{$T=0$}} \\
\midrule
$N$                   & 40{,}000 & 40{,}000 \\
$\%S$                 & 0.362    & 0.832 \\
$\hat{p}_{Y\mid S=1}$ & 0.0621   & 0.0449 \\
$\Delta\mathrm{pp}$   & $-58.35$ & $-38.42$ \\
$\mathrm{AME}$        & $+0.0541$ & $+0.0112$ \\
$p$                   & 0.6627 & 0.815 \\
\midrule
\multicolumn{3}{c}{\textbf{$T=0.05$}} \\
\midrule
$N$                   & 40{,}000 & 40{,}000 \\
$\%S$                 & 0.048    & 0.090 \\
$\hat{p}_{Y\mid S=1}$ & 0.0000   & 0.0278 \\
$\Delta\mathrm{pp}$   & $-66.64$ & $-44.42$ \\
$\mathrm{AME}$        & $-0.3735$ & $+0.1626$ \\
$p$                   & $1.84\times10^{-135441}$ & 0.2839 \\
\midrule
\multicolumn{3}{c}{\textbf{$T=0.3$}} \\
\midrule
$N$                   & 40{,}000 & 40{,}000 \\
$\%S$                 & 0.045    & 0.109 \\
$\hat{p}_{Y\mid S=1}$ & 0.0000   & 0.0000 \\
$\Delta\mathrm{pp}$   & $-66.40$ & $-45.97$ \\
$\mathrm{AME}$        & $-0.4970$ & $-0.4114$ \\
$p$                   & $8.64\times10^{-65910}$ & $3.13\times10^{-95446}$ \\
\midrule
\multicolumn{3}{c}{\textbf{$T=0.7$}} \\
\midrule
$N$                   & 40{,}000 & 40{,}000 \\
$\%S$                 & 0.022    & 0.073 \\
$\hat{p}_{Y\mid S=1}$ & 0.0000   & 0.0357 \\
$\Delta\mathrm{pp}$   & $-64.66$ & $-40.08$ \\
$\mathrm{AME}$        & $-0.5572$ & $+0.1937$ \\
$p$                   & $4.7\times10^{-54940}$ & 0.1153 \\
\bottomrule
\end{tabular}
\caption{\textbf{Formal ``Aha!'' detector (Def.~3.1): temperature sweep for Qwen2.5-7B/Llama3.1-8B.}
For each decoding temperature, \(\%S\) is the share of traces flagged by the formal detector; \(\hat{p}_{Y\mid S=1}\) is empirical accuracy among flagged traces; and \(\Delta\mathrm{pp}\) is the raw accuracy difference (percentage points) between flagged and non-flagged traces.
\(\mathrm{AME}\) is the average marginal effect from a logistic regression with problem fixed effects and a standardized training-step control.}
\label{tab:formal-aha-temp-7b8b}
\vspace{-3mm}
\end{table}

\subsection{Additional Temperature Ablations}
\label{spp:ablations}

Fig. ~\ref{fig:app-aha-grid-q15b-bytemp} provides additional temperature ablations across our suite of Qwen2.5-1.5B traces for the Xword, Math, and RHour datasets. We carry out the same analysis over our Qwen-7B and Llama-8B Math traces in Fig. ~\ref{fig:app-aha-grid-q7b-l8b-bytemp}.

\subsection{Qualitative review of formal ``Aha!'' Moments}
\label{app:qualitative-formal-aha}

Below, we show a qualitative inspection of a small set of (Formal) ``Aha!'' detections from our stored Qwen2.5--1.5B evaluation outputs. For each domain we apply the Formal criteria at the problem--checkpoint level and then show representative shifted traces.

\paragraph{Math.}
We use $(\delta_1=0.250,\,\delta_2=0.250,\,\delta_3=0.000)$ with \texttt{min\_prior\_steps}=2.

\paragraph{Xword.}
We use $(\delta_1=0.500,\,\delta_2=0.500,\,\delta_3=0.000)$ with \texttt{min\_prior\_steps}=2.

\paragraph{RHour.}
We use $(\delta_1=0.250,\,\delta_2=0.250,\,\delta_3=None)$ with \texttt{min\_prior\_steps}=2.
Because RHour accuracies are near zero in these stored outputs, we found too few events satisfying a positive gain constraint; we therefore omit the gain threshold for this qualitative inspection.

\subsection{Triggered reconsideration under uncertainty}
\label{app:pass2-entropy}

We extend \S\ref{sec:rq3-uncertainty} by analyzing when an \emph{extrinsically triggered} reconsideration cue (Pass~2) is most effective.
We report both a nonparametric entropy gate (top-20\% vs.\ bottom-80\% by pass-1 entropy) and a regression that treats entropy as a continuous predictor.

\paragraph{Entropy-gated gains (nonparametric stratification).}
For each domain, we bucket prompts by pass-1 sequence entropy using a fixed within-domain threshold at the 80th percentile (high = top 20\%, low = bottom 80\%).
We report pass-1 and pass-2 accuracies and the paired gain \(\Delta\) in percentage points.
In addition to per-domain results, we include a pooled ``ALL'' row that aggregates Xword/Math/RHour (count-weighted).

\paragraph{Entropy as a continuous predictor (regression).}
We regress pass-2 correctness on standardized pass-1 entropy, controlling for pass-1 correctness and problem fixed effects (cluster-robust SEs at the problem level).
Table~\ref{tab:pass2-entropy-regression} reports the log-odds coefficient \(\beta_{\mathrm{ent}}\) (per +1 SD entropy) and the corresponding odds ratio \(\mathrm{OR}_{1\sigma}\!=\!\exp(\beta_{\mathrm{ent}})\).

\begin{table}[h]
\centering
\small
\setlength{\tabcolsep}{4pt}
\renewcommand{\arraystretch}{1.05}
\begin{tabular*}{\columnwidth}{@{\extracolsep{\fill}} l r r r r @{}}
\toprule
\textbf{Domain} & $N$ & $\beta_{\mathrm{ent}}$ & $\mathrm{OR}_{1\sigma}$ & $p$ \\
\midrule
Xword &  99{,}840 & $-0.033$ & $0.97$ & 0.091 \\
Math  & 464{,}000 & $+0.019$ & $1.02$ & 0.146 \\
RHour & 331{,}120 & $-0.407$ & $0.67$ & $2.36\times10^{-119}$ \\
\bottomrule
\end{tabular*}
\caption{\textbf{Pass-2 accuracy vs.\ pass-1 entropy (Qwen2.5-1.5B).}
We regress pass-2 correctness on standardized pass-1 entropy, controlling for pass-1 correctness and problem fixed effects. $\beta_{\mathrm{ent}}$ is the log-odds coefficient for a 1 SD entropy increase and $\mathrm{OR}_{1\sigma}=\exp(\beta_{\mathrm{ent}})$.}
\label{tab:pass2-entropy-regression}
\vspace{-3mm}
\end{table}

\paragraph{Pass-2 entropy regression for larger models.}
Table~\ref{tab:pass2-entropy-regression-7b8b} reports the same regression for Qwen2.5--7B and Llama\,3.1--8B on \textsc{Math}.
Here, entropy has a small and non-significant association for Qwen2.5--7B, while for Llama\,3.1--8B the association is negative and statistically detectable.

\begin{table}[h]
\centering
\small
\setlength{\tabcolsep}{4pt}
\renewcommand{\arraystretch}{1.05}
\begin{tabular*}{\columnwidth}{@{\extracolsep{\fill}} l r r r r @{}}
\toprule
\textbf{Group} & $N$ & $\beta_{\mathrm{ent}}$ & $\mathrm{OR}_{1\sigma}$ & $p$ \\
\midrule
Qwen2.5-7B  & 63{,}404  & $+0.012$ & $1.01$ & 0.7586 \\
Llama3.1-8B & 102{,}232 & $-0.075$ & $0.93$ & 0.005146 \\
\bottomrule
\end{tabular*}
\caption{\textbf{Pass-2 accuracy vs.\ pass-1 entropy (Qwen2.5-7B/Llama3.1-8B).}
We regress pass-2 correctness on standardized pass-1 entropy, controlling for pass-1 correctness and problem fixed effects (cluster-robust SEs).}
\label{tab:pass2-entropy-regression-7b8b}
\vspace{-3mm}
\end{table}

\begin{table}[t]
\centering
\small
\setlength{\tabcolsep}{4pt}
\renewcommand{\arraystretch}{1.05}
\begin{tabular}{lrrr}
\toprule
\textbf{Metric} & \textbf{Crossword} & \textbf{Math} & \textbf{RHour} \\
\midrule

\multicolumn{4}{c}{\textbf{$T=0.0$}} \\
\midrule
$N$                   & 20{,}800 & 80{,}000 & 80{,}000 \\
$\%S$                 & 0.471    & 0.462    & 0.000 \\
$\hat{p}_{Y\mid S=1}$ & 0.0816   & 0.0000   & -- \\
$\Delta\mathrm{pp}$   & $-0.42$  & $-30.17$ & -- \\
$\mathrm{AME}$        & $-0.0000$& $-0.0014$& -- \\
$p$                   & 0.883    & 0.999    & -- \\

\midrule
\multicolumn{4}{c}{\textbf{$T=0.05$}} \\
\midrule
$N$                   & 20{,}800 & 80{,}000 & 80{,}000 \\
$\%S$                 & 0.212    & 0.299    & 0.000 \\
$\hat{p}_{Y\mid S=1}$ & 0.0000   & 0.0251   & -- \\
$\Delta\mathrm{pp}$   & $-8.72$  & $-29.62$ & -- \\
$\mathrm{AME}$        & $-0.0002$& $-0.0009$& -- \\
$p$                   & 0.999    & $1.92\times10^{-12}$ & -- \\

\midrule
\multicolumn{4}{c}{\textbf{$T=0.3$}} \\
\midrule
$N$                   & 20{,}800 & 80{,}000 & 80{,}000 \\
$\%S$                 & 0.312    & 0.475    & 0.000 \\
$\hat{p}_{Y\mid S=1}$ & 0.0000   & 0.0211   & -- \\
$\Delta\mathrm{pp}$   & $-9.78$  & $-36.36$ & -- \\
$\mathrm{AME}$        & $-0.0003$& $-0.0017$& -- \\
$p$                   & 0.999    & $4.17\times10^{-21}$ & -- \\

\midrule
\multicolumn{4}{c}{\textbf{$T=0.7$}} \\
\midrule
$N$                   & 20{,}800 & 80{,}000 & 80{,}000 \\
$\%S$                 & 1.438    & 0.364    & 8.191 \\
$\hat{p}_{Y\mid S=1}$ & 0.0067   & 0.0241   & 0.0002 \\
$\Delta\mathrm{pp}$   & $-11.22$ & $-26.17$ & $+0.01$ \\
$\mathrm{AME}$        & $-0.0016$& $-0.0010$& $0.0000$ \\
$p$                   & $2.41\times10^{-5}$ & $3.25\times10^{-13}$ & 0.461 \\

\bottomrule
\end{tabular}
\caption{\textbf{Formal ``Aha!'' detector (Def.~3.1): temperature sweep.}
For each domain and decoding temperature, \(\%S\) is the share of traces flagged by the formal detector;
\(\hat{p}_{Y\mid S=1}\) is empirical accuracy among flagged traces; and \(\Delta\mathrm{pp}\) is the raw accuracy
difference (percentage points) between flagged and non-flagged traces.
\(\mathrm{AME}\) is the average marginal effect of a formal-Aha flag from a logistic regression with problem fixed effects,
a standardized training-step control, and cluster-robust SEs.
Cells marked ``--'' indicate the detector never fired in that regime, making conditional quantities undefined.}
\label{tab:formal-aha-temp}
\vspace{-3mm}
\end{table}

\subsection{Pass-2 accuracy conditional on detected shifts (additional summary)}
\label{sec:app-pass2-conditional-on-shift}

Because our intervention defines a second pass (Pass~2), it is useful to verify that the negative association between \emph{spontaneous} shifts and correctness is not an artifact of evaluating only the first-pass answer.
Table~\ref{tab:shift-accuracy-pass2} reports, for each setting, the Pass~2 accuracy among traces whose \emph{Pass~1} reasoning was labeled as shifted vs.\ non-shifted, alongside the corresponding raw differences.

\begin{table*}[h]
\centering
\footnotesize
\setlength{\tabcolsep}{4pt}
\renewcommand{\arraystretch}{1.05}
\begin{tabular*}{\textwidth}{@{\extracolsep{\fill}} l c r r r r r @{}}
\toprule
\textbf{Experiment} & $T$ & $N$ & \(\%S\) & $P_2(\checkmark\mid S{=}1)$ & $P_2(\checkmark\mid S{=}0)$ & $\Delta_2$ (pp) \\
\midrule
Qwen2.5--1.5B (all domains) & all & 723{,}200 & 7.65 & 3.95 & 20.42 & $-16.47$ \\
\midrule
Qwen2.5--7B (Math) & 0.0  & 39{,}080 & 2.54 & 25.62 & 67.10 & $-41.47$ \\
Qwen2.5--7B (Math) & 0.05 & 2{,}768  & 0.85 & 13.64 & 64.20 & $-50.57$ \\
Qwen2.5--7B (Math) & 0.3  & 1{,}104  & 1.25 & 30.77 & 57.79 & $-27.03$ \\
Qwen2.5--7B (Math) & 0.7  & 20{,}180 & 1.37 & 45.07 & 64.73 & $-19.66$ \\
\midrule
Llama\,3.1--8B (Math) & 0.0  & 14{,}728 & 2.42 & 32.94 & 36.67 & $-3.74$ \\
Llama\,3.1--8B (Math) & 0.05 & 28{,}808 & 5.71 & 24.67 & 37.20 & $-12.53$ \\
Llama\,3.1--8B (Math) & 0.3  & 30{,}240 & 5.58 & 27.78 & 40.19 & $-12.41$ \\
Llama\,3.1--8B (Math) & 0.7  & 28{,}376 & 6.54 & 28.01 & 42.36 & $-14.35$ \\
\bottomrule
\end{tabular*}
\caption{\textbf{Pass-2 accuracy conditional on Pass-1 shift labels.}
$P_2(\checkmark\mid S{=}1)$ and $P_2(\checkmark\mid S{=}0)$ denote Pass~2 accuracies among traces whose Pass~1 reasoning was labeled as shifted vs.\ non-shifted, respectively, and $\Delta_2$ is the raw percentage-point difference.}
\label{tab:shift-accuracy-pass2}
\vspace{-3mm}
\end{table*}

\begin{table*}[t]
\centering
\setlength{\tabcolsep}{4pt}
\renewcommand{\arraystretch}{1.05}
\begin{tabular*}{\columnwidth}{@{\extracolsep{\fill}} l l r r r r @{}}
\toprule
\textbf{Domain} & \textbf{Bucket} & $N$ & $\hat p_{\text{P1}}$ (\%) & $\hat p_{\text{P2}}$ (\%) & $\Delta$ (pp) \\
\midrule
Xword & all  &  99{,}840 &  9.65 & 10.15 & +0.49 \\
Xword & high &  19{,}969 &  8.56 &  9.59 & +1.04 \\
Xword & low  &  79{,}871 &  9.93 & 10.29 & +0.36 \\
\midrule
Math  & all  & 464{,}000 & 32.70 & 40.43 & +7.74 \\
Math  & high &  92{,}800 & 19.70 & 35.09 & +15.38 \\
Math  & low  & 371{,}200 & 35.94 & 41.77 & +5.82 \\
\midrule
RHour & all  & 331{,}120 &  0.023 & 0.036 & +0.013 \\
RHour & high &  66{,}224 &  0.027 & 0.023 & -0.005 \\
RHour & low  & 264{,}896 &  0.022 & 0.039 & +0.017 \\
\midrule
Overall   & all  & 894{,}960 & 18.04 & 22.11 & +4.07 \\
Overall   & high & 178{,}993 & 11.18 & 19.27 & +8.09 \\
Overall   & low  & 715{,}967 & 19.75 & 22.82 & +3.07 \\
\bottomrule
\end{tabular*}
\caption{\textbf{Triggered reconsideration gains by pass-1 entropy.}
We bucket instances by pass-1 sequence entropy within each domain (high = top 20\%, low = bottom 80\%). ``Overall'' aggregates across domains using count-weighted averages.}
\label{tab:pass2-entropy-gate}
\vspace{-3mm}
\end{table*}

\begin{figure*}[t]
  \centering
  \small
  \setlength{\tabcolsep}{2pt}
  \renewcommand{\arraystretch}{1.05}

  \begin{tabular}{@{}M{2.2em}@{\hspace{2pt}}M{0.30\textwidth}@{\hspace{-2pt}}M{0.30\textwidth}@{\hspace{-2pt}}M{0.30\textwidth}@{}}
    & \textbf{Crossword} & \textbf{Math} & \textbf{RHour} \\
    \midrule

    \rotatebox[origin=c]{90}{\textbf{$T=0$}} &
    \subcaptionbox{}{\includegraphics[width=\linewidth]{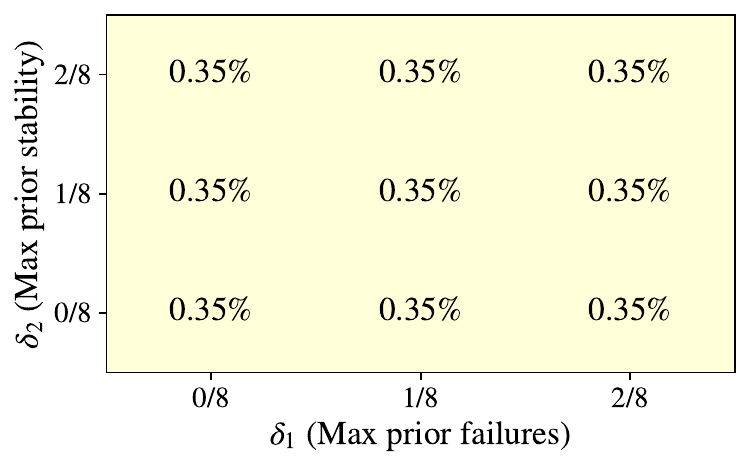}\label{fig:aha-xword-q15b-t0}} &
    \subcaptionbox{}{\includegraphics[width=\linewidth]{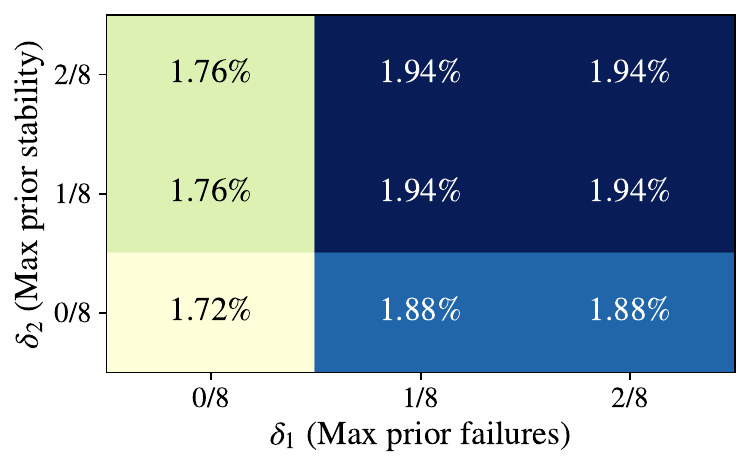}\label{fig:aha-math-q15b-t0}} &
    \subcaptionbox{}{\includegraphics[width=\linewidth]{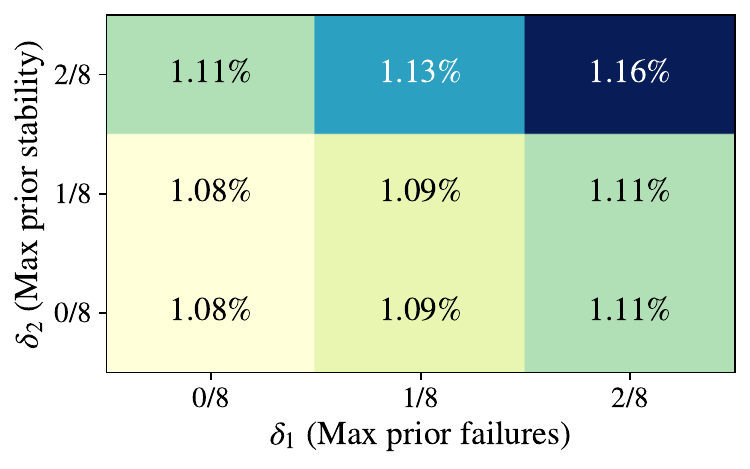}\label{fig:aha-car-q15b-t0}} \\[3pt]

    \rotatebox[origin=c]{90}{\textbf{$T=0.05$}} &
    \subcaptionbox{}{\includegraphics[width=\linewidth]{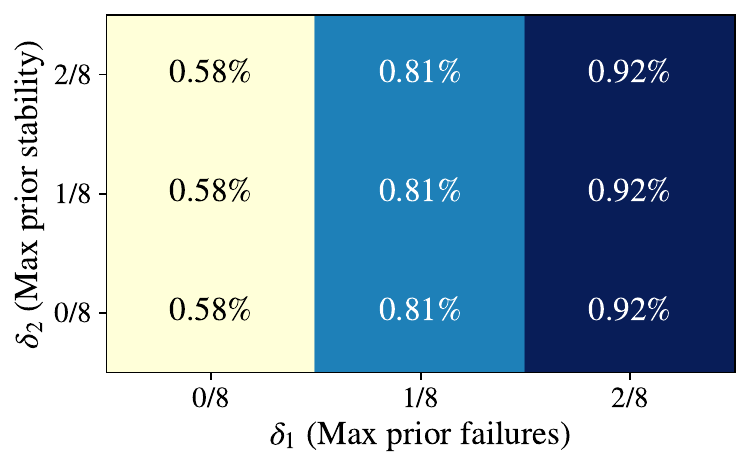}\label{fig:aha-xword-q15b-t005}} &
    \subcaptionbox{}{\includegraphics[width=\linewidth]{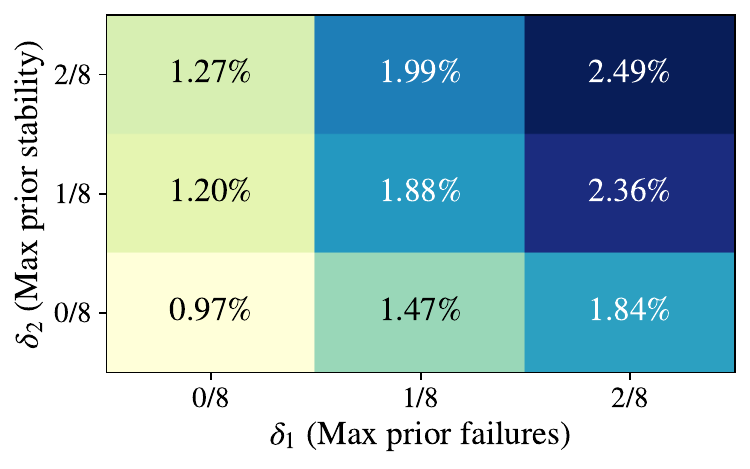}\label{fig:aha-math-q15b-t005}} &
    \subcaptionbox{}{\includegraphics[width=\linewidth]{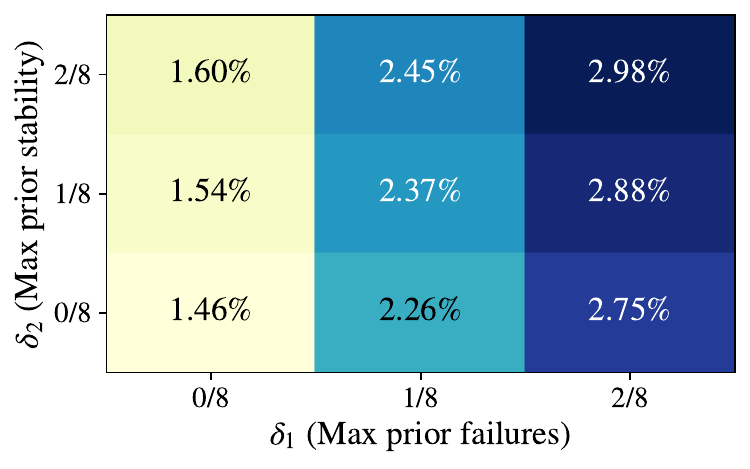}\label{fig:aha-car-q15b-t005}} \\[3pt]

    \rotatebox[origin=c]{90}{\textbf{$T=0.3$}} &
    \subcaptionbox{}{\includegraphics[width=\linewidth]{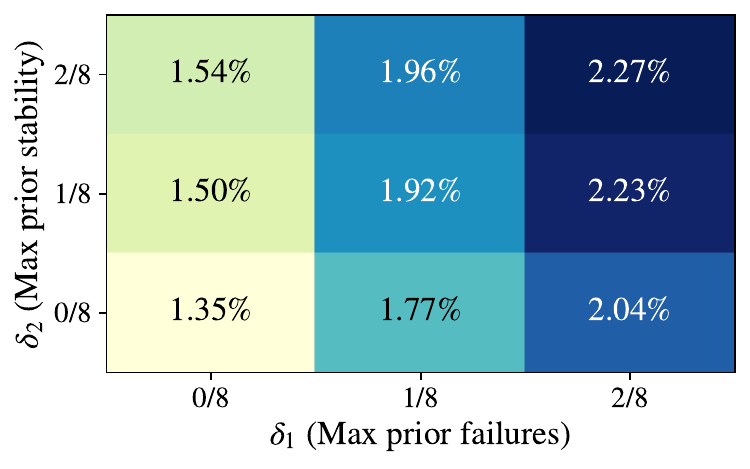}\label{fig:aha-xword-q15b-t03}} &
    \subcaptionbox{}{\includegraphics[width=\linewidth]{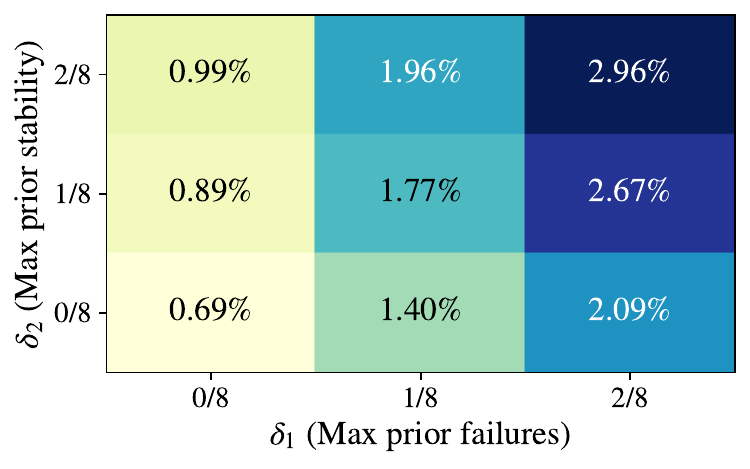}\label{fig:aha-math-q15b-t03}} &
    \subcaptionbox{}{\includegraphics[width=\linewidth]{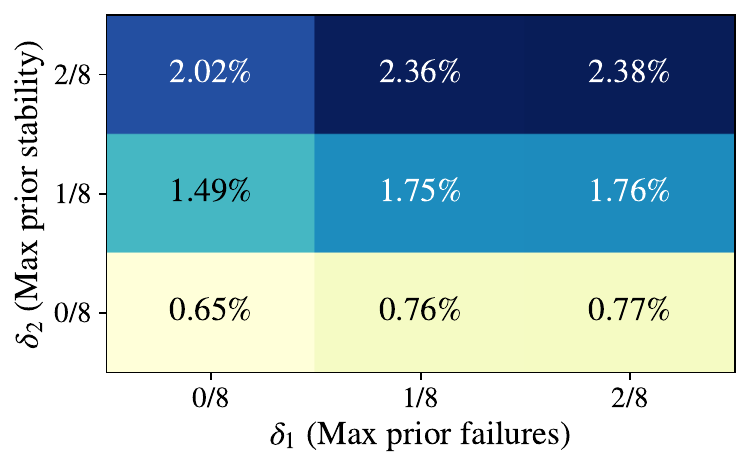}\label{fig:aha-car-q15b-t03}} \\[3pt]

    \rotatebox[origin=c]{90}{\textbf{$T=0.7$}} &
    \subcaptionbox{}{\includegraphics[width=\linewidth]{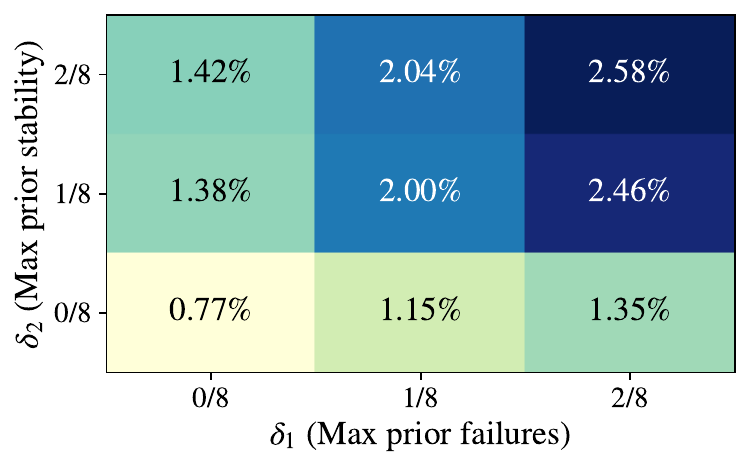}\label{fig:aha-xword-q15b-t07}} &
    \subcaptionbox{}{\includegraphics[width=\linewidth]{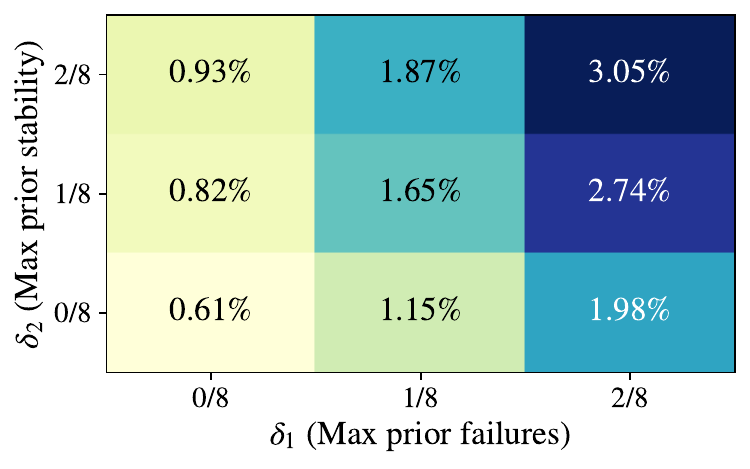}\label{fig:aha-math-q15b-t07}} &
    \subcaptionbox{}{\includegraphics[width=\linewidth]{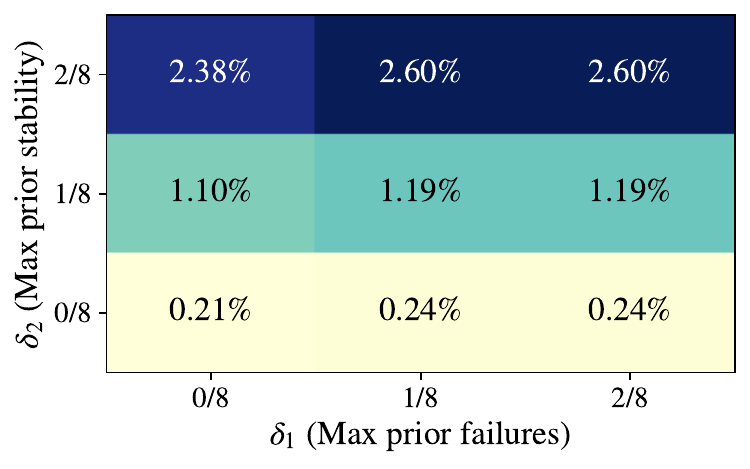}\label{fig:aha-car-q15b-t07}} \\
  \end{tabular}

  \caption{\textbf{Aha! moment prevalence heatmaps (Qwen-1.5B) across decoding temperatures.}
  Columns are domains; rows vary decoding temperature $T$.
  Cells show the share of $(q_j,k)$ pairs meeting Def.~\ref{def:aha-moment-lrms} under the threshold grid;
  see App.~\ref{app:detecting-aha} for detection details.}
  \label{fig:app-aha-grid-q15b-bytemp}
\end{figure*}

\begin{figure*}[t]
  \centering
  \small
  \setlength{\tabcolsep}{2pt}
  \renewcommand{\arraystretch}{1.05}

  \begin{tabular}{@{}M{2.2em}@{\hspace{2pt}}M{0.30\textwidth}@{\hspace{-2pt}}M{0.30\textwidth}@{}}
    & \textbf{Qwen-7B (Math)} & \textbf{Llama-8B (Math)} \\
    \midrule

    \rotatebox[origin=c]{90}{\textbf{$T=0$}} &
    \subcaptionbox{}{\includegraphics[width=\linewidth]{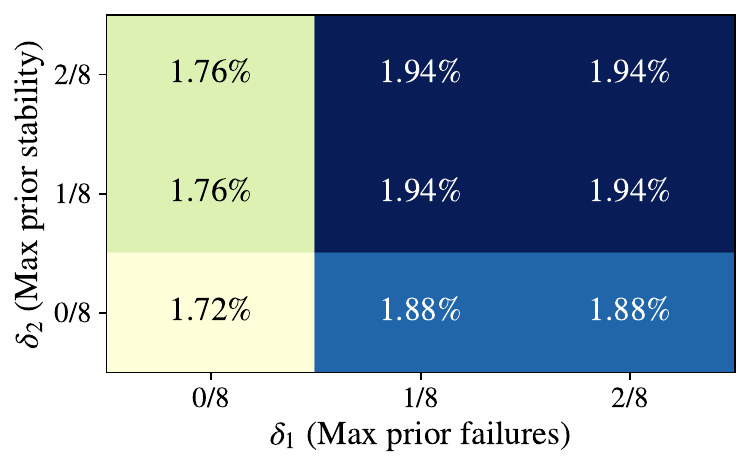}\label{fig:aha-math-q7b-t0}} &
    \subcaptionbox{}{\includegraphics[width=\linewidth]{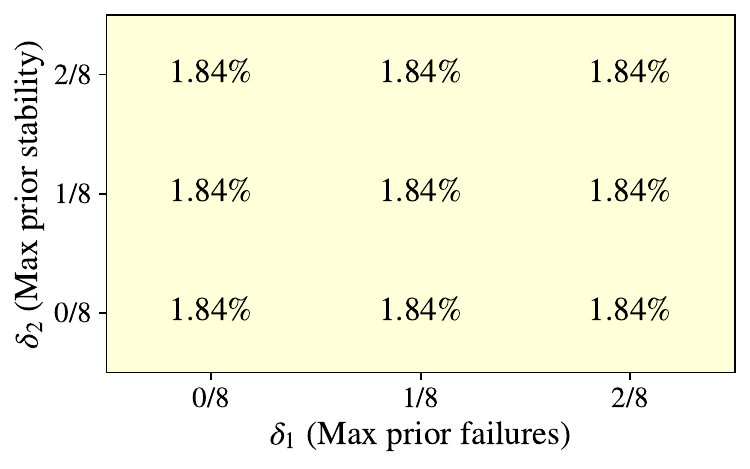}\label{fig:aha-math-l8b-t0}} \\[3pt]

    \rotatebox[origin=c]{90}{\textbf{$T=0.05$}} &
    \subcaptionbox{}{\includegraphics[width=\linewidth]{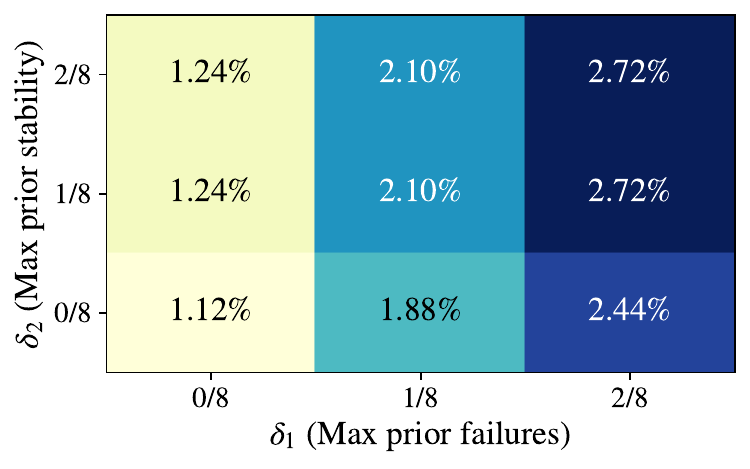}\label{fig:aha-math-q7b-t005}} &
    \subcaptionbox{}{\includegraphics[width=\linewidth]{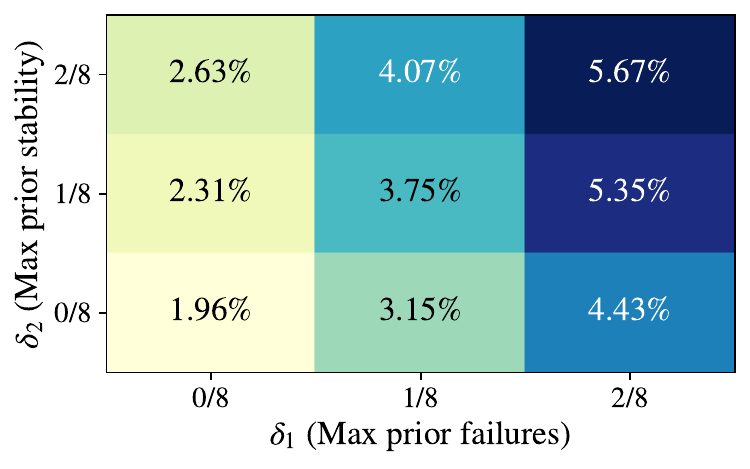}\label{fig:aha-math-l8b-t005}} \\[3pt]

    \rotatebox[origin=c]{90}{\textbf{$T=0.3$}} &
    \subcaptionbox{}{\includegraphics[width=\linewidth]{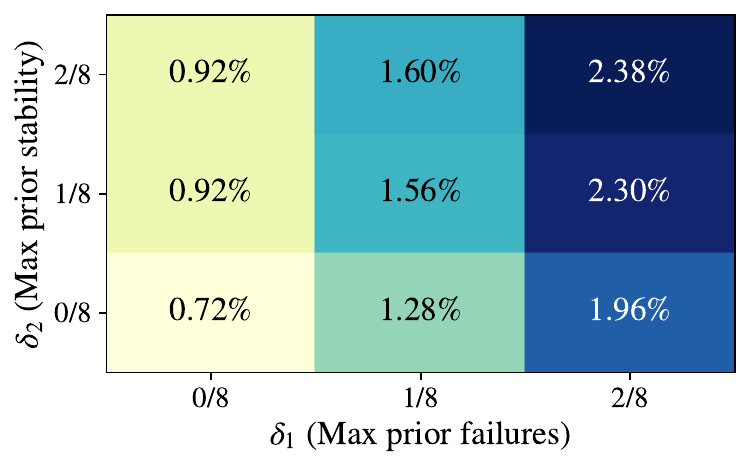}\label{fig:aha-math-q7b-t03}} &
    \subcaptionbox{}{\includegraphics[width=\linewidth]{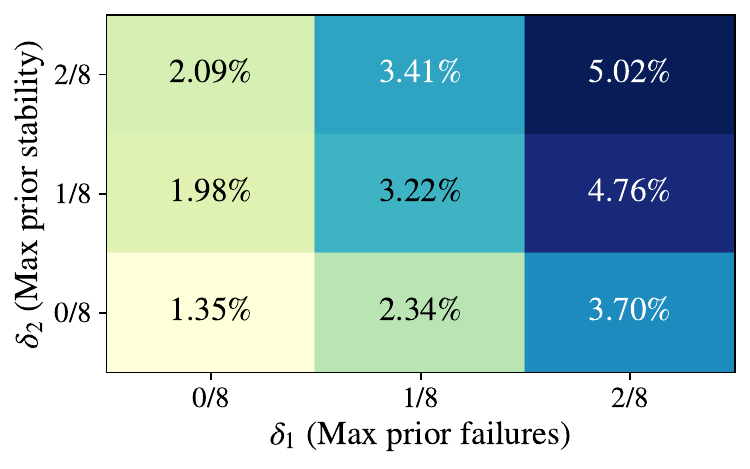}\label{fig:aha-math-l8b-t03}} \\[3pt]

    \rotatebox[origin=c]{90}{\textbf{$T=0.7$}} &
    \subcaptionbox{}{\includegraphics[width=\linewidth]{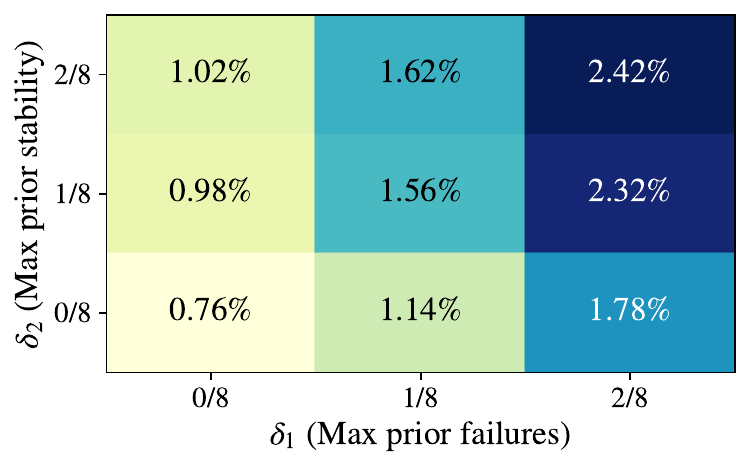}\label{fig:aha-math-q7b-t07}} &
    \subcaptionbox{}{\includegraphics[width=\linewidth]{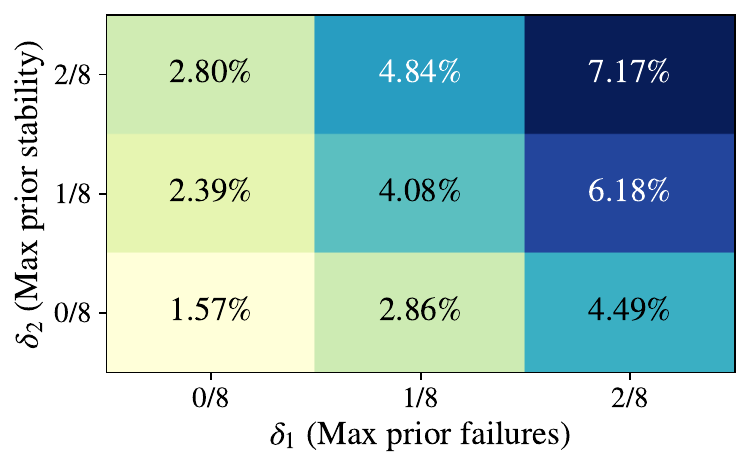}\label{fig:aha-math-l8b-t07}} \\
  \end{tabular}

  \caption{\textbf{Aha! moment prevalence heatmaps (Qwen-7B vs.\ Llama-8B; \textsc{MATH-500}) across decoding temperatures.}
  Columns are models; rows vary decoding temperature $T$.
  Cells show the share of $(q_j,k)$ pairs meeting Def.~\ref{def:aha-moment-lrms} under the threshold grid;
  see App.~\ref{app:detecting-aha} for detection details.}
  \label{fig:app-aha-grid-q7b-l8b-bytemp}
\end{figure*}

\section{Release and Artifacts}
\label{sec:artifacts}

All artifact details (contents, structure, and reproduction steps) are described in the corresponding artifact appendix sections of this document. For convenience, we provide the single entry-point link here.

\paragraph{Repository.}
The full artifact bundle (evaluation pipeline, shift-detection code, configs, and supporting documentation) can be found linked to our github repository.

\paragraph{Contact.}
For questions, bug reports, or replication issues, please use the GitHub issue tracker:
\url{https://github.com/humans-and-machines/Illusion-of-Reasoning/issues}

\FloatBarrier

\label{examples}
\begin{tightpromptbox}{Math example 1 (T=0.05, step=100)}
\textbf{root:} GRPO-1.5B-math-temp-0.05\\
\textbf{problem:} If $\arccos x + \arccos 2x + \arccos 3x = \pi,$ then $x$ satisfies a cubic polynomial of the form $ax^3 + bx^2 + cx + d = 0,$ where $a,b,c,d$ are integers and $a\neq 0$. Find the smallest possible value of $|a|+|b|+|c|+|d|$.\\
\textbf{step:} 100\quad \textbf{n\_samples:} 8\quad \textbf{shift\_rate:} 0.125\quad \textbf{freq\_correct:} 0.125\quad \textbf{$p(\checkmark\mid S{=}1)$:} 1.000\quad \textbf{gain:} +87.50 pp\\
\textbf{Shift cue:} contradiction\quad \textbf{Strategy shift:} propose $x$ $\rightarrow$ feasibility check $\rightarrow$ backtrack / replan\\
\textbf{Shift excerpt (PASS-1):} \textbf{[SHIFT at “However, substituting $x=0$ … does not satisfy”]}\\
\textbf{markers:} contradiction\\
\ldots x = 0. \textbf{However, substituting $x = 0$ into the original equation does not satisfy it} becausecontradicts $\pi$. Therefore, let's consider other values of $x$ that might work. We need to solve the system of equations formed \dots\\
\textbf{PASS-1 \tagtt{answer}:} $-1/27$
\end{tightpromptbox}

\begin{tightpromptbox}{Math example 2 (T=0.05, step=850)}
\textbf{root:} GRPO-1.5B-math-temp-0.05\\
\textbf{problem:} Let $a,b,c,d$ be positive real numbers such that $a+b+c+d=1$. Find the minimum value of $\frac{1}{a}+\frac{1}{b}+\frac{4}{c}+\frac{16}{d}$.\\
\textbf{step:} 850\quad \textbf{n\_samples:} 8\quad \textbf{shift\_rate:} 0.125\quad \textbf{freq\_correct:} 0.125\quad \textbf{$p(\checkmark\mid S{=}1)$:} 1.000\quad \textbf{gain:} +87.50 pp\\
\textbf{Shift cue:} instead\quad \textbf{Strategy shift:} dead-end/degenerate reasoning $\rightarrow$ switch to AM-GM template\\
\textbf{Shift excerpt (PASS-1):} \textbf{[SHIFT at “Instead, let’s consider … AM-GM”]}\\
\textbf{markers:} instead\\
\ldots we get:
\[
\frac{1}{a} + \frac{1}{b} + \frac{4}{c} + \frac{16}{d} = \infty.
\]
However, this approach does not yield a finite minimum.\textbf{Instead, let's consider another strategy using the Arithmetic Mean-Geometric Mean Inequality (AM-GM):}
$\cdots$ \\
\textbf{PASS-1 \tagtt{answer}:} $64$
\end{tightpromptbox}

\bigskip
\bigskip
\bigskip
\bigskip
\bigskip
\bigskip

\begin{tightpromptbox}{Math example 3 (T=0.05, step=400)}
\textbf{root:} GRPO-1.5B-math-temp-0.05\\
\textbf{problem:} Let $x_1,x_2,x_3,y_1,y_2,y_3$ be real numbers such that
\begin{align*}
(x_1 - x_2)^2 + (y_1 - y_2)^2 &= 9, \\
(x_1 - x_3)^2 + (y_1 - y_3)^2 &= 16, \\
(x_2 - x_3)^2 + (y_2 - y_3)^2 &= 25.
\end{align*}
Find $\begin{vmatrix} x_1 & y_1 & 1 \\ x_2 & y_2 & 1 \\ x_3 & y_3 & 1 \end{vmatrix}^2$.\\
\textbf{step:} 400\quad \textbf{n\_samples:} 8\quad \textbf{shift\_rate:} 0.125\quad \textbf{freq\_correct:} 0.125\quad \textbf{$p(\checkmark\mid S{=}1)$:} 1.000\quad \textbf{gain:} +87.50 pp\\
\textbf{Shift cue:} instead\quad \textbf{Strategy shift:} direct determinant manipulation $\rightarrow$ switch to “consider determinant itself” (reframe)\\
\textbf{Shift excerpt (PASS-1):} \textbf{[SHIFT at “Instead, we will consider the determinant itself …”]}\\
\textbf{markers:} instead\\
\ldots However, finding the determinant directly from these equations might not be straightforward.\textbf{Instead, we will consider the determinant itself and see if we can simplify our approach. $\cdots$ }\\
\textbf{PASS-1 \tagtt{answer}:} 144 
\end{tightpromptbox}

\begin{tightpromptbox}{Math example 4 (T=0.7, step=100)}
\textbf{root:} GRPO-1.5B-math-temp-0.7\\
\textbf{problem:} A hexagon is inscribed in a circle (diagram omitted). What is the measure of $\alpha$, in degrees?\\
\textbf{step:} 100\quad \textbf{n\_samples:} 8\quad \textbf{shift\_rate:} 0.125\quad \textbf{freq\_correct:} 0.125\quad \textbf{$p(\checkmark\mid S{=}1)$:} 1.000\quad \textbf{gain:} +87.50 pp\\
\textbf{Shift cue:} however\quad \textbf{Strategy shift:} algebra gives invalid angle $\rightarrow$ constraint check (angle must be positive) $\rightarrow$ sign correction\\
\textbf{Shift excerpt (PASS-1):} \textbf{[SHIFT at “However, since $\alpha$ … should be positive”]}\\
\textbf{markers:} however\\
\ldots Simplifying, \( \alpha + 110^\circ = 70^\circ \), leading to \( \alpha = -40^\circ \).
\textbf{However, since \( \alpha \) represents an angle within the circle, it should be positive.} $\cdots$  \\
\textbf{PASS-1 \tagtt{answer}:} $40$
\end{tightpromptbox}

\bigskip
\bigskip
\bigskip

\begin{tightpromptbox}{Math example 5 (T=0.7, step=900)}
\textbf{root:} GRPO-1.5B-math-temp-0.7\\
\textbf{problem:} Compute $\sin^3 18^\circ + \sin^2 18^\circ$.\\
\textbf{step:} 900\quad \textbf{n\_samples:} 8\quad \textbf{shift\_rate:} 0.125\quad \textbf{freq\_correct:} 0.125\quad
$\mathbf{p}(\checkmark\mid S{=}1)$\textbf{:} 1.000\quad \textbf{gain:} +87.50 pp\\
\textbf{Shift cue:} instead\quad \textbf{Strategy shift:} failed analytic attempt $\rightarrow$ abandon/switch to numerical/other-method talk (search collapse)\\
\textbf{Shift excerpt (PASS-1):} \textbf{[SHIFT at “Instead, let's try solving numerically …”]}\\
\textbf{markers:} instead\\
\ldots Since $\sqrt{5} > 3$, $y = \frac{1}{2}$ does not satisfy the equation.\textbf{Instead, let's try solving numerically or by another method.} $\cdots$ \\
\textbf{PASS-1 \tagtt{answer}:} $-1/8$
\end{tightpromptbox}

\begin{tightpromptbox}{Xwords example 1 (T=0.7, step=1000)}
\textbf{root:} GRPO-1.5B-xword-temp-0.7\\
\textbf{problem:} Front of ship; bend (3)\\
\textbf{step:} 1000\quad \textbf{n\_samples:} 8\quad \textbf{shift\_rate:} 0.125\quad \textbf{freq\_correct:} 0.125\quad \textbf{$p(\checkmark\mid S{=}1)$:} 1.000\quad \textbf{gain:} +87.50 pp\\
\textbf{Shift cue:} doesn't fit\quad \textbf{Strategy shift:} definition mismatch $\rightarrow$ reinterpret “front of ship” $\rightarrow$ BOW\\
\textbf{Shift excerpt (PASS-1):} \textbf{[SHIFT at “However, considering … ship’s front …”]}\\
\textbf{markers:} doesn't fit\\
\ldots the definition does not fit directly. \textbf{However, considering the context of a ship’s front,} \ldots \\
\textbf{PASS-1 \tagtt{answer}:} BOW
\end{tightpromptbox}

\begin{tightpromptbox}{Xwords example 2 (T=0.7, step=300)}
root: GRPO-1.5B-xword-temp-0.7\\
problem: Clue: Celebrity from stare losing tail (4)\texttt{\string\n}Gold: STAR\\
step: 300\quad n\_samples: 8\quad shift\_rate: 0.125\quad freq\_correct: 0.125\quad p\_correct\_given\_shift: 1.000\quad gain: +87.50 pp\\
Shift excerpt (PASS-1): [SHIFT: doesn't fit]\quad Strategy shift: length mismatch $\rightarrow$ simplify transformation\\
markers: doesn't fit\\
\ldots  Attempt with “TILL”: Definition: “Celebrity” could be fitting given the context, but the length is still too much.\\ This seems too long and doesn't fit easily into the 4-letter constraint. Wordplay: STARE loses its TAIL → STAR.
PASS-1 \tagtt{answer}: STAR
\end{tightpromptbox}

\begin{tightpromptbox}{Xwords example 3 (T=0.3, step=550)}
root: GRPO-1.5B-xword-temp-0.3\\
problem: Clue: Shoe liner at home on fish (6)\texttt{\string\n}Gold: INSOLE\\
step: 550\quad n\_samples: 8\quad shift\_rate: 0.250\quad freq\_correct: 0.250\quad p\_correct\_given\_shift: 0.500\quad gain: +25.00 pp\\
Shift excerpt (PASS-1): [SHIFT: doesn't fit / re-evaluate]\quad Strategy shift: reject homophone $\rightarrow$ switch to charade\\
markers: doesnt fit, re-evaluate\\
\ldots Homophone? ("shoe" sounds like "soe") - Doesn't fit here\\
Candidate attempt (WRONG PATH):
- HIDDEN: No hidden indicator present
- WRONG PATH: Consider other indicators\\
Re-evaluate as CHARADES: IN ("at home") + SOLE ("fish") → INSOLE.
Accounting: INSOLE letters: I N S O L E (6). Definition "Shoe liner" fits. Enumeration (6) OK.\\
Final answer: INSOLE\\
PASS-1 \tagtt{answer}: INSOLE
\end{tightpromptbox}

\begin{tightpromptbox}{Xwords example 4 (T=0.05, step=900)}
root: GRPO-1.5B-xword-temp-0.05\\
problem: Clue: Frighten when energy enters scar (5)\texttt{\string\n}Gold: SCARE\\
step: 900\quad n\_samples: 8\quad shift\_rate: 0.250\quad freq\_correct: 0.125\quad p\_correct\_given\_shift: 0.500\quad gain: +37.50 pp\\
Shift excerpt (PASS-1): [SHIFT: rethink / re-evaluate]\quad Strategy shift: failed parse $\rightarrow$ substitution re-parse\\
markers: rethink, re-evaluate\\
\ldots Device matches, but there’s no indication that it should be considered an abbreviation. Candidate attempt (wrong path): — fails definition, so we mustrethink. Re-evaluate as SUBSTITUTION: SCARE → SCARE. Accounting: SCARE letters: S C A R E (5). Definition “Frighten” fits. Enumeration (5) OK.\\
PASS-1 \tagtt{answer}: SCARE
\end{tightpromptbox}

\begin{tightpromptbox}{RHour example 1 (T=0, step=950)}
\textbf{root:} GRPO-1.5B-carpark-temp-0\\
\textbf{problem:} example\_id: idx\_101\texttt{\string\n}Gold: ['Cv2', 'A>1']\\
\textbf{step:} 950\quad \textbf{n\_samples:} 8\quad \textbf{shift\_rate:} 1.000\quad \textbf{freq\_correct:} 0.000\quad \textbf{$p(\checkmark\mid S{=}1)$:} 0.000\quad \textbf{gain:} +0.00 pp\\
\textbf{Shift cue:} hang on\quad \textbf{Strategy shift:} attempt $\rightarrow$ detect blocker $\rightarrow$ replan with enabling move\\
\textbf{Shift excerpt (PASS-1):} \textbf{[SHIFT at “Hang on...”]}\\
\textbf{markers:} hang on\\
\ldots S1: E>1 \(\rightarrow\) GOAL\(\times\) (blocked by D). \textbf{Hang on}\ldots\\
S2: Fv1,E>1 \(\rightarrow\) GOAL\(\checkmark\) (all valid sequences pass this check).\\
\textbf{PASS-1 \tagtt{answer}:} Fv1,E>1
\end{tightpromptbox}

\begin{tightpromptbox}{RHour example 2 (T=0, step=600)}
\textbf{root:} GRPO-1.5B-carpark-temp-0\\
\textbf{problem:} example\_id: idx\_109\texttt{\string\n}Gold: ['Cv1', 'Cv1', 'A>2']\\
\textbf{step:} 600\quad \textbf{n\_samples:} 8\quad \textbf{shift\_rate:} 0.375\quad \textbf{freq\_correct:} 0.000\quad \textbf{$p(\checkmark\mid S{=}1)$:} 0.000\quad \textbf{gain:} +0.00 pp\\
\textbf{Shift cue:} check again\quad \textbf{Strategy shift:} local fix $\rightarrow$ explicit verification pass $\rightarrow$ continue search\\
\textbf{Shift excerpt (PASS-1):} \textbf{[SHIFT at “Check again …”]}\\
\textbf{markers:} check again\\
\ldots Move B >1 to avoid collision with D. \textbf{Check again for collisions after moving B:} No more collisions detected.\ldots\\
\textbf{PASS-1 \tagtt{answer}:} C>1,E>1
\end{tightpromptbox}

\AtEndDocument{%
  \clearpage
  \section{License}
  \label{app:license}

  \begingroup\small
  \noindent\textbf{MIT License}\\
  \noindent Copyright (c) 2025 \textit{Liv d'Aliberti, Manoel Ribeiro}

  \medskip
  \noindent Permission is hereby granted, free of charge, to any person obtaining a copy
  of this software and associated documentation files (the “Software”), to deal in
  the Software without restriction, including without limitation the rights to use,
  copy, modify, merge, publish, distribute, sublicense, and/or sell copies of the
  Software, and to permit persons to whom the Software is furnished to do so,
  subject to the following conditions:

  \medskip
  \noindent The above copyright notice and this permission notice shall be included in
  all copies or substantial portions of the Software.

  \medskip
  \noindent THE SOFTWARE IS PROVIDED “AS IS”, WITHOUT WARRANTY OF ANY KIND, EXPRESS OR
  IMPLIED, INCLUDING BUT NOT LIMITED TO THE WARRANTIES OF MERCHANTABILITY, FITNESS
  FOR A PARTICULAR PURPOSE AND NONINFRINGEMENT. IN NO EVENT SHALL THE AUTHORS OR
  COPYRIGHT HOLDERS BE LIABLE FOR ANY CLAIM, DAMAGES OR OTHER LIABILITY, WHETHER
  IN AN ACTION OF CONTRACT, TORT OR OTHERWISE, ARISING FROM, OUT OF OR IN
  CONNECTION WITH THE SOFTWARE OR THE USE OR OTHER DEALINGS IN THE SOFTWARE.
  \endgroup
}

\end{document}